\documentclass{article}

\usepackage[margin=1in]{geometry}
\usepackage{authblk}

\usepackage[round]{natbib}

\usepackage[utf8]{inputenc}
\usepackage[T1]{fontenc}
\usepackage{url}
\usepackage{amsmath,amssymb,amsfonts}
\usepackage{bm}
\usepackage{graphicx}
\usepackage{rotating}
\usepackage{booktabs}
\usepackage{multirow}
\usepackage{subcaption}
\usepackage{algorithm}
\usepackage{algorithmic}
\usepackage{xcolor}
\usepackage{hyperref}
\usepackage{placeins}

\newcommand{\R}{\mathbb{R}}
\newcommand{\SO}{\mathrm{SO}(3)}

\newcommand{\hatmap}[1]{\left[#1\right]_{\times}}
\newcommand{\expmap}{\mathrm{Exp}}

\newcommand{\bI}{\mathbf{I}}
\newcommand{\bR}{\mathbf{R}}
\newcommand{\bomega}{\bm{\omega}}

\newcommand{\btau}{\bm{\tau}}
\newcommand{\bu}{\mathbf{u}}
\newcommand{\be}{\mathbf{e}}

\newcommand{\norm}[1]{\left\|#1\right\|}

\newtheorem{proposition}{Proposition}

\author[1]{Thribhuvan Rapolu\thanks{These authors contributed equally to this work.}}
\author[2]{Luke McLennan\protect\footnotemark[1]}
\author[1,2]{Chandrajit Bajaj}

\affil[1]{Department of Computer Science, University of Texas at Austin}
\affil[2]{Oden Institute of Computational Engineering \& Sciences, University of Texas at Austin}

\begin{document}

\title{Stochastic Port-Hamiltonian Gaussian Processes \\ for Learning from Noisy Dynamics on Lie Groups}

\maketitle

\begin{abstract}
    Learning predictive models for dynamics is difficult because observations are often sparse and noisy, while the underlying dynamics evolve on nonlinear configuration manifolds and are affected by environmental dissipation and stochastic disturbances. Standard neural dynamics models can fit short-term trajectories but often violate physical structure, accumulate error over long horizons, or rely on parameterizations that introduce singularities such as gimbal lock. We introduce LieSPHGP, a stochastic Gaussian-process framework for learning rigid body dynamics with port-Hamiltonian structure and stability-aware robustness. LieSPHGP represents motion on Lie groups and Lie algebras, enabling coordinate-consistent optimization without singular parameterizations. Learned mass, potential energy, dissipation, and control components retain physical roles and a separate diffusion term represents stochastic forcing. Each component is modeled by Matérn kernel Gaussian processes which capture uncertainty induced by noisy observations. To improve physical validity, Lie group exponential map updates preserve rotational constraints, while positive-definite damping ensures nonnegative deterministic energy dissipation. The implicit-explicit integrator additionally provides unconditional stability for the frozen linear damping subsystem. The resulting model learns predictive dynamics that are robust to noise, respect rigid body geometry, and capture both intrinsic motion and environmental dissipation. Experiments on various noisy rigid body benchmarks show improved prediction accuracy, stronger long-horizon stability, and better satisfaction of manifold constraints compared with neural dynamics baselines. Code, data and interactive results are available on the project website: \url{https://thribhuvanrapolu.github.io/liesphgp/}.
\end{abstract}

\section{Introduction}
Learning predictive models of rigid body motion from noisy trajectory data is a fundamental challenge in robotics, computer vision, and computational physics. Real-world systems such as robot manipulators, quadrotors, and biomechanical structures exhibit dynamics evolving on nonlinear configuration spaces and subject to dissipation, external forcing, and stochastic perturbations. Classical Hamiltonian mechanics provide a principled, energy-based foundation for modeling such systems \citep{arnold1989mechanics,marsden1999symmetry}, but they naturally describe conservative dynamics and therefore cannot directly encode dissipation and control inputs.

Port-Hamiltonian models extend Hamiltonian mechanics to controlled and dissipative systems while preserving physical structure \citep{morrison1986diss,morrison2024metriplectic,van2014port}. Learning-based Hamiltonian and port-Hamiltonian models encode this structure \citep{greydanus2019hnn,sosanya2022dhnn,desai2021phnn,duruisseaux2023liefvin,duong2024phnode}, while SDE and GP approaches model stochasticity and uncertainty \citep{boninsegna2018ssr,liu2019nsde,kidger2021infgan,tanaka2022ssgp,beckers2022gpphs,ross2023hgp}. Recent approaches have combined Gaussian processes with Lie group geometry \citep{li2025physics} and port-Hamiltonian structure \citep{leung2026learning,courteville2026,zaspel2024data}, while stochastic port-Hamiltonian neural networks have incorporated diffusion and passivity constraints \citep{di2026stochastic}. Our work focuses on identifying controlled and dissipative stochastic rigid-body dynamics on Lie groups from noisy observations, using structured Gaussian-process components that also support downstream model-based control.

\begin{figure}
    \centering
    \includegraphics[width=1\linewidth]{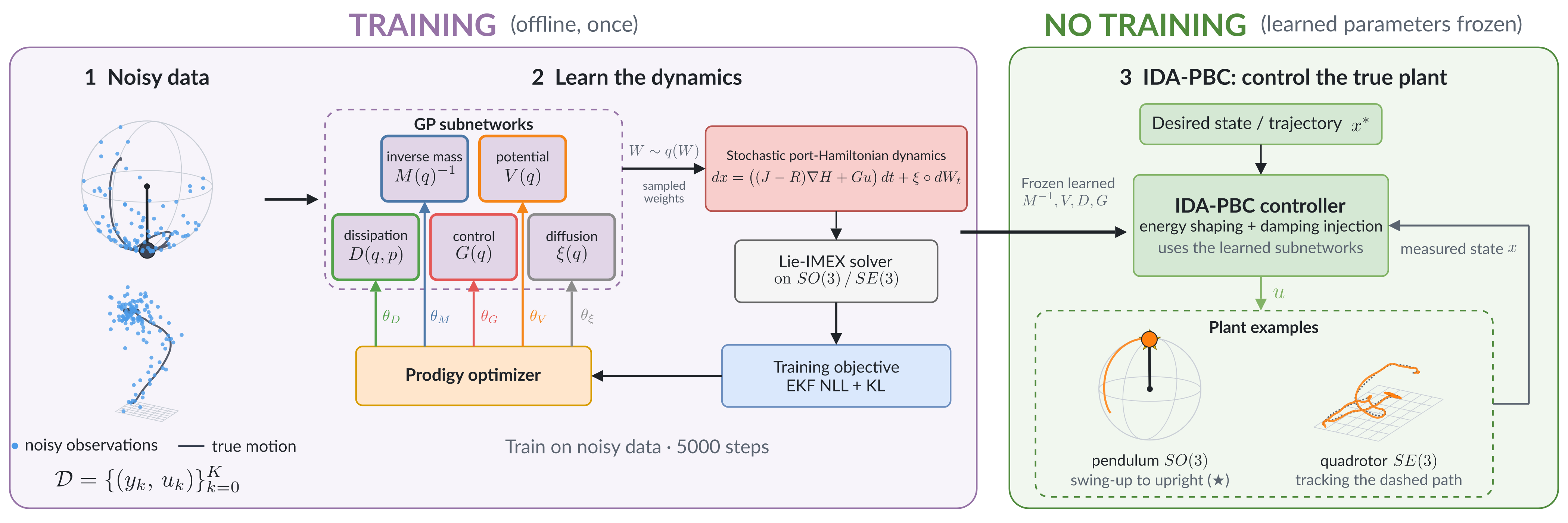}
    \caption{\textbf{Overview of the proposed framework.} Given noisy observations of a dynamical system, we learn a physics-structured model that captures both the underlying dynamics and their uncertainty while preserving the geometric structure of the state space. The learned model is then used to design a controller that drives the true system toward a desired state or trajectory without additional training. We demonstrate the framework on a 3D pendulum evolving on $SO(3)$ and a quadrotor evolving on $SE(3)$.}
    \label{fig:overview}
\end{figure}

We introduce \textit{LieSPHGP (Lie group Stochastic Port-Hamiltonian Gaussian Processes)}, a framework that unifies port-Hamiltonian structure, Lie-group geometry, Gaussian-process parameterizations, and stochastic differential equation modeling. We study two coupled learning problems. The first problem is dynamics identification from noisy state-control trajectories, which is to learn the parameters of the structured subnetworks of the stochastic port-Hamiltonian model so that it accurately explains the observed stochastic dynamics while respecting the underlying Lie-group geometry. The second problem is model-based control, in which after identifying the dynamics, the learned model is used as the plant model for a structure-preserving controller to drive the learned closed-loop dynamics toward a desired target or reference trajectory. Thus, the first problem estimates the dynamics from noisy data, while the second asks whether the learned physics is accurate enough to support downstream control. Figure~\ref{fig:overview} gives an overview of the approach.

Our contributions can be summarized as follows:
\begin{itemize}
    \item \textbf{Stochastic port-Hamiltonian learning on Lie groups.} Building on deterministic Lie-group port-Hamiltonian models \citep{duong2024phnode} and stochastic port-Hamiltonian formulations \citep{di2026stochastic}, we develop a Stratonovich SDE model for learning controlled and dissipative rigid-body dynamics on $SO(3)$ and $SE(3)$. The formulation explicitly models stochastic environmental forcing through a learned diffusion coefficient, distinguishing physical process noise from uncertainty in the learned dynamics and noisy observations (Section~\ref{sec:methods:ph_form}).

    \item \textbf{Geometry-preserving stochastic integration.} Building on Lie-group integration methods \citep{duruisseaux2023liefvin,gay2024variational} and implicit-explicit numerical schemes \citep{pareschi2000implicit,zhang2025semi}, we combine exponential-map updates with an implicit treatment of learned dissipation. The resulting Lie-IMEX integrator preserves rotational manifold constraints and provides unconditional stability for the frozen linear damping subsystem, addressing numerical difficulties caused by large learned damping coefficients (Section~\ref{sec:methods:integrator} and Appendix~\ref{sec:appendix:stability}).

    \item \textbf{Structured probabilistic identification from noisy trajectories.} Extending Gaussian process Hamiltonian learning \citep{mclennan2025rghnet,leung2026learning} and physics-informed Gaussian processes on Lie groups \citep{li2025physics}, we parameterize inverse mass, potential energy, dissipation, control, and diffusion using Mat\'ern random Fourier feature subnetworks. We jointly train these physically constrained components using variational inference and a differentiable Lie group extended Kalman filter, which approximately marginalizes latent states and accounts for both process and observation noise (Sections~\ref{sec:methods:gp} and~\ref{sec:methods:loss}).

    \item \textbf{Experimental evaluation of dynamics identification and control.} We evaluate LieSPHGP on stochastic $SO(3)$ and $SE(3)$ benchmarks and real-world rigid-body trajectories, comparing against deterministic port-Hamiltonian models and stochastic neural alternatives. We assess long-horizon prediction accuracy, probabilistic predictive quality, manifold preservation, physical parameter identification, and downstream IDA-PBC control, demonstrating improvements in several metrics (Section~\ref{sec:experiments}).
\end{itemize}

\section{Background}
\label{sec:background}

\subsection{Port-Hamiltonian Dynamics}
Classical Hamiltonian mechanics provides a geometric foundation for modeling physical systems in terms of energy-preserving flows on phase space \citep{arnold1989mechanics}. For a system with configuration $q$ and conjugate momentum $p$, the Hamiltonian $H(q,p)$ encodes the total energy, and the dynamics evolve according to Hamilton's equations. While classical formulations operate on $\mathbb{R}^{2d}$, the modern theory of geometric mechanics extends Hamiltonian dynamics to general symplectic and Poisson manifolds \citep{marsden1999symmetry}, enabling the study of systems whose configuration spaces possess nonlinear geometry, such as Lie groups. 

Many real-world systems do not evolve purely under conservative dynamics. Environmental interactions introduce dissipation, noise, and external forces. Classical Hamiltonian mechanics alone cannot capture these effects, motivating generalized formulations such as metriplectic dynamics \citep{morrison1986diss,morrison2024metriplectic} and port-Hamiltonian systems \citep{van2014port}, which introduce structured dissipative brackets and external ports while preserving the underlying geometric structure. These frameworks naturally integrate control inputs, energy exchange, and dissipation into the dynamics. Port-Hamiltonian systems \citep{van2014port} extend traditional Hamiltonian dynamics by introducing a positive semi-definite dissipation matrix $\mathcal{R}(q,p)$ and external forcing through a port term $\mathcal{G}(q,p)u$, with control input $u$ and control output $y$. The dynamics become:
\begin{equation}
\begin{aligned}
\label{eq:phs}
    \dot{x} &= 
    \begin{bmatrix}
        \dot{q} \\ \dot{p}
    \end{bmatrix}
    =
    (J(q,p) - \mathcal{R}(q,p))\,\nabla H(q,p)
    + \mathcal{G}(q,p)\, u, \\
    y &= \mathcal{G}(q,p)^\top \nabla H(q,p).
\end{aligned}
\end{equation}
This port-Hamiltonian form enables the modeling of controlled mechanical systems such as robot joints, quadrotors, and other articulated agents. These systems preserve the Hamiltonian structure while incorporating control and dissipation, making them suitable for learning-based modeling of physical dynamics \citep{desai2021phnn,duong2024phnode}.

Stochastic extensions of port-Hamiltonian systems introduce additional challenges because positive semidefinite dissipation alone does not guarantee non-increasing energy under stochastic forcing. Recent work has developed stochastic passivity conditions for SDE-based port-Hamiltonian systems \citep{ackermann2026} and investigated stochastic energy shaping through invariant measures \citep{cordoni2025}. These results motivate distinguishing deterministic dissipativity from stochastic stability when learning dynamics subject to environmental disturbances. \citet{lamoline2024dirac} introduced stochastic ports into the Dirac-structure formulation of a class of infinite-dimensional port-Hamiltonian systems using Stratonovich calculus. Their work studies the geometric structure and interconnection of these systems, while our focus is learning finite-dimensional rigid-body dynamics on Lie groups from noisy observations.

\subsection{Learning Port-Hamiltonian Dynamics}

Machine learning approaches to dynamics identification have developed from unconstrained function approximation toward models that incorporate physical structure. Neural ordinary differential equations (NODEs) \citep{chen2018node} parameterize continuous-time vector fields using neural networks. Neural stochastic differential equations \citep{liu2019nsde,kidger2021infgan} additionally model stochastic dynamics. Physics-informed approaches incorporate mechanical principles into learned dynamics. Deep Lagrangian Networks \citep{lutter2019deep,lutter2023combining} parameterize kinetic and potential energies and enforce the Euler-Lagrange equations, while Lagrangian Neural Networks \citep{cranmer2020lagrangian} learn general Lagrangian functions without requiring canonical momentum coordinates. \citet{zhong2020unsupervised} further developed unsupervised Lagrangian dynamics learning from images for prediction and control. These approaches demonstrate the advantages of continuous-time modeling and physical inductive biases, motivating the incorporation of geometric constraints, dissipation, stochastic forcing, and uncertainty quantification in learned dynamical systems.

Learning Hamiltonian mechanics from data has gained prominence due to the interpretability and structure preservation of energy-based models. Hamiltonian neural networks \citep{greydanus2019hnn}, dissipative HNNs \citep{sosanya2022dhnn}, and port-Hamiltonian neural networks \citep{desai2021phnn} learn energy functions or structure matrices directly from data. Gaussian process approaches \citep{tanaka2022ssgp,beckers2022gpphs,ross2023hgp,mclennan2025rghnet} introduce uncertainty quantification and improved robustness. More recently, \citet{leung2026learning} developed multistep port-Hamiltonian Gaussian processes that incorporate dynamical constraints as linear functionals of a Hamiltonian GP. Our method instead learns separate, physically constrained GP subnetworks through an approximate likelihood computed by a Lie group extended Kalman filter. This allows us to model nonlinear interactions between the learned physical components and account for latent-state uncertainty and stochastic diffusion, at the cost of approximate inference. \citet{zaspel2024data} considered GP-based identification of port-Hamiltonian differential-algebraic systems. \citet{courteville2026} and \citet{yang2026structure} extended GP-based port-Hamiltonian modeling to distributed systems using finite-element formulations and structure-aware spatial kernels. While these approaches incorporate physical structure into Gaussian process learning, they primarily address deterministic dynamics and do not explicitly learn stochastic diffusion or distinguish uncertainty in learned physical parameters, stochastic process noise, and measurement noise. Additionally, these approaches are restricted to Euclidean spaces, while other works \citep{duruisseaux2023liefvin, duong2024phnode, li2025physics} and our work extend to system on Lie groups and algebras.  Other approaches address complementary aspects of learning and numerical stability. \citet{bhardwaj2026phast} introduced PHAST, which combines structured port-Hamiltonian learning with Strang splitting for stable long-horizon forecasting, but does not explicitly learn stochastic diffusion or employ intrinsic Lie group integration. \citet{mclennan2025rghnet} developed a probabilistic framework for learning generalized Hamiltonian dynamics from noisy observations using Gaussian processes and stability regularization, primarily in Euclidean phase spaces and without explicitly learned SDE diffusion. \citet{di2026stochastic} introduced stochastic port-Hamiltonian neural networks (SPH-NNs) with universal approximation results and weak passivity guarantees in expectation, but they parameterize the Hamiltonian as a single unstructured network $H_\theta(q,p)$ on Euclidean coordinates, assumes clean full-state observations without separating measurement from process noise or providing uncertainty quantification, and leaves feedback control to future work. Our approach addresses these limitations by learning structured GP subnetworks on Lie groups through an extended Kalman filter likelihood on noisy observations and using them for IDA-PBC control. 

\subsection{Learning Dynamics on Lie Groups}
Many physical systems, including rigid-body systems, evolve on Lie groups. Lie group variational integrator networks \citep{duruisseaux2023liefvin} and port-Hamiltonian neural ODEs on Lie groups \citep{duong2024phnode} demonstrate the importance of respecting manifold geometry in learned dynamics. Gaussian-process learning has also been extended beyond Euclidean spaces through stationary kernels on compact Lie groups and noncompact symmetric spaces \citep{azangulov2024compact,azangulov2024noncompact}. More directly, \citet{li2025physics} developed physics-informed Gaussian-process learning for mechanical systems on Lie groups. Our approach extends this direction to controlled and dissipative stochastic port-Hamiltonian dynamics, learning physical components and stochastic diffusion with separate models for process and observation noise, and using the identified physical components for downstream control. Numerical integration is also central to preserving the geometry of these systems. \citet{gay2024variational} developed stochastic variational integrators on Lie groups that preserve geometric structures such as symplecticity, Lie-Poisson structure, and coadjoint orbits. Our Lie-IMEX integrator instead prioritizes rotational manifold preservation and stable treatment of learned dissipation rather than symplecticity of the stochastic flow. Together, these components address the joint problem of learning stochastic port-Hamiltonian dynamics on Lie groups through Gaussian processes, combining uncertainty-aware inference from noisy rigid-body trajectories with geometry-preserving stochastic integration. Table~\ref{tab:prior} summarizes the distinctions between our method and prior approaches.

\begin{table*}[t]
\centering
\caption{Summary of references by dynamics type, architecture, geometry, and differential equation type, sorted by year.}
\label{tab:prior}
\small
\begin{tabular}{lcccc}
\toprule
\textbf{Method} & \textbf{Dynamics Type} & \textbf{Architecture} & \textbf{Geometry} & \textbf{DE Type} \\
\midrule
NODE \citep{chen2018node} & No Hamiltonian & Neural Network & Euclidean & ODE \\
HNN \citep{greydanus2019hnn} & Hamiltonian & Neural Network & Euclidean & ODE \\
NSDE \citep{liu2019nsde} & No Hamiltonian & Neural Network & Euclidean & SDE \\
VIN \citep{saemundsson2020vin} & Lagrangian & Neural Network & Euclidean & ODE \\
PHNN \citep{desai2021phnn} & Port-Hamiltonian & Neural Network & Euclidean & ODE \\
InfGAN \citep{kidger2021infgan} & No Hamiltonian & Neural Network & Euclidean & SDE \\
SSGP \citep{tanaka2022ssgp} & Hamiltonian & Gaussian Process & Euclidean & ODE \\
GP-PHS \citep{beckers2022gpphs} & Port-Hamiltonian & Gaussian Process & Euclidean & ODE \\
SINN \citep{zhu2023sinn} & No Hamiltonian & Neural Network & Euclidean & SDE \\
LieFVIN \citep{duruisseaux2023liefvin} & Lagrangian & Neural Network & Lie Groups & ODE \\
HGP \citep{ross2023hgp} & Hamiltonian & Gaussian Process & Euclidean & ODE \\
PH-NODE \citep{duong2024phnode} & Port-Hamiltonian & Neural Network & Lie Groups & ODE \\
DynGMA \citep{zhu2024dyngma} & No Hamiltonian & Gaussian Mixture & Euclidean & SDE \\
GP-PHDAE \citep{zaspel2024data} & Port-Hamiltonian & Gaussian Process & Euclidean & DAE \\
LGHD \citep{mclennan2025rghnet} & Port-Hamiltonian & Gaussian Process & Euclidean & ODE \\
MoSim \citep{hao2025nms} & No Hamiltonian & Neural Network & Euclidean & ODE \\
PI-GP-Lie \citep{li2025physics} & Lagrangian & Gaussian Process & Lie Groups & ODE \\
PHAST \citep{bhardwaj2026phast} & Port-Hamiltonian & Neural Network & Euclidean & ODE \\
MS-PHS GP \citep{leung2026learning} & Port-Hamiltonian & Gaussian Process & Euclidean & ODE \\
PFEM-GP-dPHS \citep{courteville2026} & Port-Hamiltonian & Gaussian Process & Euclidean & PDE \\
SA-GP-dPHS \citep{yang2026structure} & Port-Hamiltonian & Gaussian Process & Euclidean & PDE \\
SPH-NN \citep{di2026stochastic} & Port-Hamiltonian & Neural Network & Euclidean & SDE \\
\midrule
\textbf{This paper} & Port-Hamiltonian & Gaussian Process & Lie Groups & SDE \\
\bottomrule
\end{tabular}
\end{table*}

\subsection{Controlling Port-Hamiltonian Systems}
Port-Hamiltonian models describe energy, dissipation, and input effects explicitly, so physics-aware controllers can be built to reshape energy and inject damping \citep{ortega2002idapbc,van2014port}. Previous work learns controllers for known models \citep{massaroli2022optimal,plaza2022total}. \citet{kamboj2026colearning} instead jointly learns the model and controller through alternating optimization that requires repeated data collection, but this can be costly on physical systems. Following \citet{duong2024phnode}, we first learn the dynamics and then construct the controller without further training. We use IDA-PBC \citep{ortega2002idapbc} to obtain $\dot{x} = (J_{cl}-R_{cl})\nabla H_{cl}$, where $H_{cl}=H+H_a$ has its minimum at the desired state and added damping helps the system reach it. Control performance is therefore an additional test of accuracy of the learned physics. Additional details are in Appendix~\ref{ida_pbc_SO3}.

\section{Methods}
\label{sec:methods}

\begin{figure*}[t]
    \centering
    \includegraphics[width=1\linewidth]{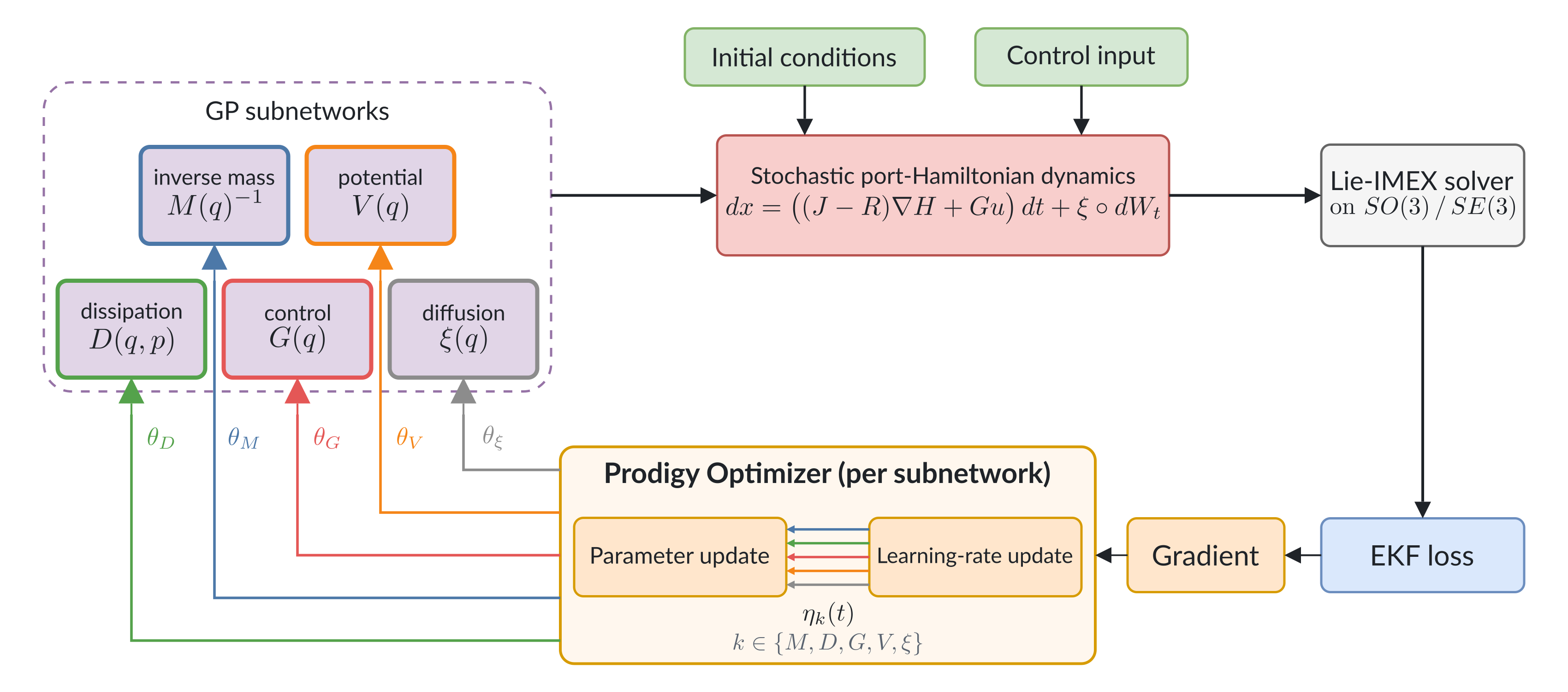}

    \caption{\textbf{LieSPHGP framework.} Five GP subnetworks ($M^{-1}$, $V$, $D$, $G$ and the diffusion $\xi$), each a level (constant, or linear for $V$) plus a Mat\'ern-kernel random-feature Gaussian process (Section~\ref{sec:methods:gp}, Algorithm~\ref{alg:gp_layer}), define a stochastic port-Hamiltonian system on $SO(3)$ or $SE(3)$ (Section~\ref{sec:methods:ph_form}), integrated with the Lie-IMEX solver (Section~\ref{sec:methods:integrator}, Algorithm~\ref{alg:lieimexsolver}). At each training step one GP weight sample $W\sim q(W)$ is drawn and held fixed throughout each rollout; starting from $y_0$, the solver gives the prediction $\hat x_k^-$ and its Jacobians $A_k$, $B_k$, and a Lie-group EKF propagates the covariance $P_k^-=A_kP_{k-1}A_k^\top+B_kB_k^\top$, so the diffusion enters only through the process covariance $B_kB_k^\top$, while the noisy observations $y_k$ and the learned measurement-noise scales $\sigma_{\mathrm{obs}}$ enter the measurement update. Training minimizes $\mathcal L=\mathbb E_{q(W)}[\text{EKF NLL}]+\beta\,\mathrm{KL}\big(q(W)\,\|\,p(W)\big)$, the EKF negative log-likelihood of the noisy windows plus a KL term on the GP weights (Section~\ref{sec:methods:loss}, Eq.~\ref{eq:ekf_nll_step}). Each parameter group has its own Prodigy optimizer, so its learning rate $\eta_k(t)$ is adapted separately during training (Algorithm~\ref{alg:phase1_sde}).}

    \label{fig:framework}
\end{figure*}

Figure~\ref{fig:framework} summarizes LieSPHGP: the structured Gaussian-process subnetworks, the stochastic port-Hamiltonian dynamics on $SO(3)$ or $SE(3)$, the Lie-IMEX integrator, and the EKF training objective, which we describe below.

\subsection{Stochastic Dynamics Form}
\label{sec:methods:ph_form}
The port-Hamiltonian dynamics described in Section \ref{sec:background} are deterministic. For some physical systems, there are stochastic forcing components, such as chaotic wind speeds. Additionally, the dynamics dataset may include measurement noise and relying on a deterministic model could cause overfitting to noise in the data. Measurement noise and stochastic forcing represent different sources of uncertainty: measurement noise corrupts observations, whereas stochastic forcing changes the underlying physical trajectories. Consequently, modeling noisy observations alone is insufficient when the physical system itself is stochastic, motivating an explicit stochastic term. Thus, we consider the addition of a stochastic term into the port-Hamiltonian dynamics \citep{cordoni2022stochastic}. These dynamics are given by an SDE rather than an ODE. The Stratonovich formulation is natural for mechanical systems on manifolds because it satisfies the ordinary chain rule under smooth coordinate transformations. This allows stochastic forcing to be expressed consistently with the underlying Lie-group geometry and the corresponding energy balance. Using a Stratonovich SDE, the dynamics are as follows:
\begin{equation}
\begin{aligned}
    dx &= \left[(J(x) - \mathcal{R}(x)) \nabla H(x) + \mathcal{G}(x) u\right]dt + \xi(x) \circ dW_t, \\
    y &= \mathcal{G}(x)^\top \nabla H(x)
\end{aligned}
\end{equation}

where $W_t$ is Brownian motion and $\xi$ is the diffusion matrix. The Hamiltonian is broken down into kinetic and potential energies, $H(x) = H(q,p) = \tfrac12 p^\top M(q)^{-1} p + V(q)$, where $x=(q,p)$, $M$ is the mass matrix, and $V$ is the potential energy.

For the $SO(3)$ Lie group, the dynamics are:
\begin{equation}
\begin{aligned}
\label{eq:ph-sde-so3}
    dq &= q\,\widehat{M^{-1}p}\,dt, \\
    dp &= \Big( p \times \nabla_p H + \sum_{i=1}^3 q_i \times \nabla_{q_i} H - D(q,p) \nabla_p H + G(q) u \Big) \, dt +  M(q)\,\xi(q) \circ dW_t,
\end{aligned}
\end{equation}
where $q\in SO(3)$, $p \in \mathfrak{so}(3)^* \cong \mathbb{R}^3$, and $D(q,p)\in\mathbb{R}^{3\times3}$ and $G(q)\in\mathbb{R}^{3\times3}$ are the momentum blocks of the full-state matrices, $\mathcal{R}=\mathrm{blkdiag}(0_{9\times9},\,D)$ and $\mathcal{G}=[\,0_{9\times3};\,G\,]$, since dissipation and control act only on $p$. The full derivation is in Appendix \ref{sec:appendix:background:so3}. Details for the $SE(3)$ Lie group are provided in Appendices~\ref{sec:appendix:background:se3} and~\ref{sec:se3-formulation}.

\paragraph{Dynamics learning problem.}
The optimization problem is to learn the stochastic port-Hamiltonian dynamics from noisy state observations and known control inputs. Let $\mathcal{D}=\{(y_k^{(j)},u_k^{(j)})\}_{j,k}$ be the observed
trajectories, where $j$ indexes trajectories and $k$ indexes samples
at times $t_k=k\Delta t$. Here, $u_k^{(j)}$ is the known control input,
and $y_k^{(j)}$ is a noisy observation of the state $x_k^{(j)}$,
expressed in $(q,\omega)$ coordinates with $\omega=M(q)^{-1}p$
(Section~\ref{sec:methods:loss}). We parameterize the Hamiltonian, dissipation, control, and diffusion terms by $\theta$, yielding $H_\theta$, $D_\theta$, $G_\theta$, and $\xi_\theta$ (equivalently $\mathcal{R}_\theta$ and $\mathcal{G}_\theta$). The goal is to identify parameters such that the model predictions are consistent with the observed trajectories while capturing both the mean evolution and the uncertainty induced by stochastic forcing. This can be formulated as a maximum marginal likelihood problem:
\begin{equation}
\begin{aligned}
\theta^\star
&=\arg\max_\theta \log p_\theta(\mathcal{D}),\\
\text{subject to}\quad
dx_t
&=\left[(J(x_t)-\mathcal{R}_\theta(x_t))\nabla H_\theta(x_t)+\mathcal{G}_\theta(x_t)u_t\right]dt+\xi_\theta(x_t)\circ dW_t.
\end{aligned}
\label{eq:dynamics_identification_problem}
\end{equation}
Here, $p_\theta(\mathcal{D})$ is the marginal likelihood of the noisy observations, accounting for both stochastic process noise and measurement noise. The resulting model describes a distribution over possible state trajectories rather than a single deterministic prediction. Computationally, each term must be discretized and estimated. In Section \ref{sec:methods:gp}, we describe the Gaussian process models and approximations used for each term of the dynamics. In Section \ref{sec:methods:integrator} we describe the numerical integration used for the stochastic port-Hamiltonian equation. In Section \ref{sec:methods:loss} we describe the loss functions used to approximate the marginal likelihood and add regularization and the method of updating the parameters of the dynamics model.

\subsection{Kernelized Gaussian Process Model}
\label{sec:methods:gp}
We replace the neural subnetworks used in Port-Hamiltonian Neural ODEs \citep{duong2024phnode} with variational Gaussian-process (GP) approximations for the structured components $M^{-1}$, $V$, $D$, $G$, and $\xi$ \citep{tanaka2022ssgp,beckers2022gpphs,ross2023hgp,mclennan2025rghnet}. Random Fourier features (RFFs) \citep{rahimi2007rff, rasmussen2006gpml} provide a finite-dimensional approximation that can be optimized jointly with the dynamics model.

\paragraph{Mat\'ern kernel.} For inputs $x,x'$ with distance $r=d(x,x')$, the Mat\'ern covariance is
\begin{equation}
k_{\mathrm M}(x,x')=\sigma_{\mathrm M}^2\frac{2^{1-\nu}}{\Gamma(\nu)}\left(\frac{\sqrt{2\nu}\,r}{\ell_{\mathrm M}}\right)^\nu K_\nu\!\left(\frac{\sqrt{2\nu}\,r}{\ell_{\mathrm M}}\right),
\end{equation}

where $\sigma_{\mathrm M}$ is the amplitude, $\ell_{\mathrm M}$ is the length scale, $\nu$ controls smoothness, and $K_\nu$ is the modified Bessel function of the second kind. We use $\nu=5/2$ in every subnetwork, and the amplitude $\sigma_{\mathrm M}$ and length scale $\ell_{\mathrm M}$ of every subnetwork are learned with the other parameters. On rotational inputs the GP takes $\mathrm{vec}(R)\in\mathbb R^9$, so the kernel depends on the chordal distance $\|R-R'\|_F=2\sqrt{2}\sin(\theta/2)$, a function of the geodesic angle $\theta$ that is invariant to rotations. Unlike a Mat\'ern kernel of the geodesic distance, which is not positive definite in general \citep{feragen2015geodesic,borovitskiy2020matern}, this kernel is the restriction of a positive-definite kernel on $\mathbb R^9$ and is therefore positive definite on $SO(3)$. Thus, although the dynamics and integration respect the Lie group structure, the Gaussian process covariance is constructed extrinsically by restricting a Euclidean kernel to embedded rotation matrices.

We use Mat\'ern-distributed frequencies to construct the finite random-feature parameterization \citep{rahimi2007rff, rasmussen2006gpml}, $\phi_n^{\mathrm M}(x)=\sigma_{\mathrm M}\sqrt{2/F}\cos(\omega_n^\top x/\ell_{\mathrm M}+b_n)$, $n=1,\dots,F$, with $\omega_n\sim\operatorname{StudentT}_{2\nu}(0,I_d)$, $b_n\sim U(0,2\pi)$ and $F=64$ features per subnetwork.

For each dynamics subnetwork we use a sum of a level (mean function) and the GP approximation. Thus, the form is:
\begin{equation}
a_\theta(x) = \lambda_f^\theta + w^\top\phi^{\mathrm M}(x), \quad w \sim \mathcal{N}(w_\mathrm{mean}^\theta, w_\mathrm{var}^\theta),
\end{equation}
where $a$ is one of the $M^{-1}, V, D, G, \xi$ subnetworks, $\lambda_f^\theta$ is a trainable constant vector/matrix level for $M^{-1}$, $D$, $G$ and $\xi$. For the potential the level is linear, $\lambda_V^\top z$ with $z=\mathrm{vec}(R)$ on $SO(3)$ and $z=x$ on $SE(3)$ (Algorithm~\ref{alg:gp_layer}), because a constant potential exerts no force. $w$ are the RFF weights, $w^{\theta}_{\mathrm{mean}}$ is a trainable vector, and $w^{\theta}_{\mathrm{var}}$ is a trainable diagonal matrix. Refer to Algorithm~\ref{alg:gp_layer} in the appendix for the kernel algorithm. 

\paragraph{Symmetric positive definite matrix subnetworks} The inverse mass $M^{-1}(R)$ and the dissipation $D(\omega)$ must be symmetric positive definite for every input and every sampled weight. Each is built using a Cholesky factorization. The model described above returns $a = \lambda+\phi(z)^\top W\in\mathbb R^6$, where $\lambda$ is a trainable constant level, $\phi$ are the Mat\'ern random features and $W$ is the sampled GP weight matrix. We set:
\begin{equation}
L(a)=\begin{bmatrix} e^{a_1} & 0 & 0\\ a_4 & e^{a_2} & 0\\ a_5 & a_6 & e^{a_3}\end{bmatrix}
\end{equation}

\begin{equation}
    \qquad M^{-1}(R)\ \text{or}\ D(\omega)=L(a)\,L(a)^\top 
\end{equation}

The exponential keeps the diagonal of $L$ positive, so $\det L>0$ and the product is strictly positive definite without any added regularizer. Note that this construction means that the resulting subnetwork $M^{-1}$ or $D$ is not Gaussian distributed. Details are given in Appendix~\ref{sec:appendix:computation:psd_gp}.

\subsection{Numerical Integration}
\label{sec:methods:integrator}
We use a Lie-IMEX (implicit-explicit) \citep{pareschi2000implicit, zhang2025semi} integrator for the learned port-Hamiltonian dynamics. The motivation is that the learned dissipation is positive definite but is not a priori bounded in magnitude; an explicit integration step can therefore become unstable when a damping eigenvalue is large relative to the timestep. Let the velocity/twist variable be $\zeta$ (for the $SO(3)$ pendulum, $\zeta=\omega\in\mathbb R^3$; for the $SE(3)$ quadrotor, $\zeta=(v_b,\omega_b)\in\mathbb R^6$). We split its drift as
\begin{equation}
\dot\zeta = F_\theta(s,u)-K_\theta(s)\zeta,
\label{eq:imex_split}
\end{equation}
where $K_\theta$ is the effective damping block and $F_\theta$ collects all remaining drift terms. For the $SO(3)$ pendulum, $K_\theta=M^{-1}(R)\,D(\omega)$; for the quadrotor, $K_\theta=\operatorname{blkdiag}(M_1^{-1}D_v,\,M_2^{-1}D_\omega)$. Each block is a product of two symmetric positive definite matrices, so its eigenvalues are real and positive.

The learned model is written in the momentum $p=M_\theta(q)\,\zeta$ (Equation~\ref{eq:ph-sde-so3}), while the integrator advances $\zeta$. Because $q$ carries no Brownian increment,

\begin{equation}
\begin{aligned}
d\zeta &= M_\theta^{-1}(q)\,dp + \Big(\tfrac{d}{dt}M_\theta^{-1}(q)\Big)\,p\,dt \\
&= M_\theta^{-1}(q)\,dp  -M_\theta^{-1}\dot M_\theta M_\theta^{-1} \,p \, dt \\
&= M_\theta^{-1}(q)\,dp  -M_\theta^{-1}\dot M_\theta \, \zeta \, dt
\label{eq:momentum_to_velocity}
\end{aligned}
\end{equation}
The additional drift term vanishes for a constant mass but not for the learned, configuration-dependent $M_\theta^{-1}(q)$; we evaluate it as the directional derivative of the $M^{-1}$ subnetwork along $\dot q$ by forward-mode automatic differentiation. Substituting the momentum drift of Equation~\ref{eq:ph-sde-so3} gives, for the $SO(3)$ pendulum,
\begin{equation}
F_\theta = M_\theta^{-1}\Big(p\times\nabla_pH+\sum_{i=1}^3 q_i\times\nabla_{q_i}H+G_\theta u\Big)+\Big(\tfrac{d}{dt}M_\theta^{-1}\Big)p,
\end{equation}
and the momentum diffusion $M_\theta(q)\,\xi_\theta(q)$ becomes the twist diffusion $\Gamma_\theta=\xi_\theta(q)$, which is the quantity the diffusion subnetwork parameterizes. On $SE(3)$ the same holds blockwise: $v_b=M_1^{-1}(x)\,p_v$ and $\omega_b=M_2^{-1}(R)\,p_\omega$, so the momentum diffusions are $M_1(x)\,\xi_v$ and $M_2(R)\,\xi_\omega$ and the twist diffusion is $\Gamma_\theta=(\xi_v,\xi_\omega)$.
Restricting the diffusion to a function of the configuration is a modelling choice as it covers attitude-dependent external forcing such as wind acting through $R^\top$. Velocity-dependent diffusion, e.g.\ turbulence scaling with airspeed, would require $\xi(q,\zeta)$ and an explicit choice of stochastic calculus.

At a substep of length $h$, evaluate $(F_n,K_n)$ at $s_n$ and solve the linearly implicit predictor
\begin{equation}
(I+hK_n)\zeta_p=\zeta_n+hF_n+\Gamma_n\Delta W,
\qquad \Delta W\sim\mathcal N(0,hI),
\label{eq:imex_predictor}
\end{equation}
where the stochastic increment is omitted for deterministic experiments. Configuration variables are predicted explicitly, while rotation is advanced on the group, e.g. $R_p=R_n\exp(h\widehat\omega_n)$. Re-evaluating $(F_p,K_p)$ at the predictor, the corrected velocity/twist solves
\begin{equation}
\begin{aligned}
\left(I+\frac h2K_p\right)\zeta_{n+1}
&=\zeta_n+\frac h2\left(F_n+F_p-K_n\zeta_n\right) +\Gamma_n\Delta W.
\end{aligned}
\label{eq:imex_corrector}
\end{equation}

Since the diffusion $\xi(q)$ depends only on the configuration and acts only on the velocity/twist, its directional derivative along each diffusion vector field vanishes. Under this assumption, the It\^o and Stratonovich forms of the SDE coincide. The step therefore evaluates $\Gamma_n=\xi(q_n)$ once at the start of the step and uses the same increment $\Gamma_n\Delta W$ in the predictor and the corrector.

The translational configuration is corrected by the trapezoidal rule, and the rotation correction uses the Lie exponential map:
\begin{equation}
\begin{aligned}
x_{n+1} &= x_n+\tfrac h2(\dot x_n+\dot x_p) \\
R_{n+1} &= R_n\exp\!\left(\frac h2\widehat{(\omega_n+\omega_p)}\right).
\end{aligned}
\end{equation}
On $SO(3)$ the exponential is evaluated with Rodrigues' formula \citep{rodrigues1840lois,lynch2017modern}, so $R_n\in SO(3)$ implies $R_{n+1}\in SO(3)$ up to floating-point error. Algorithm~\ref{alg:lieimexsolver} gives the computational form used in the experiments.

\begin{proposition}[Lie-IMEX Properties]
Consider the Lie-IMEX update. The rotational update preserves the Lie-group constraint exactly in exact arithmetic: if $R_n\in SO(3)$, then $R_{n+1}\in SO(3)$. Moreover, for the frozen scalar dissipative subsystem $\dot y=-\mu y$ with $\mu\geq0$, the trapezoidal damping corrector has amplification factor
\begin{equation}
\rho(h\mu)=\frac{1-h\mu/2}{1+h\mu/2},
\end{equation}
and therefore satisfies $|\rho(h\mu)|\leq1$ for every $h\mu\geq0$. Hence the frozen scalar damping update is unconditionally stable with respect to the timestep.
\end{proposition}

The proposition establishes two properties of the implemented integrator: exact preservation of rotational group membership and unconditional stability of the frozen linear damping subproblem. These properties do not imply symplecticity, invariant-measure preservation, or nonlinear stochastic stability of the complete learned stochastic integrator. A proof and discussion of the scope of the result are given in Appendix~\ref{sec:appendix:stability}.

\begin{algorithm}[t]
\caption{Training Lie Group Stochastic Port-Hamiltonian Gaussian Process}
\label{alg:phase1_sde}
\small
\begin{algorithmic}[1]

\REQUIRE Dataset $\mathcal{D}=\{(y_{0:T-1}^{(j)},u_{1:T-1}^{(j)})\}_{j=1}^{N_{\mathrm{win}}}$; GP subnetworks $f\in\mathcal{F}=\{M^{-1},V,D,G,\xi\}$, each $f(x)=\lambda_f+\phi_f(x)^\top W_f$; $N_{\mathrm{train}},B,N_{\mathrm{sub}}$
\ENSURE Trained parameters $\theta=(\lambda,\psi)$: levels $\lambda=\{\lambda_f\}_{f\in\mathcal{F}}$; GP weight posteriors, kernel hyperparameters and observation noise; $\psi=\{m_f,s_f,\sigma_{\mathrm M,f},\ell_{\mathrm M,f}\}_{f\in\mathcal{F}}\cup\{\sigma_{\mathrm{obs}}\}$

\STATE \textbf{\COMMENT{Initialize}}
Draw fixed RFFs $(\Omega,b)$ (Alg.~\ref{alg:gp_layer}); initialize $\lambda$ and $\psi$ (Sec.~\ref{sec:methods:loss}).

\STATE $N_{\mathrm{obs}}\gets N_{\mathrm{win}}(T-1)$, \quad $h\gets\Delta t/N_{\mathrm{sub}}$.

\FOR{$k=1,\ldots,N_{\mathrm{train}}$}

    \STATE \textbf{\COMMENT{Sample GP posterior}}
    Sample minibatch $\mathcal{B}$ and GP weights $W\sim q_\psi(W)=\prod_{f}\mathcal{N}\big(m_f,\mathrm{diag}(s_f^2)\big)$ via reparameterization (Sec.~\ref{sec:methods:gp}).

    \STATE For each $j\in\mathcal{B}$, set $\hat{x}_0\gets y_0^{(j)}$, $P_0\gets R_{\mathrm{obs}}$, and $\mathcal{N}^{(j)}\gets0$; for $t=1,\ldots,T-1$:

    \STATE \textbf{\COMMENT{Forward dynamics}}
    Compute $\hat{x}_t^{-}$ and Jacobians $(A_t,B_t)$ with $N_{\mathrm{sub}}$ Lie-IMEX steps under $u_t^{(j)}$ (Sec.~\ref{sec:methods:integrator}, Alg.~\ref{alg:lieimexsolver}).

    \STATE \textbf{\COMMENT{EKF step}}
    Compute $(\hat{x}_t,P_t,\ell_t^{(j)})$ by the Lie-group EKF and accumulate $\mathcal{N}^{(j)}\gets\mathcal{N}^{(j)}+\ell_t^{(j)}$ (Sec.~\ref{sec:methods:loss}).

    \STATE \textbf{\COMMENT{Compute loss}}


    Compute the loss $\mathcal{L}(\lambda,\psi)$ from $\{\mathcal{N}^{(j)}\}_{j\in\mathcal{B}}$, $\mathrm{KL}(q_\psi\|p)$, and annealing $\beta_k$ (Sec.~\ref{sec:methods:loss}).

    \STATE \textbf{\COMMENT{Backpropagate}}
    $g\gets\nabla_{(\lambda,\psi)}\mathcal{L}$ through EKF + Lie-IMEX; clip $\|g\|\leq c$.

    \STATE \textbf{\COMMENT{Parameter update}}
    Update every parameter group (the level $\lambda_f$, the GP weights $(m_f,s_f)$, the amplitude $\sigma_{\mathrm M,f}$ and the length scale $\ell_{\mathrm M,f}$ of each $f\in\mathcal{F}$, and $\sigma_{\mathrm{obs}}$) with its own Prodigy optimizer; all learning rates cosine-decayed (Sec.~\ref{sec:methods:loss}).

\ENDFOR

\end{algorithmic}
\end{algorithm}

\subsection{Loss Functions}
\label{sec:methods:loss}

\paragraph{Initialization:} The levels start at $M^{-1}=I$ and $D=I$, that is $\lambda_{M^{-1}}=\lambda_D=(0,0,0,0,0,0)$ in the Cholesky parameterization, and at $\lambda_V=0$, $\lambda_G=0$ and $\lambda_{\xi,i}=\log 0.1$, so the diffusion starts at $0.1$ in every channel. Each entry of the GP posterior means is drawn from $\mathcal N(0,\,1/F^2)$, except those of $G$, which start at zero. Every posterior log-standard-deviation starts at $-5$, so each weight has posterior standard deviation $e^{-5}\approx0.007$, and the kernel hyperparameters start at $\sigma_{\mathrm M}=\ell_{\mathrm M}=1$. Every level and (log) noise scale is then shifted by an independent $\mathcal N(0,s^2)$ draw, with $s=0.1$, so each seed starts from a different point. The observation noise scales start at $0.1$: $\sigma_R=\sigma_\omega=0.1$ on $SO(3)$, and $\sigma_p=\sigma_R=\sigma_v=\sigma_\omega=0.1$ on $SE(3)$.

\paragraph{Observation model.} A training window is $T$ noisy samples $y_0,\dots,y_{T-1}$ taken every $\Delta t$, with the control $u_k$ held during the interval $k-1\to k$. We write $x\oplus\delta=(R\,\mathrm{Exp}(\delta_\theta),\ \omega+\delta_\omega)$ for moving a state $x=(R,\omega)$ by a small error $\delta=(\delta_\theta,\delta_\omega)\in\mathbb R^6$, and $x_a\ominus x_b=(\mathrm{Log}(R_b^\top R_a),\ \omega_a-\omega_b)\in\mathbb R^6$ for the difference of two states \citep{sola2018micro}. Each sample is
\begin{equation}
y_k=x_k\oplus\varepsilon_k,\qquad \varepsilon_k\sim\mathcal N(0,R_{\mathrm{obs}}) 
\end{equation}
\begin{equation}
R_{\mathrm{obs}}=\mathrm{diag}(\sigma_R^2I_3,\ \sigma_\omega^2I_3),
\end{equation}
where $\sigma_R$ (rad) and $\sigma_\omega$ (rad/s) are the attitude and angular-velocity noise scales; both are learned.

\paragraph{Extended Kalman filter (EKF) marginal likelihood.} The training objective is the marginal likelihood of the noisy window, $p(y_{1:T-1}\mid y_0)$, in which the unknown true states are integrated out. By the chain rule it is the product of one-step predictive densities, $p(y_{1:T-1}\mid y_0)=\prod_{k=1}^{T-1}p(y_k\mid y_{0:k-1})$, and each factor is approximated by an extended Kalman filter; this is the prediction-error form of maximum likelihood for state-space models \citep{schweppe1965evaluation,sarkka2013bayesian}. Because the state lives on $SO(3)\times\mathbb R^3$ and the dynamics are nonlinear, we use an error-state extended Kalman filter on the Lie group \citep{bourmaud2015continuous,barrau2017invariant,sola2018micro}. The filter keeps a mean $\hat x_k$ and a covariance $P_k\in\mathbb R^{6\times6}$ of the error $x_k\ominus\hat x_k$, starting from $\hat x_0=y_0$ and $P_0=R_{\mathrm{obs}}$. For each time step $k=1,\dots,T-1$, the EKF computation has the following steps.

\emph{Predict.} Let $\Psi_h(x,u,z)$ be the Lie-IMEX step of Section~\ref{sec:methods:integrator} with standard-normal noise $z\in\mathbb R^3$ (the Brownian increment is $\Delta W=\sqrt h\,z$; $z\in\mathbb R^6$ on $SE(3)$). The predicted mean is $\hat x_k^-=\Psi_h(\hat x_{k-1},u_k,0)$. Let $A_k\in\mathbb R^{6\times6}$ be the Jacobian of the step with respect to the error state, which tells how an error in $\hat x_{k-1}$ is carried to $\hat x_k^-$, and $B_k\in\mathbb R^{6\times3}$ its Jacobian with respect to the noise, which tells how the noise $z$ enters the error state:
\begin{equation}
[A_k\ \ B_k]=\frac{\partial}{\partial(\delta,z)}\Big[\Psi_h(\hat x_{k-1}\oplus\delta,u_k,z)\ominus\hat x_k^-\Big]_{\delta=0,\,z=0}
\end{equation}
\begin{equation}
    P_k^-=A_kP_{k-1}A_k^\top+B_kB_k^\top 
\end{equation}
Both Jacobians are computed by forward-mode automatic differentiation through the integrator step, so they are exact for the discrete-time model; because $z$ is standard normal, $B_k$ already contains the factor $\sqrt h$, so $B_kB_k^\top$ is the linearized approximation to the process-noise covariance of one step; it contains the learned diffusion $\xi$.


\emph{Likelihood.}  $r_k\in\mathbb R^6$ is the difference between the sample and the prediction, and $S_k\in\mathbb R^{6\times6}$ is its covariance:
\begin{equation}
r_k=y_k\ominus\hat x_k^-,\qquad S_k=P_k^-+R_{\mathrm{obs}},
\end{equation}
\begin{equation}
\begin{aligned}
\label{eq:ekf_nll_step}
&\ell_k=-\log p(y_k\mid y_{0:k-1})\\
&=\tfrac12\big(r_k^\top S_k^{-1}r_k+\log\det S_k+n_{out}\log2\pi\big).
\end{aligned}
\end{equation}
$S_k$ is factorized by Cholesky, so no matrix inverse is formed.

\emph{Update.} With the Kalman gain $\mathcal K_k=P_k^-S_k^{-1}\in\mathbb R^{6\times6}$ and the correction $m_k=\mathcal K_kr_k\in\mathbb R^6$, whose first three entries $m_{k,\theta}$ are the attitude correction,
\begin{equation}
\hat x_k=\hat x_k^-\oplus m_k,
\end{equation}
\begin{equation}
P_k=(I-\mathcal K_k)P_k^-(I-\mathcal K_k)^\top+\mathcal K_kR_{\mathrm{obs}}\mathcal K_k^\top
\end{equation}
(Joseph form, which keeps $P_k$ symmetric positive definite \citep{bucy2005filtering}). $P_k$ was computed in the tangent space at $\hat x_k^-$; it is moved to the tangent space at the new mean by $P_k\leftarrow\Theta_kP_k\Theta_k^\top$, $\Theta_k=\mathrm{blkdiag}\big(J_r(m_{k,\theta}),I_3\big)$ \citep{bourmaud2015continuous}, where $J_r$ is the right Jacobian of $SO(3)$ \citep{sola2018micro}

\begin{equation}
J_r(\phi)=I-\frac{1-\cos\theta}{\theta^2}\,\phi^\wedge+\frac{\theta-\sin\theta}{\theta^3}\,(\phi^\wedge)^2
\end{equation}
\begin{equation}
    \theta=\|\phi\|
\end{equation}

The EKF approximation to the negative log marginal likelihood of the window is $\sum_{k=1}^{T-1}\ell_k$. A per-step likelihood of the increments $y_k \ominus y_{k-1}$ reflects contributions from both process noise and observation noise, approximately $\xi^2\Delta t+2\sigma_{\mathrm{obs}}^2$ under a locally linear, constant-diffusion approximation. The EKF models these sources separately by propagating process uncertainty through $P_k$ and incorporating observation noise independently at each measurement. Note that the statistical identifiability is not guaranteed and depends on factors such as sampling frequency, trajectory length, and model accuracy.

\paragraph{Weight KL term.} Every GP subnetwork (SO(3): $M^{-1}$, $V$, $D$, $G$ and the diffusion $\xi$; on $SE(3)$: $M_1^{-1}$, $M_2^{-1}$, $V$, $D_v$, $D_\omega$, $G$ and $\xi$) has a weight matrix
 $W\in\mathbb R^{F\times d}$ with the fixed prior $p(W)=\mathcal N(0,I)$ and the variational posterior $q(W)=\mathcal N(\mu,\mathrm{diag}(e^{2\rho}))$, where the mean $\mu$ and the log-standard-deviation $\rho$ are learned. For one subnetwork the Kullback--Leibler divergence is
\begin{equation}
\mathrm{KL}\big(q(W)\,\|\,p(W)\big)=\tfrac12\sum_{a=1}^{F}\sum_{b=1}^{d}\Big(e^{2\rho_{ab}}+\mu_{ab}^2-1-2\rho_{ab}\Big),
\end{equation}
and $\mathrm{KL}(q\,\|\,p)$ in the objective is the sum of this quantity over all subnetworks. The levels $\lambda$, the kernel hyperparameters $(\sigma_{\mathrm M},\ell_{\mathrm M})$ and the observation noise have no prior and do not enter this term.

\paragraph{Training objective.} 

Let $\theta$ collect all trainable parameters: for every GP subnetwork, the posterior means and log-standard-deviations $(\mu,\rho)$, the level $\lambda$ and the kernel hyperparameters $(\sigma_{\mathrm M},\ell_{\mathrm M})$, together with the observation noise ($(\sigma_R,\sigma_\omega)$ on $SO(3)$; $(\sigma_p,\sigma_R,\sigma_v,\sigma_\omega)$ on $SE(3)$). We minimize a variational objective for the noisy training windows surrogate of the negative evidence lower bound (ELBO) in which the likelihood is replaced by its EKF approximation,

\begin{equation}
\mathcal J_{\mathrm{EKF}}(\theta)=\mathbb E_{q(W)}\Big[\sum_{\text{windows}}\sum_{k=1}^{T-1}\ell_k\Big]+\mathrm{KL}(q\,\|\,p),    
\end{equation}

where $\ell_k$ (Equation \ref{eq:ekf_nll_step}) is the EKF-estimated negative log predictive density of sample $k$ and the expectation is over the GP weights $W$ of all subnetworks. Because overlapping windows reuse observations and restart the filter from their first sample, the sum over windows is a composite likelihood, not the likelihood of the original trajectories. This loss function is based on a negative evidence lower bound (ELBO), but since the EKF approximates the predictive densities, $-\mathcal J_{\mathrm{EKF}}$ is not guaranteed to lower-bound the log marginal likelihood of the SDE. At each training step we draw a batch of $B$ windows and one reparameterized weight sample $W=\mu+e^{\rho}\odot\epsilon$, $\epsilon\sim\mathcal N(0,I)$, shared by the whole batch and held fixed across all timesteps and integration substeps. We use the stochastic estimate
\begin{equation}
\mathcal L(\theta)=\frac{1}{n_{out}\,B\,(T-1)}\sum_{b=1}^{B}\sum_{k=1}^{T-1}\ell_k^{(b)}\;+\;\beta\,\frac{\mathrm{KL}(q\,\|\,p)}{n_{out}\,N_{\mathrm{obs}}},
\end{equation}
where $N_{\mathrm{obs}}$ is the total number of predicted observations in the training set and  $n_{out}$ is the dimension of the error state ($n_{out}=6$ on $SO(3)\times\mathbb R^3$: three for the attitude, three for the angular velocity; $n_{out}=12$ on $SE(3)\times\mathbb R^6$, with three more each for position and linear velocity), so that $\mathcal L$ is $\mathcal J_{\mathrm{EKF}}$ per observation and per dimension, with the KL term weighted by $\beta$. The weight $\beta$ rises linearly from $0$ to $1$ over the first $1000$ steps (KL annealing). All parameters are optimized with Prodigy \citep{mishchenko2024prodigy}, a learning-rate-free method, with a separate Prodigy optimizer for each parameter group: the level, the GP weights $(\mu,\rho)$, the amplitude $\sigma_{\mathrm M}$ and the length scale $\ell_{\mathrm M}$ of every subnetwork, and the observation noise. Each subnetwork therefore has its own learning rate $\eta_k(t)$, which Prodigy adapts during training. Every optimizer uses base rate $1$, initial distance estimate $d_0=10^{-6}$ and a cosine decay to $0.1$ of the base rate; gradients are clipped to global norm $1$. The model is trained for $5000$ steps on windows of $2$\,s, with batch size $B=64$ on $SO(3)$ and $B=16$ on $SE(3)$. We use $N_{\mathrm{sub}}=1$ Lie-IMEX substeps per observation interval, giving an integration timestep $h=\Delta t/N_{\mathrm{sub}}$. During stochastic simulation, each GP weight draw is held fixed throughout a rollout, while independent Brownian increments are sampled across integration substeps. The predictor and corrector within each substep reuse the same Brownian increment. During EKF training, the mean prediction uses zero noise input, and process noise is accounted for through covariance propagation. Algorithm~\ref{alg:phase1_sde_long} in Appendix~\ref{sec:appendix:algorithms} gives the complete training procedure.

\subsection{Computational complexity}
Let $F$ be the number of random Fourier features, $d$ the error-state dimension, $m$ the dimension of the Brownian motion $W_t$, $n$ the velocity/twist dimension, $B$ the minibatch size, $T$ the trajectory-window length, and $N_{\mathrm{sub}}$ the number of Lie-IMEX substeps per observation interval. 
A GP evaluation scales as $O(F)$ for fixed physical state dimension, while one Lie-IMEX substep requires $O(F+n^3)$ operations due to GP evaluation and the implicit $n\times n$ damping solve. The Lie-group EKF requires $O(d^3)$ operations for dense covariance propagation and factorization. 

Including forward-mode differentiation with respect to the $d$ state and $m$ noise directions at each substep, with Jacobians propagated substep by substep, one training iteration scales approximately as
\begin{equation}
O\left(
B(T-1)
\left[
(d+m)N_{\mathrm{sub}}(F+n^3)+d^3
\right]
\right).
\end{equation}

The total training complexity is this quantity multiplied by $N_{\mathrm{train}}$. Since $d=6$, $n=3, m=3$ for the $SO(3)$ system and $d=12$, $n=6, m=6$ for the $SE(3)$ system, the matrix dimensions remain small and computation scales primarily with the number of RFF features, trajectory samples, and integration substeps.

\section{Experiments}
\label{sec:experiments}

We evaluate LieSPHGP on three systems with different geometries, actuation,
and sources of dissipation (Figures~\ref{fig:windy_pendulum},
\ref{fig:quadrotor}, and \ref{fig:underwater_vehicle}). The pendulum and quadrotor are
simulated, allowing direct comparison of the learned energy, dissipation,
input map, and noise levels with their known values.

The \emph{windy pendulum} is a fully actuated rigid body on $SO(3)$.
It rotates about a fixed pivot under gravity, with velocity-dependent
pivot friction and random wind torques. The \emph{quadrotor} translates
and rotates on $SE(3)$. Its four rotor thrusts control six degrees of
freedom, making it underactuated. It experiences velocity-dependent air
drag and random wind forces and torques.

The \emph{BlueROV2 Heavy} is a real underwater vehicle with eight thrusters,
recorded in a test tank. Its true dynamics and noise levels are unknown.
Water produces strong velocity-dependent drag in all six degrees of
freedom and increases the vehicle's effective inertia through added mass.
Buoyancy largely balances gravity, giving this system different energy
and dissipation characteristics from the pendulum and quadrotor.
Appendix~\ref{sec:appendix:idsia} provides a fourth evaluation on real
Crazyflie flights.

\subsection{Noisy SO(3) Pendulum with Stochastic Wind}
\begin{figure}[t]
    \centering
    \includegraphics[width=\linewidth]{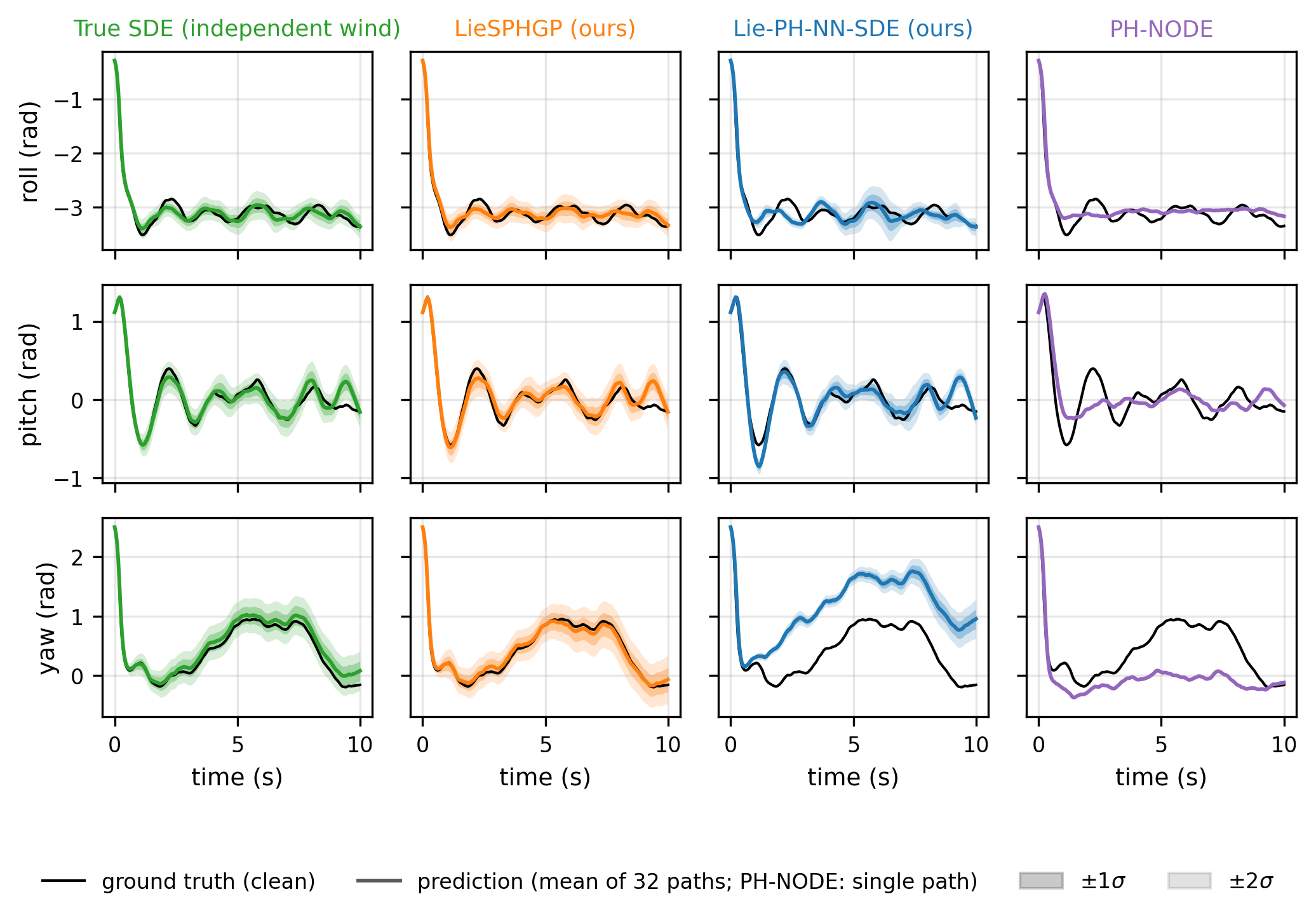}
\caption{\textbf{Open-loop prediction on the windy pendulum.}
Roll, pitch, and yaw (Euler $xyz$ angles of $R$, as in
Figure~\ref{fig:ida_pbc}) predicted over $10$\,s from the true initial
state using recorded inputs. Roll $=\pm\pi$ corresponds to the hanging
position. Black shows ground truth. Lines show the mean of $32$ sampled
paths, except for PH-NODE's single deterministic path. Shading shows
$\pm1\sigma$ and $\pm2\sigma$. The true SDE uses independently sampled
wind. We show the held-out trajectory with the largest difference in
mean rotation error between Lie-PH-NN-SDE and LieSPHGP
($43.1^\circ$ versus $7.0^\circ$). Over all $10$ held-out trajectories,
Lie-PH-NN-SDE has the larger error on $9$. On the shown trajectory, $2\sigma$
coverage is $0.97$ for both the true SDE and LieSPHGP, and $0.47$ for
Lie-PH-NN-SDE.}
    \label{fig:pend-prediction}
\end{figure}

\paragraph{SDE}
\label{sec:experiment:sde}
We consider an SDE system of a pendulum moving in 3D space with a stochastic wind force. The details are given in Appendix \ref{sec:appendix:experiment:ode}. We compare against port-Hamiltonian Neural ODE (PH-NODE) \citep{duong2024phnode}, against an unstructured neural SDE (NeuralSDE) \citep{liu2019nsde} in Euclidean coordinates, and against Lie-PH-NN-SDE, which shares the Hamiltonian structure, Lie-IMEX solver and EKF likelihood of LieSPHGP but uses neural subnetworks. The results can be seen in Table \ref{tab:3d-windy-pendulum-comparison}. We compare using metrics for rotation, velocity, and energy error, and $SO(3)$ constraints of orthogonality and determinant equal to one, and continuous ranked probability score (CRPS) \citep{hersbach2000decomposition} and $2\sigma$ coverage for both rotation and velocity. We also compare the accuracy of the learned components $M^{-1}D, M^{-1}G,$ gravity. LieSPHGP and Lie-PH-NN-SDE share the port-Hamiltonian structure, Lie-IMEX integrator, EKF likelihood, and optimizer. They differ only in function parameterization (GP versus neural networks) and in the weights (a variational posterior versus a point estimate). LieSPHGP has the lowest mean rotation, angular-velocity and energy errors and CRPS in Table~\ref{tab:3d-windy-pendulum-comparison}, the $2\sigma$ coverage closest to nominal, and the lowest identification errors ($M^{-1}G$, gravity, $M^{-1}D$ and the diffusion; Table~\ref{tab:pend-physics-id}); both Lie-IMEX models keep $SO(3)$ at round-off level, whereas NeuralSDE leaves the group (orthogonality error $\sim10^{6}$) and its $2\sigma$ bands cover at most $5\,\%$ of the truth. LieSPHGP's $2\sigma$ bands cover $90$--$93\,\%$ of the truth. Lie-PH-NN-SDE has a $62\,\%$ higher rotation error and under-covers ($61\,\%$). PH-NODE's damping is far off (relative $M^{-1}D$ error $4.96$; several times too large along the diagonal, Figure~\ref{fig:pend-physics-id}) and is $3.7\times$ less accurate. LieSPHGP identifies the control map and gravity to within $6\,\%$ (Appendix~\ref{sec:appendix:physics-id}, Table~\ref{tab:pend-physics-id}). Ablations of the training loss, the weight posterior and the window length are in Appendix~\ref{sec:appendix:pend-ablations} (Table~\ref{tab:pend-ablations}), and a comparison with SPH-NN \citep{di2026stochastic} is in Appendix~\ref{sec:appendix:sphnn}. The learned observation noise is within $2\,\%$ of the simulator's value, the learned process noise is close to it, and its $2\sigma$ coverage stays at $0.90$--$0.96$ over the $10$-s horizon (Appendix~\ref{sec:appendix:uncertainty}). Figure~\ref{fig:pend-prediction} shows the held-out trajectory with the
largest difference between Lie-PH-NN-SDE and LieSPHGP. Lie-PH-NN-SDE
drifts in yaw, and its narrow uncertainty band often excludes the true
trajectory ($2\sigma$ coverage of $0.47$). LieSPHGP tracks the true SDE
trajectory with an uncertainty band that covers it. PH-NODE's overdamped
prediction loses the oscillations after about $1$\,s.

\paragraph{ODE}
We also consider the ODE case of the system. Table~\ref{tab:3d-pendulum-comparison} trains PH-NODE and LieSPHGP (with $\xi=0$) methods on identical data in two styles. With PH-NODE's
5-point windows LieSPHGP already improves every metric, and with 2 second windows (41 points) the gap widens to an order of magnitude. Longer windows help PH-NODE only
moderately ($1.8\times$) possibly because its loss does not account for noise in the data so errors may accumulate in long rollouts. The EKF likelihood of LieSPHGP instead treats every sample as a noisy
observation and turns the longer horizon into a $16\times$ gain, recovering
$M^{-1}D$, $M^{-1}G$ and gravity to within $1$--$3\,\%$, while the Lie-IMEX integrator keeps $R$ on $SO(3)$ with $10^{-6}$ precision (Section~\ref{sec:experiments:ablations}, Table~\ref{tab:solver-ablation}).

\begin{table*}[t]
\centering
\caption{\textbf{Open-loop prediction and identification on the noisy deterministic $SO(3)$ pendulum.}
Constant friction $0.5$, observation noise $0.25$, no diffusion ($\xi=0$ for LieSPHGP). Both methods are trained on the
same data in two styles: PH-NODE's original 5-point windows (stride 1, full batch) and 41-point windows (stride 30,
batch 64). PH-NODE is the published model and loss \citep{duong2024phnode} with $M^{-1}\!\to I$ pretraining. Errors over $10$-s
rollouts of $10$ held-out trajectories, mean $\pm$ std over $5$ training seeds; best in bold. With the same windows, LieSPHGP has the lower mean error on every row.}
\label{tab:3d-pendulum-comparison}
\begin{tabular}{lcccc}
\toprule
 & \multicolumn{2}{c}{\textbf{5-point (0.2 seconds) windows}} & \multicolumn{2}{c}{\textbf{41-point (2 seconds) windows}} \\
\cmidrule(lr){2-3}\cmidrule(lr){4-5}
\textbf{Error} & PH-NODE & LieSPHGP ($\xi{=}0$) (Ours) & PH-NODE & LieSPHGP ($\xi{=}0$) (Ours) \\
\midrule
Rotation (rad)           & $1.381 \pm 0.016$ & $1.064 \pm 0.067$ & $0.773 \pm 0.083$ & $\mathbf{0.066 \pm 0.017}$ \\
Velocity (rad/s)         & $0.872 \pm 0.004$ & $0.611 \pm 0.017$ & $0.636 \pm 0.034$ & $\mathbf{0.048 \pm 0.007}$ \\
Energy                   & $1.154 \pm 0.007$ & $0.787 \pm 0.037$ & $0.714 \pm 0.033$ & $\mathbf{0.044 \pm 0.005}$ \\
Det.\ ($\times10^{-6}$)  & $66.0 \pm 0.1$    & $0.98 \pm 0.03$   & $51.7 \pm 2.8$    & $\mathbf{0.99 \pm 0.03}$   \\
Orth.\ ($\times10^{-6}$) & $84.3 \pm 0.2$    & $1.92 \pm 0.05$   & $72.3 \pm 4.4$    & $\mathbf{1.93 \pm 0.07}$   \\
\midrule
$M^{-1}D$                & $1.340 \pm 0.010$ & $0.724 \pm 0.026$ & $0.934 \pm 0.119$ & $\mathbf{0.031 \pm 0.002}$ \\
$M^{-1}G$                & $0.094 \pm 0.003$ & $0.060 \pm 0.029$ & $0.419 \pm 0.019$ & $\mathbf{0.014 \pm 0.003}$ \\
Gravity                  & $0.158 \pm 0.001$ & $0.045 \pm 0.003$ & $0.131 \pm 0.008$ & $\mathbf{0.011 \pm 0.001}$ \\
\bottomrule
\end{tabular}
\end{table*}

All models are trained for 5,000 steps with $\Delta t=0.05$s, in two data styles: 5-point windows (stride 1, all 9,700 windows per step; PH-NODE's original setting) and 41-point windows (stride 30, batch 64). PH-NODE uses the RK4 solver \citep{hairer1993solving} and $M^{-1}\!\to I$ pretraining; LieSPHGP and Lie-PH-NN-SDE use the Lie-IMEX solver (Algorithm~\ref{alg:lieimexsolver}). For the deterministic pendulum the diffusion subnetwork of LieSPHGP is set to $\xi_\theta=0$.

\begin{table}[t]
\centering
\caption{\textbf{Open-loop prediction and identification on the stochastic windy $SO(3)$ pendulum.}
Rate-dependent friction, wind $\xi=0.5$, observation noise $0.5$. Errors over $10$-s rollouts of $10$ held-out
trajectories, mean $\pm$ std over $5$ training seeds. Stochastic models are
scored on $32$ sampled paths with an independent wind draw; CRPS and $2\sigma$ coverage measure the
predictive distribution. The lower block is the relative error of the learned port-Hamiltonian products. PH-NODE is the published model in \citet{duong2024phnode}. NeuralSDE is an
unstructured neural SDE \citep{liu2019nsde} in Euclidean coordinates. LieSPHGP and Lie-PH-NN-SDE differ only in function parameterization (GP versus neural networks) and in the weights (variational posterior versus point estimate). They share the optimizer, the state-dependent diffusion and all other modeling and numerical settings. Rotation error is the geodesic angle $\arccos\big((\operatorname{tr}(R^\top\hat R)-1)/2\big)$.
For sampled models, $\hat R$ is the chordal mean, obtained by projecting
the arithmetic mean of the sampled rotation matrices onto $SO(3)$ using
SVD. For single-path models, $\hat R$ is the predicted rotation. Rotation CRPS (rad) and $2\sigma$ coverage use the tangent coordinates $\operatorname{Log}(\hat R^\top R)$ at $\hat R$ and angular-velocity CRPS is in rad/s. The $SO(3)$ constraint errors are computed on the raw, unprojected rollouts.}
\label{tab:3d-windy-pendulum-comparison}
\setlength{\tabcolsep}{3pt}
\begin{tabular}{@{}lcccc@{}}
\toprule
\textbf{Error} & \textbf{PH-NODE} & \textbf{NeuralSDE} & \textbf{Lie-PH-NN-SDE}  & \textbf{LieSPHGP}  \\
\midrule
Rotation (rad)              & $1.066 \pm 0.038$ & $1.78 \pm 0.10$   & $0.471 \pm 0.047$ & $\mathbf{0.291 \pm 0.025}$ \\
Velocity (rad/s)            & $0.776 \pm 0.007$ & $50.2 \pm 3.7$    & $0.538 \pm 0.015$ & $\mathbf{0.443 \pm 0.012}$ \\
Energy                      & $0.703 \pm 0.010$ & $(3.1 \pm 0.5)\times10^{3}$ & $0.385 \pm 0.014$ & $\mathbf{0.320 \pm 0.011}$ \\
Det.                        & $(5.2 \pm 0.2)\times10^{-4}$ & $(4.9 \pm 1.2)\times10^{8}$ & $(8.8 \pm 0.1)\times10^{-7}$ & $(9.9 \pm 0.1)\times10^{-7}$ \\
Orth.                       & $(6.1 \pm 0.2)\times10^{-4}$ & $(2.9 \pm 0.3)\times10^{6}$ & $(1.71 \pm 0.02)\times10^{-6}$ & $(1.91 \pm 0.01)\times10^{-6}$ \\
\midrule
Rotation CRPS               & $0.443 \pm 0.014$ & $0.887 \pm 0.049$ & $0.174 \pm 0.019$ & $\mathbf{0.100 \pm 0.010}$ \\
Velocity CRPS               & $0.387 \pm 0.003$ & $24.5 \pm 2.0$    & $0.192 \pm 0.008$ & $\mathbf{0.143 \pm 0.006}$ \\
Rotation $2\sigma$ coverage & --                & $0.00 \pm 0.00$   & $0.61 \pm 0.03$   & $\mathbf{0.90 \pm 0.02}$ \\
Velocity $2\sigma$ coverage & --                & $0.05 \pm 0.02$   & $0.75 \pm 0.03$   & $\mathbf{0.93 \pm 0.01}$ \\
$M^{-1}D$                   & $4.96 \pm 0.07$   & --                & $0.67 \pm 0.11$   & $\mathbf{0.44 \pm 0.11}$ \\
$M^{-1}G$                   & $0.347 \pm 0.009$ & --                & $0.392 \pm 0.024$ & $\mathbf{0.057 \pm 0.003}$ \\
Gravity                     & $0.402 \pm 0.006$ & --                & $0.285 \pm 0.040$ & $\mathbf{0.054 \pm 0.011}$ \\
\bottomrule
\end{tabular}
\end{table}

\begin{figure*}[t]
    \centering
    \includegraphics[width=0.9\linewidth]{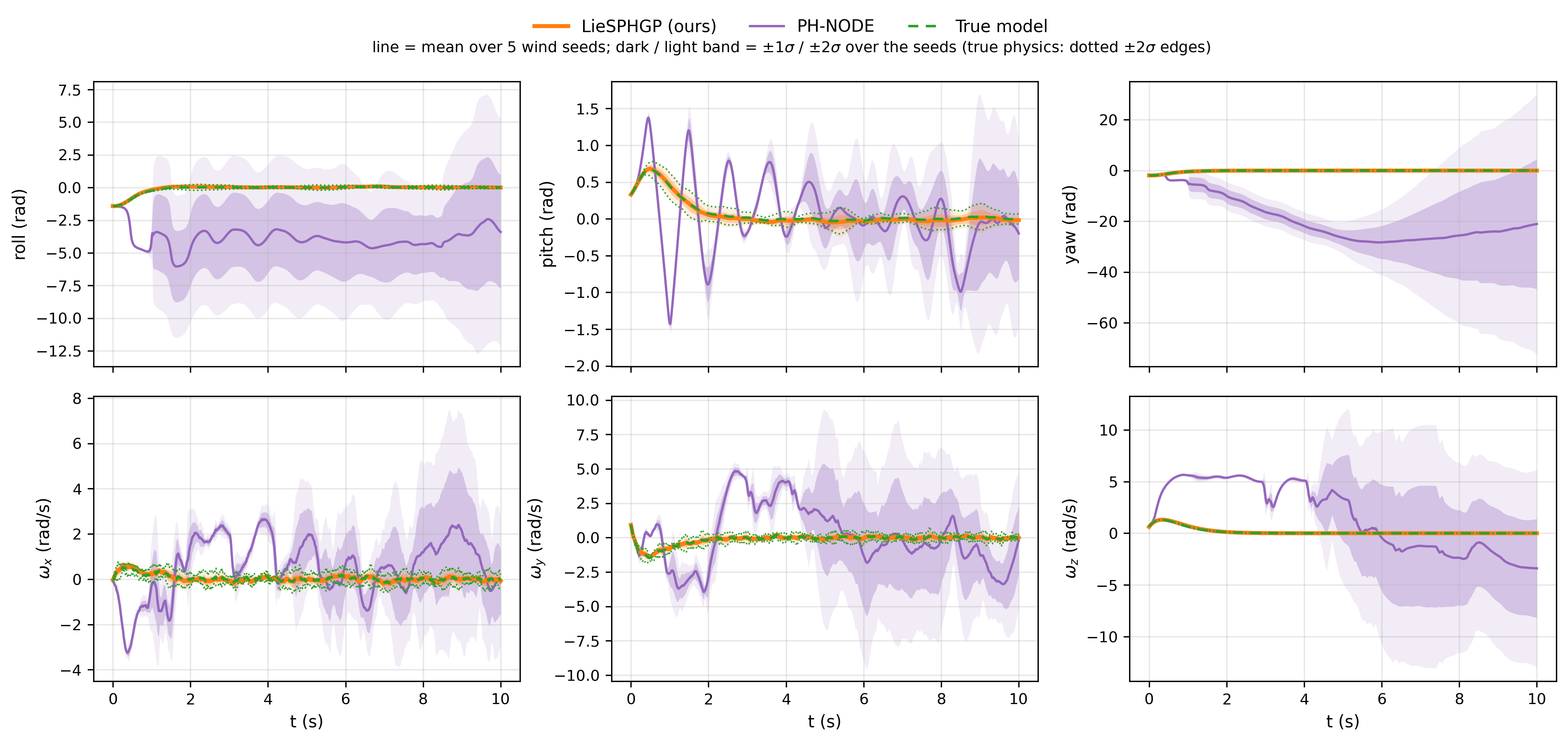}
    \caption{\textbf{Closed-loop IDA-PBC control of the stochastic $SO(3)$ pendulum using learned dynamics.}
Each model supplies its learned dynamical subnetworks to the same IDA-PBC controller acting in the windy pendulum environment, with the objective of stabilizing the pendulum at the upright configuration. LieSPHGP (orange) closely follows the controller constructed from the true model, which uses ground truth components (green), and reaches the target, whereas PH-NODE (purple) does not reach the upright. The experiment demonstrates that differences in learned physical structure that are modest in open-loop prediction can become consequential for downstream control.}
    \label{fig:ida_pbc}
\end{figure*}

\begin{figure*}[t]
    \centering
    \includegraphics[width=1\linewidth]{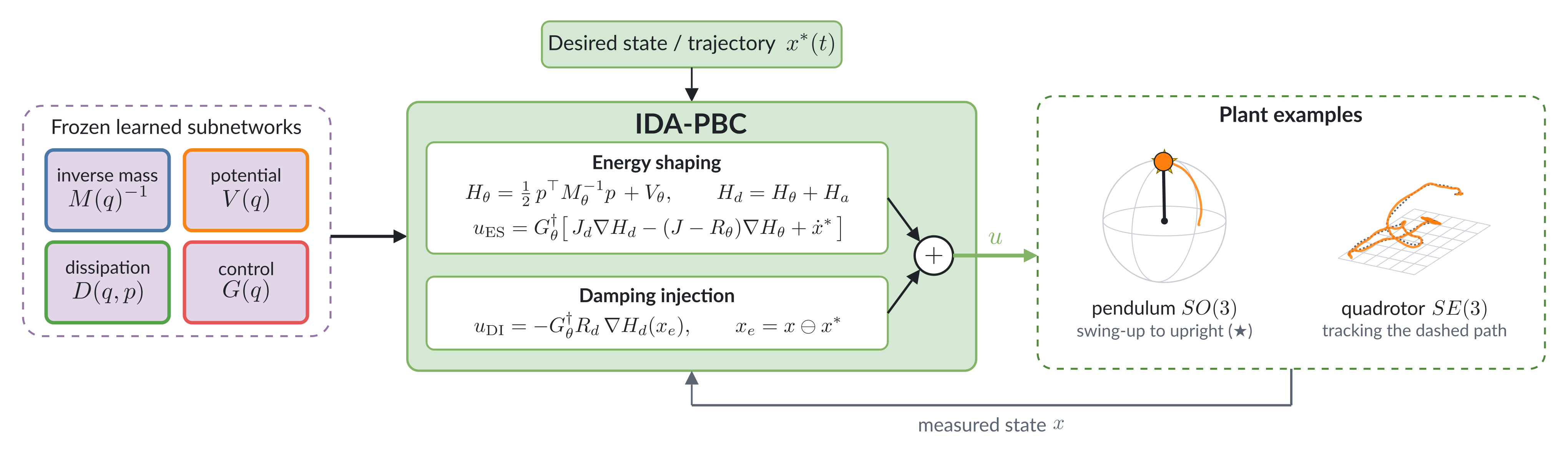}
    \caption{\textbf{Control using the learned dynamics.}
The learned subnetworks $M^{-1}$, $V$, $D$, and $G$ define the IDA-PBC
control law. Energy shaping replaces the learned potential with a desired
potential whose minimum is at the target, while damping injection adds
damping $K_\omega$. The errors are Lie-group differences,
$x_e=x\ominus x^\ast$ (Section~\ref{sec:methods:loss}), whose attitude
part is $e_R=\operatorname{Log}(R^{\ast\top}R)^\vee$ with $R^\ast$ the
target attitude. For the quadrotor,
IDA-PBC forms the inner attitude loop of an outer position-tracking
controller (Appendix~\ref{ida_pbc_SO3}).}

    \label{fig:ida-pbc-overview}
\end{figure*}

\paragraph{IDA-PBC Controller}
\label{sec:ida-pbc-pendulum}
We use each learned model inside the IDA-PBC controller \citep{ortega2002idapbc} (Figure~\ref{fig:ida-pbc-overview}) to swing the true windy pendulum
up to the upright position (Appendix~\ref{ida_pbc_SO3}), in 25 test cases with $5$ initial attitudes and $5$ Brownian motion samples. The
controller cancels the learned gravity and damping and applies gains $K_R=25s$ and $K_\omega=10\sqrt{s}$. A smaller gain scale $s$ makes it rely more on the learned model (Table~\ref{tab:ida-pbc-gain-sweep}). At normal gains ($s=1$), LieSPHGP reaches the upright in every flight and ends as close as the true model ($0.027$
vs.\ $0.024$\,rad), whereas PH-NODE succeeds in only $31\,\%$ of the flights. At low gains ($s=0.25$,
Figure~\ref{fig:ida_pbc}), LieSPHGP succeeds in $78\,\%$ of the flights, close to the true model ($84\,\%$), and PH-NODE
in none. PH-NODE's learned damping is about several times too large (Table~\ref{tab:3d-windy-pendulum-comparison}), so the
controller over-compensates it during the fast swing-up and the pendulum does not settle at the upright.

\begin{table}[t]
\centering
\caption{\textbf{IDA-PBC swing-up of the windy pendulum at normal and low controller gains.} Gains $K_R=25s$,
$K_\omega=10\sqrt{s}$. Success: angle to upright stays below $10^\circ$ ($0.175$\,rad) over the last
$2$\,s of the $10$-s run. Final angle: angle to upright at $t=10$\,s. Mean $\pm$ std over $5$ training seeds, each with
$5$ initial attitudes $\times$ $5$ Brownian motion samples. True model uses the ground-truth dynamics in the same controller.}
\label{tab:ida-pbc-gain-sweep}
\setlength{\tabcolsep}{4pt}
\begin{tabular}{@{}llccc@{}}
\toprule
 & \textbf{Gain scale} & \textbf{PH-NODE} & \textbf{LieSPHGP (Ours)} & \textbf{True model} \\
\midrule
\multirow{2}{*}{Success (\%)}
 & $s=1$    & $31 \pm 9$           & $\mathbf{100 \pm 0}$ & $100$ \\
 & $s=0.25$ & $0 \pm 0$            & $\mathbf{78 \pm 5}$  & $84$ \\
\midrule
\multirow{2}{*}{Final angle (rad)}
 & $s=1$    & $1.67 \pm 0.33$   & $\mathbf{0.027 \pm 0.001}$ & $0.024$ \\
 & $s=0.25$ & $2.06 \pm 0.16$   & $\mathbf{0.088 \pm 0.004}$ & $0.071$ \\
\bottomrule
\end{tabular}
\end{table}

\subsection{Noisy SE(3) Quadrotor with Stochastic Wind }
\label{sec:experiment:se3_quadrotor}

We next evaluate LieSPHGP on a six-degree-of-freedom quadrotor whose configuration evolves on $SE(3)$ and whose inputs are the four rotor thrusts, so that the learned input map must recover the rotor mixing matrix. We train the models on $300$ flights of $5$\,s of a Crazyflie quadrotor simulated in PyBullet with the gym-pybullet-drones model \citep{panerati2021learning}, with observation noise $0.25$, rate-dependent friction and stochastic force and torque wind ($\xi=0.5$), and evaluate on held-out $10$\,s flights. We compare LieSPHGP with PH-NODE \citep{duong2024phnode}, each trained with $3$ random seeds. LieSPHGP is trained on $2$-s windows with the EKF likelihood and PH-NODE on its original $5$-sample windows with its rollout loss, both for $5{,}000$ steps. The complete environment and port-Hamiltonian formulation are given in Appendix~\ref{sec:appendix:experiment:se3_quadrotor}.

\begin{table}[t]
\centering
\caption{\textbf{Closed-loop tracking using learned $SE(3)$ quadrotor dynamics.}
Each learned model supplies the drift and input map of the same geometric tracking controller with an IDA-PBC attitude
loop (Appendix~\ref{ida_pbc_SO3}), which commands the four rotor thrusts, flying $10$ held-out $10$-s references under
$5$ Brownian motion samples on the PyBullet plant. Values are mean $\pm$ std over $3$ training seeds. True model uses the
ground-truth components in the same controller. LieSPHGP nearly matches the true-model controller and halves PH-NODE's
tracking error. Position, linear-velocity and angular-velocity RMSE are measured against the reference and the attitude error is the RMS geodesic angle between the flown attitude and the attitude $R_d$ commanded by each model's controller. Valid tracking time is the first time the position leaves the true-model flight under the same wind sample by more than $0.158$ times the RMS distance of the reference from its starting point. Tracking CRPS compares the states reached under five wind realizations with the deterministic reference. It measures tracking performance under
stochastic disturbances and does not assess predictive calibration.}
\label{tab:quadrotor-closed-loop-comparison}
\setlength{\tabcolsep}{3pt}
\renewcommand{\arraystretch}{1.1}
\begin{tabular}{@{}lccc@{}}
\toprule
\textbf{Metric} & \textbf{PH-NODE} & \textbf{LieSPHGP} (Ours) & \textbf{True model} \\
\midrule
Position RMSE (m)             & $0.251 \pm 0.005$ & $\mathbf{0.133 \pm 0.009}$ & $0.127$ \\
Linear velocity RMSE (m/s)    & $0.856 \pm 0.015$ & $\mathbf{0.485 \pm 0.018}$ & $0.464$ \\
Attitude RMSE (rad)           & $0.220 \pm 0.006$ & $\mathbf{0.203 \pm 0.003}$ & $0.201$ \\
Angular velocity RMSE (rad/s) & $4.18 \pm 0.06$   & $\mathbf{3.93 \pm 0.04}$   & $3.89$ \\
Energy RMSE                   & $0.027 \pm 0.001$ & $\mathbf{0.016 \pm 0.000}$ & $0.016$ \\
Valid tracking time (s)       & $8.39 \pm 0.17$   & $\mathbf{9.22 \pm 0.14}$   & -- \\
Rotor saturation share        & $0.115 \pm 0.011$ & $\mathbf{0.048 \pm 0.001}$ & $0.040$ \\
Position tracking CRPS                 & $0.062 \pm 0.001$ & $\mathbf{0.027 \pm 0.003}$ & $0.025$ \\
Angular velocity tracking CRPS         & $\mathbf{1.23 \pm 0.02}$ & $1.40 \pm 0.01$     & $1.38$ \\
\bottomrule
\end{tabular}
\end{table}

\begin{figure}[t]
    \centering
    \includegraphics[width=0.9\linewidth]{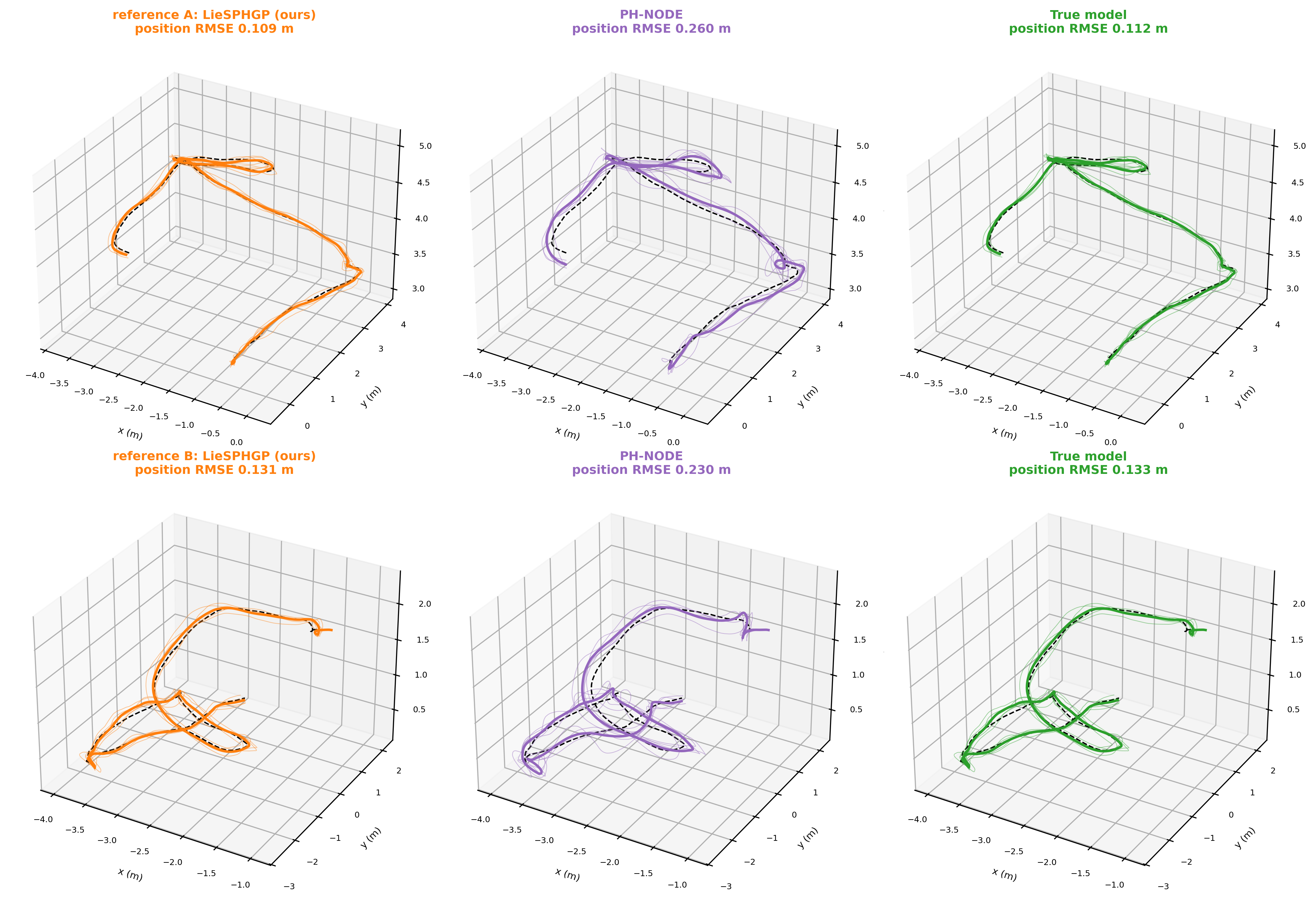}
    \caption{\textbf{Closed-loop $SE(3)$ quadrotor tracking using learned dynamics.}
    Three-dimensional trajectories for two of the ten held-out $10$-s flights
(A and B). Thin lines show individual Brownian motion samples, thick
lines show the mean over those five samples, and dashed lines show the
reference. Each model supplies the dynamics to the same geometric tracking
controller with an IDA-PBC attitude loop. The controller's commands are
allocated to the four rotors of the contact-free PyBullet CF2P plant~\citep{panerati2021learning}. LieSPHGP (orange) tracks close to the controller using the true dynamics
(green), while PH-NODE (purple) deviates during turns. Across all flights
and seeds, the position RMSE is $0.133$\,m for LieSPHGP and $0.251$\,m
for PH-NODE (Table~\ref{tab:quadrotor-closed-loop-comparison}).
    }
    \label{fig:se3_quadrotor_heldout}
\end{figure}

We use the same geometric tracking controller for all models. Each model
supplies its learned drift and input map, which the controller uses to
compute the four rotor thrust commands. The attitude loop uses IDA-PBC
(Appendix~\ref{ida_pbc_SO3}). We evaluate tracking on the PyBullet plant
using $10$ held-out reference flights and $5$ Brownian wind realizations
(Table~\ref{tab:quadrotor-closed-loop-comparison},
Figure~\ref{fig:se3_quadrotor_heldout}).

LieSPHGP achieves a position RMSE of $0.133$\,m, close to the $0.127$\,m
obtained with the true dynamics. Its position error, velocity error, and
rotor saturation are each roughly half those of PH-NODE. PH-NODE
overestimates translational damping ($M_1^{-1}D_v$ error of $8.0$,
Table~\ref{tab:quad-physics-id}), which biases the feedforward thrust.

To evaluate closed-loop prediction, we replace the PyBullet plant with
each learned model. LieSPHGP has a position error of $0.121$\,m, about
$17\,\%$ higher than the true model's $0.103$\,m. PH-NODE has a position
error of $0.431$\,m (Table~\ref{tab:quadrotor-model-in-loop}). The effect of the training window length on both methods is in Appendix~\ref{sec:appendix:quad-windows} (Table~\ref{tab:quad-window-ablation}).

\begin{table}[t]
\centering
\caption{\textbf{Model-in-the-loop prediction on the quadrotor.} The controller of
Table~\ref{tab:quadrotor-closed-loop-comparison}, built from the true dynamics, computes the rotor thrusts from the
learned model's own predicted state, and the learned model (with its learned diffusion, $8$ paths) is propagated as the
plant; the prediction is compared with the same controller flying the real PyBullet plant ($10$ held-out references
$\times$ $5$ Brownian motion samples). Errors averaged over $0$--$10$\,s; mean $\pm$ std over $3$ training seeds. True
model: the ground-truth SDE as the plant with an independent Brownian motion sample.}
\label{tab:quadrotor-model-in-loop}
\setlength{\tabcolsep}{3pt}
\begin{tabular}{@{}lccc@{}}
\toprule
\textbf{Metric} & \textbf{PH-NODE} & \textbf{LieSPHGP} (Ours) & \textbf{True model} \\
\midrule
Position error (m)              & $0.431 \pm 0.009$ & $\mathbf{0.121 \pm 0.011}$ & $0.103$ \\
Attitude error (rad)            & $0.313 \pm 0.004$ & $\mathbf{0.174 \pm 0.008}$ & $0.159$ \\
Angular velocity error (rad/s)  & $1.995 \pm 0.026$ & $\mathbf{1.165 \pm 0.039}$ & $1.040$ \\
Position CRPS (m)               & $0.199 \pm 0.006$ & $\mathbf{0.043 \pm 0.003}$ & $0.037$ \\
Position $2\sigma$ coverage     & $0.00$            & $\mathbf{0.66 \pm 0.02}$   & $0.69$ \\
Thrust difference (/ hover)     & $0.360 \pm 0.019$ & $\mathbf{0.192 \pm 0.004}$ & $0.175$ \\
Torque difference (mN\,m)       & $1.79 \pm 0.02$   & $\mathbf{1.16 \pm 0.02}$   & $1.09$ \\
\bottomrule
\end{tabular}
\end{table}

\subsection{Real World Underwater Vehicle Dataset}
\begin{figure}[t]
    \centering
    \includegraphics[width=0.7\linewidth]{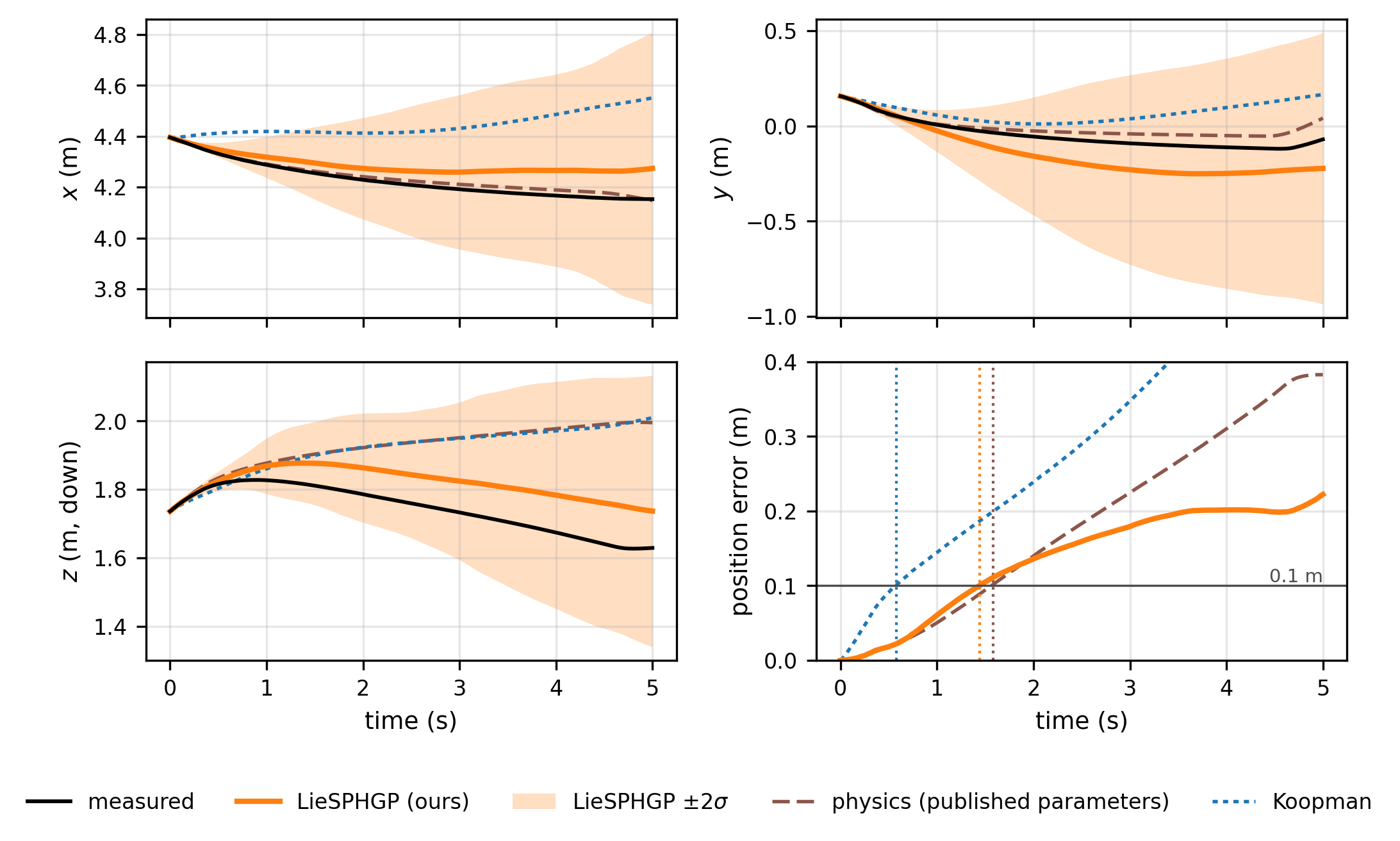}
\caption{\textbf{Open-loop prediction on a held-out BlueROV2 test segment.}
Predicted positions $x$, $y$, and $z$ (positive downward), and position
error, using recorded thruster commands and the measured initial state.
Vertical dotted lines mark the first time position error exceeds
$0.1$\,m (valid prediction time). LieSPHGP shows the mean of $32$ sampled
paths with a $\pm2\sigma$ band (training seed $0$). The physics model uses
published vehicle parameters. LieSPHGP learns $M^{-1}$, $V$, $D$, and $G$
from training data alone. The shown segment has the median LieSPHGP valid
prediction time among the $15$ test segments.}
    \label{fig:rov-prediction}
\end{figure}

We evaluate LieSPHGP on real BlueROV2 Heavy recordings from the Marinarium
dataset \citep{torroba2026marinarium}, using the benchmark's chronological
train--test split and multi-step prediction protocol. The thruster commands
were issued by a pilot responding to the vehicle's motion. We use these
recorded commands to predict up to $2$\,s ahead from each test sample
(Table~\ref{tab:rov-real}). The data processing, training details, and the physics and Koopman baselines are described in Appendix~\ref{app:underwater}. After one step, all models have errors comparable to the measurement noise.
LieSPHGP has the lowest error at $0.2$\,s. At $2$\,s, its RMSE is $16\,\%$
lower than that of the physics model, which is given the vehicle's published mass, damping and added mass, restoring forces and thruster map, whereas LieSPHGP learns $M^{-1}$, $V$, $D$ and $G$ from the recorded trajectories alone,
and comparable to that of the Koopman model
\citep{torroba2026marinarium} trained on the same data. LieSPHGP also has
the longest valid prediction time, with position error remaining below
$0.1$\,m for $1.98$\,s on average, compared with $1.62$\,s for the physics
model and $0.81$\,s for the Koopman model. Figure~\ref{fig:rov-prediction} shows the test segment with the median LieSPHGP valid prediction time. LieSPHGP and the physics model leave the $0.1$\,m band at about the same time ($1.4$ and $1.6$\,s); afterwards the error of LieSPHGP grows more slowly, mainly because it predicts the depth better. On unseen trajectories from a recording flown in attitude-stabilized mode, without retraining, LieSPHGP's error at $2$\,s is half that of the Koopman model and $39\,\%$ lower than that of the physics model, and its position error over $3$\,s is the lowest (Appendix~\ref{app:underwater}, Table~\ref{tab:rov-transfer}); persistence is competitive there because the vehicle hardly rotates. On real Crazyflie flights from the benchmark of \citet{busetto2026nonlinear}, LieSPHGP has the lowest position and velocity prediction errors (Appendix~\ref{sec:appendix:idsia}, Table~\ref{tab:multistep-prediction}).

\begin{table}[t]
\centering
\caption{\textbf{Prediction on real BlueROV2 trajectories.}
Endpoint RMSE across the 12 state variables (position, Euler angles,
body linear and angular velocities) after $T$ steps of $0.02$\,s. As in the benchmark, the RMSE pools all 12 components with their native units (m, rad, m/s, rad/s). Predictions start from each of the $9{,}164$ test samples in the Marinarium recording \citep{torroba2026marinarium} and use the recorded
thruster commands. Lower RMSE is better. Valid prediction time (VPT)
is the time until the position error first exceeds $0.1$\,m, averaged
over $15$ held-out $10$-s segments. Higher VPT is better.
Koopman (EDMDc) uses the code and hyperparameters of
\citet{torroba2026marinarium} and the same training trajectories as
LieSPHGP. Physics uses the published BlueROV2 parameters.
LieSPHGP uses the rollout with posterior-mean GP weights (no process noise) for endpoint RMSE and the
mean of $32$ sampled paths for VPT. Its results are reported as
mean $\pm$ standard deviation over two training seeds.}
\label{tab:rov-real}
\setlength{\tabcolsep}{4pt}
\begin{tabular}{@{}lcccc@{}}
\toprule
\textbf{Model} & $T=1$ (0.02 s) & $T=10$ (0.2 s) & $T=100$ (2 s) & \textbf{VPT (s)} \\
\midrule
Persistence                       & $0.091$          & $0.122$                    & $0.402$                    & $0.94$ \\
Physics (published parameters)    & $\mathbf{0.086}$ & $0.128$                    & $0.301$                    & $1.62$ \\
Koopman (EDMDc)                   & $0.096$          & $0.126$                    & $0.259$                    & $0.81$ \\
\midrule
LieSPHGP (ours)                   & $0.090 \pm 0.000$ & $\mathbf{0.105 \pm 0.002}$ & $\mathbf{0.252 \pm 0.034}$ & $\mathbf{1.98 \pm 0.28}$ \\
\bottomrule
\end{tabular}
\end{table}

\subsection{Ablations: Rotation Kernel and Integrator}
\label{sec:experiments:ablations}
We isolate two design choices with matched data, feature count ($F=64$), seed and training budget ($5{,}000$ steps, final checkpoint).

\paragraph{Comparison of embedded and intrinsic kernels on $SO(3)$.}
We replace the embedded Mat\'ern features of every GP with a rotational input by random features of the intrinsic Mat\'ern kernel of $SO(3)$ \citep{borovitskiy2020matern,azangulov2024compact}, and by its product with a Euclidean Mat\'ern kernel for inputs $(R,x)$; everything else is unchanged. Table~\ref{tab:kernel-ablation} shows no consistent gain on the pendulum where every metric agrees within the trajectory-to-trajectory spread, while training takes $1.6\times$ longer with the intrinsic kernel. Both kernels depend only on the relative rotation $R^\top R'$ and are bi-invariant on $SO(3)$, and each intrinsic feature costs a degree-8 Chebyshev recurrence instead of one cosine (Appendix~\ref{sec:appendix:kernel}).

\paragraph{Comparison of Lie-IMEX and RK4 integrators.}
To separate integrator effects from model effects, we roll out every trained pendulum model with both Lie-IMEX and classical RK4 \citep{hairer1993solving} at the same step $h=0.05$\,s, without retraining. The integrator does not change accuracy (rotation error $0.089$ vs.\ $0.084$ for LieSPHGP ($\xi=0$), seed $0$), but it alone determines the $SO(3)$ constraint error (Table~\ref{tab:solver-ablation}). Lie-IMEX keeps every model at the float32 round-off level ($\approx10^{-6}$), whereas RK4 leaves the group by $3\times10^{-5}$ to $4\times10^{-4}$ within $10$\,s. The constraint rows of Tables~\ref{tab:3d-pendulum-comparison} and~\ref{tab:3d-windy-pendulum-comparison} therefore reflect the integrator, and their accuracy rows reflect the learned model.

\begin{table}[t]
\centering
\caption{\textbf{Embedded vs.\ intrinsic $SO(3)$ kernel.} Same data, $F=64$ features, training seed 0, learned kernel hyperparameters and $5{,}000$ training steps; only the features of the GPs with rotational input change. Pendulum errors are $0$--$10$\,s open-loop values and the quadrotor position RMSE is the closed-loop tracking error, as in the main tables (mean $\pm$ std over held-out trajectories). Damping error: relative error of $M^{-1}D$ along the ground-truth closed-loop states (protocol of Tables~\ref{tab:pend-physics-id} and~\ref{tab:quad-physics-id}).}
\label{tab:kernel-ablation}
\setlength{\tabcolsep}{4pt}
\begin{tabular}{@{}llcc@{}}
\toprule
\textbf{System} & \textbf{Metric} & \textbf{Embedded} & \textbf{Intrinsic} \\
\midrule
Pendulum & Test EKF NLL & $0.7875$ & $0.7875$ \\
 & Rotation error (rad) & $0.259 \pm 0.13$ & $0.266 \pm 0.13$ \\
 & Rotation CRPS & $0.091 \pm 0.038$ & $0.093 \pm 0.040$ \\
 & Damping error & $0.861$ & $0.880$ \\
 & Training time (min) & $15.8$ & $25.6$ \\
\midrule
Quadrotor & Test EKF NLL & $0.1137$ & $0.1137$ \\
 & Position RMSE, closed loop (m) & $0.134 \pm 0.038$ & $0.134 \pm 0.038$ \\
 & Damping error $D_v$ / $D_\omega$ & $0.244$ / $0.173$ & $0.251$ / $0.171$ \\
 & Training time (min) & $56.0$ & $78.1$ \\
\bottomrule
\end{tabular}
\end{table}

\begin{table}[t]
\centering
\caption{\textbf{$SO(3)$ constraint error of the same trained models under two integrators.} Each trained pendulum model is rolled out over $10$\,s with Lie-IMEX and with RK4 at the same step $h=0.05$\,s (no retraining). Values: mean $\pm$ std over the $10$ held-out trajectories of $\max_t|\det R_t-1|$ and $\max_t\|R_t^\top R_t-I\|_F$, computed in float64 from float32 rollouts. RK4 rollouts carry no diffusion.}
\label{tab:solver-ablation}
\setlength{\tabcolsep}{3pt}
\begin{tabular}{@{}lcccc@{}}
\toprule
 & \multicolumn{2}{c}{\textbf{Determinant}} & \multicolumn{2}{c}{\textbf{Orthogonality}} \\
\cmidrule(lr){2-3}\cmidrule(lr){4-5}
\textbf{Model} & Lie-IMEX & RK4 & Lie-IMEX & RK4 \\
\midrule
\multicolumn{5}{@{}l}{\emph{Deterministic pendulum}} \\
LieSPHGP ($\xi=0$) & $(9.8\pm0.5)\!\times\!10^{-7}$ & $(3.3\pm1.7)\!\times\!10^{-5}$ & $(1.9\pm0.08)\!\times\!10^{-6}$ & $(5.1\pm2.8)\!\times\!10^{-5}$ \\
PH-NODE & $(9.6\pm3.8)\!\times\!10^{-7}$ & $(6.1\pm1.3)\!\times\!10^{-5}$ & $(2.1\pm0.3)\!\times\!10^{-6}$ & $(7.8\pm1.1)\!\times\!10^{-5}$ \\
\midrule
\multicolumn{5}{@{}l}{\emph{Windy pendulum}} \\
LieSPHGP & $(9.3\pm0.8)\!\times\!10^{-7}$ & $(8.9\pm0.4)\!\times\!10^{-5}$ & $(1.8\pm0.1)\!\times\!10^{-6}$ & $(1.0\pm0.04)\!\times\!10^{-4}$ \\
Lie-PH-NN-SDE & $(9.8\pm0.5)\!\times\!10^{-7}$ & $(1.1\pm0.2)\!\times\!10^{-4}$ & $(1.9\pm0.06)\!\times\!10^{-6}$ & $(1.3\pm0.2)\!\times\!10^{-4}$ \\
PH-NODE & $(1.1\pm0.4)\!\times\!10^{-6}$ & $(3.4\pm1.8)\!\times\!10^{-4}$ & $(1.9\pm0.5)\!\times\!10^{-6}$ & $(4.1\pm2.2)\!\times\!10^{-4}$ \\
\bottomrule
\end{tabular}
\end{table}

Additional results are given in the Appendix. The code and data are available at \url{https://github.com/CVC-Lab/LieSPHGP}. The above experiments are done on NVIDIA A100 GPU with 40GB VRAM. We use JAX \citep{jax2018github} and Optax \citep{deepmind2020jax} in our implementation.

\FloatBarrier

\section{Conclusion}
\label{sec:conclusion}
We introduced LieSPHGP, a framework that combines (i) port-Hamiltonian structure, (ii) Lie group geometry, (iii) stochastic dynamics, and (iv) Gaussian-process modeling to learn from noisy trajectories. The dynamics are formulated as a SDE on $SO(3) \times \mathfrak{so}(3)^*$ or $SE(3) \times \mathfrak{se}(3)^*$ and integrated with Lie-IMEX, which combines exponential-map rotation updates with linearly implicit treatment of learned dissipation. Across the pendulum and quadrotor experiments, the method improves
prediction accuracy, uncertainty calibration, and geometric consistency
under noisy conditions. The EKF likelihood and longer training windows
contribute to these gains, which also extend to closed-loop tracking
with the learned dynamics. On real underwater-vehicle recordings, LieSPHGP predicts as accurately as or better than physics-based and data-driven baselines at short horizons and has the longest valid prediction time. Beyond prediction, the learned energy, dissipation, and input components support the construction of IDA-PBC controllers without additional controller training or repeated data collection. Our simulated closed-loop results demonstrate this capability, motivating future deployment of controllers learned from offline trajectory datasets on physical systems.
The current formulation and evaluation focus on single rigid-body dynamics on SO(3) and SE(3), including a single-body pendulum, rather than general multi-joint articulated systems. The main model uses extrinsic Mat\'ern kernels on
embedded coordinates, with intrinsic kernels evaluated as an ablation. Inference relies on an approximate EKF likelihood, so the variational objective is not a certified lower bound on the exact SDE evidence. The numerical guarantees concern rotational group preservation and frozen-damping contractivity; they do not establish stability or a general stochastic convergence order for the complete
nonlinear scheme. Likewise, the closed-loop results
provide empirical evidence of control performance
rather than a stochastic stability guarantee. Extending the method to multiple connected bodies requires accounting for inter-body dynamic coupling and joint constraints in the configuration space. This remains an important direction for future work. Another direction for future work is extending to metriplectic dynamics \citep{morrison2024metriplectic}, which separate energy and entropy terms. The present results establish a foundation for extending geometry-preserving, data-driven dynamics learning and control to more complex articulated systems.

\subsection*{Acknowledgements}
This research was supported in part  from the Peter O’Donnell Foundation, the Jim Holland-Backcountry Foundation and in part from a grant from the Army Research Office accomplished under Cooperative Agreement Number W911NF-19-2-0333.

\subsection*{AI use statement}

In this work, we used generative AI tools to implement methods, clean and reformat datasets, support qualitative and thematic data analysis, source and search for information, identify relevant literature, help design or provide feedback on research methodology and experiments, interpret results, help formulate mathematical claims, and assist in the writing of proofs.
We have not used generative AI tools to generate synthetic data sets, help develop theoretical models or conceptual frameworks, assist with translation, or propose or refine hypotheses.
We have reviewed all AI-assisted work. We verified all LLM-suggested references against their original sources. We take responsibility for the final content of this work, including text, claims, proofs, or artifacts produced with the aid of generative AI.

\bibliography{main}
\bibliographystyle{plainnat}

\clearpage
\appendix
\thispagestyle{empty}
\section{Additional Background}

\subsection{Lie-Poisson Systems}
A Lie group $G$ is a smooth manifold equipped with smooth group operations \citep{varadarajan1984lie}. Many physical systems evolve on Lie groups (e.g., rotations on $\mathrm{SO}(3)$, rigid body motions on $\mathrm{SE}(3)$). The associated Lie algebra $\mathfrak{g}=T_eG$ is the tangent space of Lie group at the identity. It is a vector space equipped with a Lie bracket $[\cdot,\cdot]$ satisfying bilinearity, antisymmetry, and the Jacobi identity:
\begin{equation}
    [A,[B,C]] + [B,[C,A]] + [C,[A,B]] = 0.
\end{equation}
The phase space is the cotangent bundle $T^*G$. Using left trivialization, we write $T^*G \simeq G\times\mathfrak{g}^*$, so each state is represented by $(q,p)$, where $q\in G$ is the configuration and $p\in\mathfrak{g}^*$ is the body-frame momentum. Here $\mathfrak{g}^*$ is the dual of the Lie algebra \citep{marsden1999symmetry}.
The Hamiltonian $H \in C^\infty(T^*G)$ is a function on this space. The Poisson bracket is defined by:
\begin{equation}
    \{f,g\}(q,p) = \nabla f(q,p)^\top J(q,p)\, \nabla g(q,p),
\end{equation}
where $J(q,p)$ is a skew-symmetric structure matrix that also satisfies the Jacobi identity, encoding the underlying geometry \citep{duong2024phnode} and for the rigid-body Lie-Poisson structure used here, the Jacobi identity follows from the Lie bracket. Dissipative effects can be added via a dissipation bracket \citep{morrison1986diss}:
\begin{equation}
    (f,g)(q,p) = \nabla f(q,p)^\top D(q,p)\, \nabla g(q,p),
\end{equation}
where $D$ is positive semi-definite. With an entropy $S$ and the degeneracy conditions $J\nabla S=0$ and $D\nabla H=0$, combining the Poisson and dissipative brackets yields metriplectic dynamics, which conserve $H$ and produce entropy:
\begin{equation}
    \frac{d}{dt}f(q,p) = \{f,H\} + (f,S),
\end{equation}
where $S$ is an entropy functional. Alternatively, port-Hamiltonian dynamics \citep{van2014port} of the form,
\begin{equation}
    \begin{bmatrix}
        \dot{q} \\ \dot{p}
    \end{bmatrix}
    = (J(q,p) - \mathcal{R}(q,p)) \nabla H(q,p) + \mathcal{G}(q,p)u,
\end{equation}
capture conservative, dissipative, and external forcing/control effects in a unified framework.

Our mechanical model instead dissipates energy through $-D\nabla H$ and has external stochastic forcing. It is not a closed thermodynamic model.

\subsection{Port-Hamiltonian System on the SO(3) Manifold}
\label{sec:appendix:background:so3}
Given a rotation matrix $R \in SO(3)$ with rows $r_1,r_2,r_3$:
\begin{equation}
    q =R = \begin{bmatrix} r_{1}^T \\ r_{2}^T \\ r_{3}^T \end{bmatrix} = \begin{bmatrix} r_{1x} & r_{1y} & r_{1z} \\ r_{2x} & r_{2y} & r_{2z} \\ r_{3x} & r_{3y} & r_{3z} \end{bmatrix}
\end{equation}

and angular velocity as $\omega = [\omega_x,\omega_y,\omega_z]$

A port-Hamiltonian system has the form:
\begin{equation}
    \begin{bmatrix} \dot{q} \\ \dot{p} \end{bmatrix} = ({J}(q, p) - \mathcal{R}(q, p)) \begin{bmatrix} \nabla_q \mathcal{H} \\ \nabla_p \mathcal{H} \end{bmatrix} + \mathcal{G}(q) u
\end{equation}

The interconnection matrix ($J$) captures the energy-conserving internal structure (kinematics and Coriolis effects). It is skew-symmetric ($J = -J^\top$).
\begin{equation}
    {J}(q, p) = \begin{bmatrix} 0_{9 \times 9} & q^\times_{9\times3} \\ -(q^\times)^\top_{3 \times 9} & p^\times_{3\times3} \end{bmatrix}_{12\times12}
\end{equation}

where $q^\times$ maps body-frame angular velocity to the
time derivatives of the rotation rows, stacked as column
vectors. It is formed by stacking their skew-symmetric
matrices without transposing:
\begin{equation}
q^\times
=
\begin{bmatrix}
\hat r_1\\
\hat r_2\\
\hat r_3
\end{bmatrix}
=
\begin{bmatrix}
0 & -r_{1z} & r_{1y}\\
r_{1z} & 0 & -r_{1x}\\
-r_{1y} & r_{1x} & 0\\
\hline
0 & -r_{2z} & r_{2y}\\
r_{2z} & 0 & -r_{2x}\\
-r_{2y} & r_{2x} & 0\\
\hline
0 & -r_{3z} & r_{3y}\\
r_{3z} & 0 & -r_{3x}\\
-r_{3y} & r_{3x} & 0
\end{bmatrix}.
\label{eq:rotation_matrix}
\end{equation}

(Notation: $\hat{a}$ is the $3 \times 3$ skew-symmetric matrix such that $\hat{a}b = a \times b$. Note that $\hat{r}_i^\top = -\hat{r}_i$.) \\

$p^\times$ implements the cross product for the momentum dynamics.
\begin{equation}    
    p^\times = \hat{p} = \begin{bmatrix} 0 & -p_z & p_y \\ p_z & 0 & -p_x \\ -p_y & p_x & 0 \end{bmatrix}
\end{equation}

where $p_x,p_y,p_z$ is momentum around X,Y,Z axis respectively. These are calculated from $p = M(q)\omega$ where $M(q)$ is the mass matrix, which is symmetric positive definite ($M\succ0$).

The dissipation matrix $R$ captures energy loss (friction/damping).
\begin{equation}
    \mathcal{R}(q, p) = \begin{bmatrix} 0_{9 \times 9} & 0_{9 \times 3} \\ 0_{3 \times 9} & D(q, p)_{3\times3} \end{bmatrix}_{12\times12}
\end{equation}
where $D(q,p)$ is symmetric positive semi-definite (PSD) matrix.

$G(q)u$ represents external work. The input matrix $G$ maps your control inputs $u$ (motors/thrusters) into forces that add energy to the system
\begin{equation}
    \mathcal{G}(q)u = \begin{bmatrix} 0_{9\times3} \\ G(q)_{3\times3} \end{bmatrix} u
\end{equation}

Hamiltonian energy $\mathcal H$ is determined by 
\begin{equation}
    \mathcal H(q,p) = \frac 1 2 p^TM^{-1}(q) p+V(q)
\end{equation} where $V(q)$ is potential energy, $M^{-1}(q)$ is inverse mass and p is momentum $p=[p_x,p_y,p_z]$.

The port-Hamiltonian system equation now is
\begin{equation}
\begin{split}
\begin{bmatrix} \dot{q} \\ \dot{p} \end{bmatrix} &= 
\left( 
\begin{bmatrix} 0_{9\times9} & q^\times_{9\times3} \\ -(q^\times)^\top_{3\times9} & p^\times_{3\times3} \end{bmatrix} - 
\begin{bmatrix} 0_{9\times9} & 0_{9\times3} \\ 0_{3\times9} & D(q,p)_{3\times3} \end{bmatrix} 
\right) \\
&\quad \times \begin{bmatrix} \nabla_q \mathcal{H}_{(9\times1)} \\ \nabla_p \mathcal{H}_{(3\times1)} \end{bmatrix} + 
\begin{bmatrix} 0_{9\times3} \\ G(q)_{3\times3} \end{bmatrix} u
\end{split}
\end{equation}
From above, the rate of change of rotation and momentum is calculated as: 
\begin{equation}
\begin{aligned}
    \dot q &= ( q^\times )  \nabla_p H, \\
    \dot p &=  -(q^\times)^T \nabla_q H + p^\times \nabla_p H - D(q,p)\nabla_p H + G(q)u.
\end{aligned}
\end{equation}

The above is the ODE form. To model the effect of stochastic forces, we consider an SDE form. A diffusion term $M(q)\,\xi(q)\circ dW_t$ is added to the momentum dynamics, where $\xi(q)$ is the diffusion of the angular velocity $\omega=M^{-1}(q)\,p$, and the resulting SDE is:
\begin{equation}
\begin{aligned}
    dq &= ( q^\times )  \nabla_p H \, dt, \\
    dp &=  \left(-(q^\times)^T \nabla_q H + p^\times \nabla_p H - D(q,p)\nabla_p H + G(q)u \right) \, dt + M(q)\,\xi(q)\circ dW_t.
\end{aligned}
\end{equation}

The angular-velocity drift follows from $\omega = M^{-1}(q)\,p$ (Equation~\ref{eq:momentum_to_velocity}):
\begin{equation}
    \dot{\omega} = M(q)^{-1} \dot{p} + \left( \frac{d}{dt} M(q)^{-1} \right) p ,
\end{equation}
where $\dot p$ is the drift of the momentum equation above and the diffusion $\xi(q)$ acts on $\omega$ directly. This is the form the integrator advances; the second term is nonzero because the learned $M^{-1}(q)$ depends on the configuration.

\subsection{Port-Hamiltonian System on the SE(3) Manifold}
\label{sec:appendix:background:se3}
The pose is $(x, R) \in SE(3)$, where $x = [x_1, x_2, x_3]^\top \in \mathbb{R}^3$ is the position in the world frame and $R \in SO(3)$ is the rotation matrix with rows $r_1, r_2, r_3$. The generalized coordinate stacks both:
\begin{equation}
    q = \begin{bmatrix} x \\ r_1 \\ r_2 \\ r_3 \end{bmatrix}_{12 \times 1}, \qquad
    R = \begin{bmatrix} r_{1}^\top \\ r_{2}^\top \\ r_{3}^\top \end{bmatrix} = \begin{bmatrix} r_{1x} & r_{1y} & r_{1z} \\ r_{2x} & r_{2y} & r_{2z} \\ r_{3x} & r_{3y} & r_{3z} \end{bmatrix}
\end{equation}

The body-frame twist is $[v, \omega]$, with linear velocity $v = [v_x, v_y, v_z]$ and angular velocity $\omega = [\omega_x, \omega_y, \omega_z]$.

A port-Hamiltonian system has the form:
\begin{equation}
    \begin{bmatrix} \dot{q} \\ \dot{p} \end{bmatrix} = ({J}(q, p) - \mathcal{R}(q, p)) \begin{bmatrix} \nabla_q \mathcal{H} \\ \nabla_p \mathcal{H} \end{bmatrix} + \mathcal{G}(q) u
\end{equation}

The interconnection matrix ($J$) captures the energy-conserving internal structure (kinematics and Coriolis effects). It is skew-symmetric ($J = -J^\top$).
\begin{equation}
    {J}(q, p) = \begin{bmatrix} 0_{12 \times 12} & q^\times_{12\times6} \\ -(q^\times)^\top_{6 \times 12} & p^\times_{6\times6} \end{bmatrix}_{18\times18}
\end{equation}
where $q^\times$ acts as the kinematic map. It is block diagonal: the rotation $R$ maps the body linear velocity to the world frame, and $R^\times$ is built by stacking the skew-symmetric matrix of each row vector $r_i$ vertically

\begin{equation}
q^\times
=
\begin{bmatrix}
R & 0_{3\times3}\\
0_{9\times3} & R^\times
\end{bmatrix},
\qquad
R^\times
=
\begin{bmatrix}
\hat r_1\\
\hat r_2\\
\hat r_3
\end{bmatrix}.
\label{eq:se3_kinematic_map}
\end{equation}

Here $\hat a b=a\times b$, and $r_i$ denotes the column
vector whose transpose is row $i$ of $R$. Thus
$\dot r_i=\hat r_i\omega=r_i\times\omega$, consistently
with $\dot R=R\hat\omega$.

$p^\times$ implements the cross products for the momentum dynamics. With linear momentum $p_v = [p_{vx}, p_{vy}, p_{vz}]$ and angular momentum $p_\omega = [p_{\omega x}, p_{\omega y}, p_{\omega z}]$,
\begin{equation}
    p^\times = \begin{bmatrix} 0_{3\times3} & \hat{p}_v \\ \hat{p}_v & \hat{p}_\omega \end{bmatrix}_{6\times6}, \qquad
    \hat{p}_v = \begin{bmatrix} 0 & -p_{vz} & p_{vy} \\ p_{vz} & 0 & -p_{vx} \\ -p_{vy} & p_{vx} & 0 \end{bmatrix}, \quad
    \hat{p}_\omega = \begin{bmatrix} 0 & -p_{\omega z} & p_{\omega y} \\ p_{\omega z} & 0 & -p_{\omega x} \\ -p_{\omega y} & p_{\omega x} & 0 \end{bmatrix}
\end{equation}
These are calculated from $p_v = M_1(x)\, v$ and $p_\omega = M_2(R)\, \omega$, where $M_1(x)$ is the mass and $M_2(R)$ is the inertia, both symmetric positive definite (SPD) matrices.

The dissipation matrix $\mathcal{R}$ captures energy loss (drag/damping).
\begin{equation}
    \mathcal{R}(q, p) = \begin{bmatrix} 0_{12 \times 12} & 0_{12 \times 3} & 0_{12 \times 3} \\ 0_{3 \times 12} & D_v(q,p)_{3\times3} & 0_{3\times3} \\ 0_{3 \times 12} & 0_{3\times3} & D_\omega(q,p)_{3\times3} \end{bmatrix}_{18\times18}
\end{equation}
where $D_v(q,p)$ and $D_\omega(q,p)$ are symmetric positive definite (SPD) matrices.

$G(q)u$ represents external work. The input matrix $G$ maps the control inputs $u = [F_1, F_2, F_3, F_4]$ (the thrusts of the four rotors) into a body wrench that adds energy to the system
\begin{equation}
    \mathcal{G}(q)u = \begin{bmatrix} 0_{12\times4} \\ G(q)_{6\times4} \end{bmatrix} u, \qquad
    G(q) = \begin{bmatrix} G_v(q)_{3\times4} \\ G_\omega(q)_{3\times4} \end{bmatrix}
\end{equation}

Hamiltonian energy $\mathcal H$ is determined by
\begin{equation}
\label{eq:se3-hamiltonian}
    \mathcal H(q,p) = \frac 1 2 p_v^\top M_1^{-1}(x)\, p_v + \frac 1 2 p_\omega^\top M_2^{-1}(R)\, p_\omega + V(x)
\end{equation}
where $V(x)$ is potential energy, $M_1^{-1}(x)$ and $M_2^{-1}(R)$ are the inverse mass and inverse inertia, and $p = [p_v, p_\omega]$ is the momentum.

The port-Hamiltonian system equation now is
\begin{equation}
\begin{split}
\begin{bmatrix} \dot{q} \\ \dot{p} \end{bmatrix} &=
\left(
\begin{bmatrix} 0_{12\times12} & q^\times_{12\times6} \\ -(q^\times)^\top_{6\times12} & p^\times_{6\times6} \end{bmatrix} -
\begin{bmatrix} 0_{12\times12} & 0_{12\times6} \\ 0_{6\times12} & \mathrm{diag}\big(D_v(q,p), D_\omega(q,p)\big)_{6\times6} \end{bmatrix}
\right) \\
&\quad \times \begin{bmatrix} \nabla_q \mathcal{H}_{(12\times1)} \\ \nabla_p \mathcal{H}_{(6\times1)} \end{bmatrix} +
\begin{bmatrix} 0_{12\times4} \\ G(q)_{6\times4} \end{bmatrix} u
\end{split}
\end{equation}
with $\nabla_q \mathcal H = [\nabla_x \mathcal H, \nabla_{r_1} \mathcal H, \nabla_{r_2} \mathcal H, \nabla_{r_3} \mathcal H]$ and $\nabla_p \mathcal H = [\nabla_{p_v} \mathcal H, \nabla_{p_\omega} \mathcal H] = [v, \omega]$. From above, the rate of change of pose and momentum is calculated as:
\begin{equation}
\begin{aligned}
    \dot x &= R\, \nabla_{p_v} \mathcal H, \\
    \dot r_i &= r_i \times \nabla_{p_\omega} \mathcal H, \quad i = 1,2,3, \\
    \dot p_v &= p_v \times \nabla_{p_\omega} \mathcal H - R^\top \nabla_x \mathcal H - D_v(q,p) \nabla_{p_v} \mathcal H + G_v(q) u, \\
    \dot p_\omega &= p_\omega \times \nabla_{p_\omega} \mathcal H + p_v \times \nabla_{p_v} \mathcal H + \sum_{i=1}^{3} r_i \times \nabla_{r_i} \mathcal H - D_\omega(q,p) \nabla_{p_\omega} \mathcal H + G_\omega(q) u.
\end{aligned}
\end{equation}

The above is the ODE form. To model the effect of stochastic forces (wind), we consider an SDE form. A diffusion term is added to the momentum dynamics: $M_1(x)\,\xi_v(q)\circ dW_t^{v}$ to the linear momentum and $M_2(R)\,\xi_\omega(q)\circ dW_t^{\omega}$ to the angular momentum and $\xi=(\xi_v,\xi_\omega)$ is the diffusion of the body twist $(v_b,\omega_b)=(M_1^{-1}(x)\,p_v,\;M_2^{-1}(R)\,p_\omega)$ (Equation~\ref{eq:momentum_to_velocity}). The resulting SDE is:

\begin{equation}
\begin{aligned}
    dx &= R\, \nabla_{p_v} \mathcal H \, dt, \\
    dr_i &= r_i \times \nabla_{p_\omega} \mathcal H \, dt, \\
    dp_v &= \left( p_v \times \nabla_{p_\omega} \mathcal H - R^\top \nabla_x \mathcal H - D_v(q,p) \nabla_{p_v} \mathcal H + G_v(q) u \right) dt + M_1(x)\,\xi_v(q)\circ dW_t^{v}, \\
    dp_\omega &= \left( p_\omega \times \nabla_{p_\omega} \mathcal H + p_v \times \nabla_{p_v} \mathcal H + \sum_{i=1}^{3} r_i \times \nabla_{r_i} \mathcal H - D_\omega(q,p) \nabla_{p_\omega} \mathcal H + G_\omega(q) u \right) dt + M_2(R)\,\xi_\omega(q)\circ dW_t^{\omega},
\end{aligned}
\end{equation}
where $W_t = [W_t^v, W_t^\omega]$ is a 6-dimensional Wiener process.

\subsection{IDA-PBC Controller}
\label{ida_pbc_SO3}
Interconnection and damping assignment passivity-based control (IDA-PBC) \citep{ortega2002idapbc, duong2024phnode} is a structure-preserving method for stabilizing or tracking with port-Hamiltonian systems. Rather than designing an arbitrary feedback law, it constructs a control input so that the closed-loop system again has port-Hamiltonian form, but with a desired energy function and desired interconnection and damping structure.

Consider a port-Hamiltonian system with state $x \in \mathbb{R}^{n}$,
\begin{equation}
\dot{x}
=
\big(J(x) - \mathcal{R}(x)\big)\nabla H(x) + \mathcal{G}(x) u,
\end{equation}
where $J(x) = -J^\top(x)$, $\mathcal{R}(x) = \mathcal{R}^\top(x) \succeq 0$, and $H(x)$ is the Hamiltonian. IDA-PBC seeks a desired Hamiltonian
\begin{equation}
H_d(x) = H(x) + H_a(x),
\end{equation}
together with desired interconnection and damping matrices $J_d(x)$ and $\mathcal{R}_d(x)$ such that the closed-loop system satisfies
\begin{equation}
\dot{x}
=
\big(J_d(x) - \mathcal{R}_d(x)\big)\nabla H_d(x).
\end{equation}

The function $H_a$ is chosen so that $H_d$ has a strict
local minimum at the desired equilibrium $x^\ast$.
Under exact deterministic matching, skew-symmetric $J_d$,
and symmetric positive-semidefinite $\mathcal{R}_d$, the shaped
energy satisfies
\[
\dot H_d=-\nabla H_d^\top \mathcal{R}_d\nabla H_d\leq0.
\]

For tracking, let $e=\psi(x,t)\in\mathbb R^{m}$ be a local error coordinate that is smooth on the region of interest, e.g. $e=x-x^\ast(t)$ for Euclidean components and $e_R=\operatorname{Log}\!\big(R_d(t)^\top R\big)^\vee$ for an attitude. With $f(x)=(J-R)\nabla H$ the open-loop drift,
\begin{equation}
\dot e = \frac{\partial \psi}{\partial x}\,\big[f(x)+\mathcal{G}(x)u\big]+\frac{\partial \psi}{\partial t} .
\end{equation}
Matching with the desired error dynamics $\dot e=v_d(e):=(J_d-\mathcal{R}_d)\nabla_e H_d(e)$ gives the linear matching equation
\begin{equation}
A\,u=b,\qquad
A=\frac{\partial \psi}{\partial x}\,G,\qquad
b=v_d(e)-\frac{\partial \psi}{\partial x}\,f-\frac{\partial \psi}{\partial t} ,
\label{eq:tracking_matching}
\end{equation}
solved in the minimum-norm sense by $u=A^\dagger b$. The desired dynamics are realized exactly iff
\begin{equation}
(I-AA^\dagger)\,b=0 .
\label{eq:tracking_feasibility}
\end{equation}
For $e=x-x^\ast(t)$, $A=\mathcal{G}$ and $b=v_d-f+\dot x^\ast$, where $\dot x^\ast$ is the known reference derivative; the $J_d$ and $\mathcal{R}_d$ parts of $v_d$ give the energy-shaping and damping-injection terms $u_{\mathrm{ES}}+u_{\mathrm{DI}}$.
Using the Moore--Penrose pseudoinverse $A^\dagger$,
we write the controller as
\begin{equation}
u=u_{\mathrm{ES}}+u_{\mathrm{DI}},
\end{equation}
where
\begin{equation}
\begin{aligned}
u_{\mathrm{ES}}
&=
A^\dagger
\left[
J_d(e)\nabla_e H_d(e)
-\frac{\partial\psi}{\partial x}\,f(x)
-\frac{\partial\psi}{\partial t}
\right],\\
u_{\mathrm{DI}}
&=
-A^\dagger \mathcal{R}_d(e)\nabla_e H_d(e),
\qquad
A=\frac{\partial\psi}{\partial x}\,\mathcal{G}(x).
\end{aligned}
\end{equation}
Their sum is $u=A^\dagger b$, the minimum-norm
least-squares solution of the matching equation.
Under exact matching, a strict local minimum of $H_d$ at $e=0$, $J_d=-J_d^\top$ and $\mathcal{R}_d=\mathcal{R}_d^\top\succeq0$ yield $\dot H_d=-\nabla_eH_d^\top \mathcal{R}_d\nabla_eH_d\le0$, which bounds the error but does not by itself imply convergence.
The controllers are implemented in configuration - twist coordinates $s=(q,\zeta)$, $\zeta=M_\theta^{-1}(q)\,p$. The matching equations use the full-state drift $f_s=(f_q,f_\zeta)$ and input map
\[
G_s(s)=\begin{bmatrix}0\\ M_\theta^{-1}(q)\,G_\theta(q)\end{bmatrix},
\]
where $f_q$ is the configuration kinematics and $f_\zeta$ the learned zero-input twist drift, including the mass-derivative term of Equation~\ref{eq:momentum_to_velocity}. Both $f_\zeta$ and $M_\theta^{-1}G_\theta$ are invariant under the gauge $(M^{-1},V,D,G)\to(cM^{-1},V/c,D/c,G/c)$, so the controller does not depend on the learned gauge.

The preceding energy-shaping argument concerns the
deterministic dynamics under exact matching. With
stochastic forcing and learned-model errors, it does not by itself guarantee stochastic passivity or asymptotic convergence. We evaluate closed-loop
performance empirically and do not claim a stochastic stability guarantee.

To describe the effects of noise and model error, write
the actual closed-loop dynamics in smooth local error
coordinates $e$ in It\^o form as
\begin{equation}
de=\big[(J_d-\mathcal R_d)\nabla_e H_d+r\big]\,dt
+\Xi\,dW_t.
\end{equation}
Here $r$ is the drift mismatch, including model errors,
actuator limitations, and any It\^o correction required
by the chosen coordinates. The matrix $\Xi$ describes
the actual plant diffusion in these coordinates.
For a twice differentiable, time-independent $H_d$,
with $J_d=-J_d^\top$ and
$\mathcal R_d=\mathcal R_d^\top\succeq0$,
the It\^o generator $\mathcal A$ gives the local expected
rate of energy change:
\begin{equation}
\begin{aligned}
\mathcal A H_d
={}&-\nabla_e H_d^\top\mathcal R_d\nabla_e H_d
+\nabla_e H_d^\top r+\frac12\operatorname{tr}
\big(\Xi\Xi^\top\nabla_e^2H_d\big).
\end{aligned}
\end{equation}
The damping term is nonpositive, while the remaining
terms need not be.
For the pendulum law below, with a fixed reference,
unchanged configuration kinematics, and within the
smooth attitude-error domain,
$H_d=\tfrac12\|\omega\|^2+V_d$ gives
\begin{equation}
\mathcal A H_d
=-K_\omega\|\omega\|^2
+\omega^\top r_\omega
+\frac12\|\Xi_\omega(q)\|_F^2,
\end{equation}
where $r_\omega$ is the angular-velocity drift mismatch
and $\Xi_\omega$ is the actual plant's angular-velocity
diffusion. Thus, even with $r_\omega=0$, diffusion
contributes a nonnegative term that can offset damping.
This balance alone does not establish convergence to
zero energy or to a stationary energy level.
We do not bound these terms for the reported experiments.

\paragraph{Equilibrium Matching feasibility.}
For underactuated systems, the desired closed-loop dynamics
can be realized exactly only if the matching condition
\begin{equation}
\left(I-\mathcal{G}(x)\mathcal{G}^\dagger(x)\right)
\left[
(J_d-\mathcal{R}_d)\nabla H_d-(J-\mathcal{R})\nabla H
\right]=0
\end{equation}
is satisfied, where all quantities are evaluated at the
same state $x$. This condition requires the desired
closed-loop vector field to differ from the open-loop drift
only along the actuated directions \citep{ortega2002idapbc}.
Otherwise, the pseudoinverse-based controller realizes only
the projection of the desired correction onto the range of
$\mathcal{G}(x)$, and exact IDA-PBC matching is not guaranteed.

\paragraph{$SO(3)$ pendulum example.} For the windy pendulum, we consider equilibrium regulation
on $SO(3)\times\mathbb R^3$ with state $(q,p)$ and target
$(q_d,0)$. We choose the desired potential
\begin{equation}
V_d(q)=\frac{K_R}{2}\,\big\|\operatorname{Log}(q_d^\top q)^\vee\big\|^2 ,
\end{equation}

with gradient torque $K_R\,e_R$, $e_R=\operatorname{Log}(q_d^\top q)^\vee$, defined for rotation angles $\theta<\pi$ where the swing-up starts are the recorded initial states of five held-out test trajectories, none at $\theta=\pi$. 
The principal logarithm is smooth for rotation angles below $\pi$.
Our implementation remains accurate to within about $10^{-6}$\,rad of
$\pi$. In the reported swing-up trials, LieSPHGP and the true model remain
at or below their initial angle of $2.64$\,rad. Only controllers that
have already failed approach the $\pi$ limit in these trials.
We do not use the Frobenius potential $K(3-\operatorname{tr}(q_d^\top q))$ because it gives the torque $2K\,\tfrac12(q_d^\top q-q^\top q_d)^\vee$ of magnitude $2K\sin\theta$, which vanishes at $\theta=\pi$.
The desired energy replaces both the learned kinetic and potential terms, $H_d(q,\omega)=\tfrac12\|\omega\|^2+V_d(q)$, i.e.\ $H_a=H_d-H_\theta$, with desired damping $\mathcal{R}_d=K_\omega I$. Along the target dynamics $\dot\omega=-K_Re_R-K_\omega\omega$ one has $\dot H_d=-K_\omega\|\omega\|^2$.

For $K_R>0$, $V_d$ has a strict minimum at $q=q_d$.

With $e_R=\operatorname{Log}(q_d^\top q)^\vee$ and target dynamics $\dot\omega=-K_Re_R-K_\omega\omega$, Equation~\ref{eq:tracking_matching} in twist coordinates gives
\begin{equation}
u=\big(M_\theta^{-1}G_\theta\big)^{\dagger}\big[-K_Re_R-K_\omega\omega-f_\zeta(q,\omega)\big],
\end{equation}
where $f_\zeta$ is the learned twist drift at $u=0$, so the learned gravity, Coriolis and damping terms are cancelled and the damping $-K_\omega\omega$ is injected. If the learned input map $M_\theta^{-1}G_\theta$ is nonsingular, the desired angular acceleration is matched exactly until the commands saturate; the clip $|u_i|\le25$ can prevent exact matching. We use $K_R=25s$ and $K_\omega=10\sqrt{s}$ with gain scale $s\in\{1,0.25\}$ (Table~\ref{tab:ida-pbc-gain-sweep}). With model errors or stochastic forcing this is a nominal shaping law, not a convergence guarantee.

\paragraph{$SE(3)$ quadrotor.} With four inputs for six twist directions, Equation~\ref{eq:tracking_feasibility} fails for a generic position error, and a full-state IDA-PBC law would require solving matching partial differential equations for the two unactuated horizontal directions. We therefore use the standard geometric tracking cascade \citep{lee2010geometric}, with the learned model supplying the drift and the input map, and apply IDA-PBC only to the fully actuated attitude subsystem.
The outer loop is a PD law with feed-forward, $a_w=a_d-K_p(x-x_d)-K_v(v_w-v_d)$. Subtracting the learned translational drift gives the required thrust acceleration $a_T=R\big(R^\top a_w-\omega\times v_b-f_v\big)$, whose direction, together with the yaw reference, defines the desired attitude $R_d$.
The inner loop applies Equation~\ref{eq:tracking_matching} with $e_R=\tfrac12(R_d^\top R-R^\top R_d)^\vee$, $e_\omega=\omega-R^\top R_d\omega_d$ and target dynamics $\dot\omega=-K_Re_R-K_\omega e_\omega$, where $\omega_d$ is obtained by finite differences of $R_d$ and the feed-forward term $\partial_t\psi$ is neglected. The rotor thrusts $u$ solve the four actuated rows (body-$z$ and angular accelerations) of $M_\theta^{-1}G_\theta u=b$ in the least-squares sense. For a constant $R_d$, $e_R$ is the gradient of $\Psi(R)=\tfrac12\operatorname{tr}(I-R_d^\top R)$, so the attitude loop is IDA-PBC with full cancellation: $H_d=\tfrac12\lVert e_\omega\rVert^2+K_R\Psi(R)$, damping $K_\omega I$ and $\dot H_d=-K_\omega\lVert e_\omega\rVert^2$. For a moving $R_d$ this holds only approximately, and the full cascade has no single $H_d$ because the thrust acts along $Re_3$ rather than $R_de_3$. We therefore use it as a nominal design, without a stability guarantee for the learned model. Gains, in acceleration units: $K_p=(10,10,50)$, $K_v=3$, $K_R=250$, $K_\omega=20$, at 200~Hz. Negative thrusts are set to zero and each rotor speed is limited to $\Omega_{\max}$ and the rotor saturation share in Table~\ref{tab:quadrotor-closed-loop-comparison} is the fraction of control steps in which at least one rotor reaches this limit.

\section{Computational Details}
\label{sec:appendix:computation}

\subsection{Symmetric Positive-Definite Gaussian Process Subnetworks}
\label{sec:appendix:computation:psd_gp}
The inverse mass matrix and the dissipation matrix are symmetric positive definite. Each is produced in four steps:
\begin{enumerate}
    \item \textbf{Feature mapping.} The input $z$ ($z=\mathrm{vec}(R)\in\mathbb R^9$ for $M^{-1}$, $z=\omega\in\mathbb R^3$ for $D$) is mapped to Mat\'ern random Fourier features $\mathbf h=\phi(z)\in\mathbb R^{F}$ (Section~\ref{sec:methods:gp}).
    \item \textbf{Weight sampling.} The GP weights $W\in\mathbb R^{F\times 6}$ have the variational posterior $q(W)=\mathcal N(\mu,\mathrm{diag}(\sigma^2))$ and the prior $\mathcal N(0,I)$. One sample $W=\mu+\sigma\odot\epsilon$, $\epsilon\sim\mathcal N(0,I)$, is drawn by the reparameterization trick and added to a trainable level $\lambda\in\mathbb R^6$:
    \begin{equation}
        a=\lambda+W^\top\mathbf h\in\mathbb R^6 .
    \end{equation}
    \item \textbf{Cholesky factor.} The first three entries of $a$ give the diagonal of $L$ through the exponential and the last three fill the strictly lower triangle:
    \begin{equation}
        L(a)=\begin{bmatrix} e^{a_1} & 0 & 0\\ a_4 & e^{a_2} & 0\\ a_5 & a_6 & e^{a_3}\end{bmatrix}.
    \end{equation}
    \item \textbf{Output.} The matrix is $L(a)L(a)^\top$. Mathematically, the Cholesky decomposition of $L$ guarantees positive definiteness without an added regularizer
\end{enumerate}

The entries of $a$ are Gaussian processes; the matrix $L(a)L(a)^\top$ is a nonlinear transformation of them and is not itself a Gaussian process.

For the $SE(3)$ quadrotor, the rotational inverse mass $M_2^{-1}(R)$ and both dissipation matrices $D_v(v)$, $D_\omega(\omega)$ use this construction, while the translational inverse mass is a positive scalar times the identity, $M_1^{-1}(x)=\exp\!\big(\lambda_1+\phi(x)^\top w\big)\,I_3$.

\subsection{Geometric Preservation and Frozen-Damping Stability of Lie-IMEX}
\label{sec:appendix:stability}

We prove the geometric preservation and frozen-damping stability properties stated in Section~\ref{sec:methods:integrator}. The result concerns preservation of $SO(3)$ by the exponential-map update and stability of the trapezoidal corrector for a frozen dissipative subsystem. It does not constitute a general nonlinear or stochastic stability proof for the complete Lie-IMEX scheme.

\paragraph{Preservation of $SO(3)$.}
The rotational corrector has the form
\begin{equation}
R_{n+1}
=
R_n
\exp\left(
\frac{h}{2}
\widehat{\omega_n+\omega_p}
\right).
\end{equation}
Define
\begin{equation}
\Delta\theta_n
=
\frac{h}{2}(\omega_n+\omega_p), \qquad
R_{n+1}
=
R_n\exp(\widehat{\Delta\theta_n}),
\end{equation}
where $\widehat{\Delta\theta_n}\in\mathfrak{so}(3)$ is skew-symmetric. By definition of the exponential map, $\exp(\widehat{\Delta\theta_n})
\in SO(3).$ Since $SO(3)$ is closed under matrix multiplication, $R_n\in SO(3)$ implies $R_{n+1} = R_n\exp(\widehat{\Delta\theta_n})
\in SO(3)$. Thus, the rotational Lie-IMEX update preserves the $SO(3)$ constraint in exact arithmetic. Numerically we use Rodrigues' rotation formula \citep{rodrigues1840lois, lynch2017modern} to compute the exponential map.

\paragraph{Frozen scalar damping stability.}
Consider the scalar dissipative test equation
\begin{equation}
\dot{y}
=
-\mu y,
\qquad
\mu\geq 0.
\end{equation}
Applying the trapezoidal damping update gives
\begin{equation}
y_{n+1}
=
y_n
-
\frac{h\mu}{2}
\left(
y_n+y_{n+1}
\right).
\end{equation}
Rearranging gives
\begin{equation}
\left(
1+\frac{h\mu}{2}
\right)y_{n+1}
=
\left(
1-\frac{h\mu}{2}
\right)y_n.
\end{equation}
Let $a=h\mu$. Then the amplification factor is
\begin{equation}
\rho(a)
=
\frac{1-a/2}{1+a/2}.
\end{equation}
For $a\geq0$,
\begin{equation}
\left|1-\frac{a}{2}\right|
\leq
1+\frac{a}{2},
\end{equation}
and consequently
\begin{equation}
|\rho(a)|
=
\frac{
\left|1-a/2\right|
}{
1+a/2
}
\leq 1.
\end{equation}
Thus, the frozen scalar damping update is stable for every $h>0$ and $\mu\geq0$.

For comparison, explicit Heun applied to the same equation has amplification factor
\begin{equation}
\rho_{\mathrm{Heun}}(a)
=
1-a+\frac{a^2}{2},
\end{equation}
which is stable along the negative real axis only for $0\leq a\leq2$. The implicit damping corrector therefore removes this timestep restriction for the frozen scalar dissipative subsystem.

\paragraph{Frozen matrix damping stability and contractivity.}
Consider the frozen dissipative subsystem
\begin{equation}
M\dot z=-Dz,
\qquad
M=M^\top\succ0,
\qquad
D=D^\top\succeq0,
\end{equation}
where both matrices are held fixed over the timestep.

The trapezoidal damping update is
\begin{equation}
\left(M+\frac{h}{2}D\right)z_{n+1}
=
\left(M-\frac{h}{2}D\right)z_n.
\end{equation}
Since $M+\frac{h}{2}D$ is positive definite, this
update is well defined for every $h\geq0$.

Define the mass-weighted norm and kinetic energy by
\begin{equation}
\|z\|_M^2=z^\top Mz,
\qquad
E(z)=\frac12\|z\|_M^2.
\end{equation}
Rearranging the update gives
\begin{equation}
M(z_{n+1}-z_n)
=
-\frac{h}{2}D(z_{n+1}+z_n).
\end{equation}
Multiplying from the left by
$(z_{n+1}+z_n)^\top$ and using the symmetry of $M$
yields
\begin{equation}
\begin{aligned}
\|z_{n+1}\|_M^2-\|z_n\|_M^2
&=
-\frac{h}{2}(z_{n+1}+z_n)^\top
D(z_{n+1}+z_n)\\
&\leq0.
\end{aligned}
\end{equation}
Consequently,
\begin{equation}
E(z_{n+1})-E(z_n)
=
-\frac{h}{4}(z_{n+1}+z_n)^\top
D(z_{n+1}+z_n)
\leq0.
\end{equation}
Thus, the frozen trapezoidal damping update is
contractive in the mass-weighted norm for every
$h\geq0$. This statement uses the same frozen
mass matrix at both endpoints of the step.

\paragraph{Scope of the result.}
The preceding stability result applies to the dissipative subsystem with its coefficients frozen over a timestep. In the complete LieSPHGP dynamics, the mass matrix, dissipation, Hamiltonian forces, control forces, and stochastic diffusion may depend on the evolving state and are reevaluated during the predictor--corrector step. The result therefore does not establish unconditional stability of the complete nonlinear stochastic Lie-IMEX method. Similarly, preservation of $R_n\in SO(3)$ establishes manifold membership but does not imply symplecticity or preservation of a stationary stochastic measure. The result instead establishes the two properties used to motivate Lie-IMEX in this work: preservation of the rotational configuration manifold and stable implicit treatment of frozen positive-definite damping.
The frozen trapezoidal damping step is A-stable but not L-stable: its amplification factor $(1-h\mu/2)/(1+h\mu/2)$ tends to $-1$ as $h\mu\to\infty$, where $\mu>0$ is a damping rate. Thus, very stiff modes can alternate in sign and decay slowly numerically, even though they remain bounded. We do not establish a general strong or weak stochastic convergence order for the complete Lie-IMEX scheme.

\section{Additional Experiment Details}

\subsection{3D Windy Rigid-Body Pendulum (SO(3) Lie Group)}
\label{sec:appendix:experiment:ode}

\subsubsection{Environment Description}
The windy pendulum is a controlled rigid-body pendulum: a rigid-body model of mass $m$, whose center of mass
is a distance $l$ from the fixed pivot along the body's $e_3$ axis, with configuration the full attitude $R\in SO(3)$,  body angular rate $\omega\in\mathbb R^3$, isotropic inertia $I=ml^2I_3$, and three independent body torques $u\in\mathbb R^3$ applied through $G$. Unlike the minimal spherical pendulum on $S^2$, which tracks only the bob direction $Re_3$, the rotation about the rod axis is a dynamic degree of freedom here: gravity and wind exert no torque about it, but the control and the damping act on it.
We simulate a realistic sensor by adding noise directly to the underlying physical states. To ensure the noisy observations remain physically valid, we apply geometrically-correct noise to the rotation matrix using the exponential map on the $SO(3)$ manifold, and standard Gaussian noise to the 3D angular velocity vector.

\subsubsection{State Space}
Following the Lie group formulation, we represent the pendulum's 3D orientation as a full rotation matrix in $SO(3)$:
\[
q = R \in SO(3)
\]
The corresponding angular velocity is $ \omega = [\omega_x, \omega_y, \omega_z]^\top \in \mathbb{R}^3$. Our model's state observation consists of the flattened 9-dimensional rotation matrix $R_{\text{flat}}$ and the 3-dimensional angular velocity vector $\omega$, resulting in a 12-dimensional state space.

\subsubsection{System Dynamics}
The underlying dynamics are governed by Euler's rigid body equations. The complete equations of motion for the pendulum's orientation $q_t$ and angular velocity $\omega_t$ in the body frame are:
\begin{align}
    dq_t &= q_t \hat{\omega}_t dt \\
    d\omega_t &= I^{-1} \left( \tau_g + \tau_{\text{wind}} + G u - \tau_{\text{fric}} - \omega_t \times (I \omega_t) \right) dt + I^{-1} \tau_{\text{stoch}} \circ dW_t
\end{align}
where $I = ml^2 \mathbb{I}_{3 \times 3}$ is the inertia tensor, $u \in \mathbb{R}^3$ is the control input, applied in the body frame through the control map matrix G, $\tau_g = q_t^\top (l q_t e_3 \times -mg e_3)$ is the gravity torque, and $dW_t$ is the stochastic Wiener increment. We use $m=1$, $l=1$, and $g=9.81$.

\subsubsection{Port-Hamiltonian Dynamics Formulation}
\begin{equation}
\begin{aligned}
    q &= R \in SO(3), \\
    p &= M(q)\, \omega, \quad \omega \in \mathbb{R}^3,\\
    H(q, p) &= \frac{1}{2}\, p^\top M^{-1}(q) p + V(q) \\
    F &= G(q) u\\
    dq &= q \left[ \frac{\partial H}{\partial p} \right]_{\times} dt \\
    dp &= \{\underbrace{p \times \frac{\partial H}{\partial p}}_{\text{Coriolis}} + \underbrace{\sum_{i=1}^3 q_i \times \frac{\partial H}{\partial q_i}}_{\text{Gravity/Potential}} - \underbrace{D(q) \frac{\partial H}{\partial p}}_{\text{Friction}} + \underbrace{F}_{\text{Control}}\}dt + \tau_{\text{stoch}}\circ dW_{t}
\end{aligned}
\end{equation}

where the \textbf{potential energy} function is determined by the height of the bob: $V(q) = mgl\, e_3^\top q e_3$ (up to a constant), with $e_3 = [0, 0, 1]^\top$, which is maximal at the upright $q=I$. We assume the same inertia about all three body axes,
   $I=ml^2I_3$. With $m=l=1$, the inverse inertia is:
\[
M^{-1}(q) = \mathbb{I}_{3 \times 3} = 
\begin{bmatrix}
1 & 0 & 0 \\
0 & 1 & 0 \\
0 & 0 & 1
\end{bmatrix}.
\]
The actuation applies a 3D torque in the body frame through the control map matrix $G$.

The dissipation matrix $D(\omega) = d(\omega) \mathbb{I}_{3 \times 3}$ accounts for friction, where the scalar $d(\omega)$ follows the conditional logic:
\[
d(\omega) = 
\begin{cases} 
0.5 & \text{if constant friction} \\[10pt]
0.5 \cdot \left( 1 + \|\omega\| \right) & \text{if varying friction}
\end{cases}.
\]

\begin{figure}[H]
    \centering
    \includegraphics[width=0.5\linewidth]{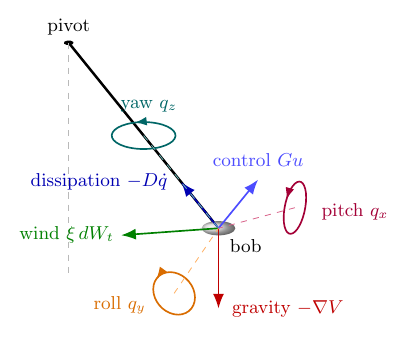}
    \caption{3D Windy Rigid-Body Pendulum}
    \label{fig:windy_pendulum}
\end{figure}

Figure \ref{fig:windy_pendulum} provides a visualization of this $SO(3)$ windy pendulum system.

\subsubsection{Experimental Conditions}
Our custom 3D windy pendulum environment augments the standard dynamics with distinct conditions:

\paragraph{Viscous Friction (Dissipation):} 
We include a linear friction term $\tau_{\text{fric}} = D(\omega)\,\omega$, which opposes the angular velocity. We evaluate both a constant friction baseline and a varying friction that grows with the angular speed.

\paragraph{Observation Noise:} 
We simulate a realistic sensor by applying geometrically-correct noise directly to the physical states. To maintain the $SO(3)$ manifold structure, we apply noise to the rotation matrix using the exponential map with a sampled Lie algebra element $\epsilon \sim \mathcal{N}(0, \sigma^2 \mathbb{I}_3)$ where $R_{\text{noisy}} = R \exp(\hat{\epsilon})$

For the angular velocity, we directly add standard Gaussian noise $\nu \sim \mathcal{N}(0, \sigma^2 \mathbb{I}_3)$ where $\omega_{\text{noisy}} = \omega + \nu$

This process creates a challenging dataset where the rotational observations still strictly lie on the $SO(3)$ manifold.

\paragraph{Datasets.} Each dataset has $100$ training trajectories of $5$\,s ($\Delta t=0.05$\,s, $10$ simulator substeps per sample), with $G=\mathrm{diag}(0.5,0.7,0.8)$, inputs $u\sim U(-7,7)^3$ redrawn at every sample, initial attitudes uniform on $SO(3)$ and $\omega_0\sim U(-1,1)^3$. The deterministic dataset has constant friction $0.5$, no wind and observation noise $\sigma=0.25$; the windy dataset has rate-dependent friction, an isotropic world-frame wind force $0.5\,dW$ on the bob, i.e.\ a true angular-velocity diffusion $\xi=(0.5,0.5,0)$, and observation noise $\sigma=0.5$. All models are evaluated on $10$ held-out trajectories of $10$\,s.

\subsection{SE(3) Quadrotor with stochastic wind}
\label{sec:appendix:experiment:se3_quadrotor}

\subsubsection{Environment Description}
The windy quadrotor environment extends the $SO(3)$ pendulum experiment to a full rigid body with translation and rotation. The plant is the ``+''-configuration Crazyflie~2.x (CF2P) model of gym-pybullet-drones \citep{panerati2021learning}, simulated contact-free in PyBullet \citep{coumans2021} at $1000$\,Hz, to which we add the rate-dependent damping and the stochastic wind described below.
The vehicle evolves on $SE(3)$, is actuated by four rotors, experiences gravitational and viscous forces, and is disturbed by deterministic and stochastic wind. The formulation uses left-trivialized body-frame velocities so that the rigid-body mass matrix, dissipation matrix, and rotor input map are constant. As in the pendulum experiment, observation noise is added in a geometry-preserving manner: Euclidean components receive additive Gaussian noise, whereas the attitude is perturbed through the $SO(3)$ exponential map.

\subsubsection{State Space}
The configuration is
\begin{equation}
q=(x_w,R)\in SE(3), \qquad x_w\in\mathbb{R}^3, \qquad R\in SO(3),
\end{equation}
where $x_w$ is the center-of-mass position in the world frame and $R$ maps body-frame vectors to the world frame. The left-trivialized velocity is $\zeta=(v_b,\omega_b)\in\mathfrak{se}(3)\cong\mathbb{R}^6$, where $v_b$ and $\omega_b$ are body-frame linear and angular velocities. The configuration is stored as $x_w$ together with the flattened $3\times3$ rotation matrix, and the full observation is therefore
\begin{equation}
\mathrm{obs}=[x_w,\operatorname{vec}(R),v_b,\omega_b]\in\mathbb{R}^{18}.
\end{equation}
The kinematic equations are
\begin{equation}
\dot{x}_w=Rv_b, \qquad \dot{R}=R[\omega_b]_{\times}.
\end{equation}

\subsubsection{Hamiltonian and Momenta}
Let $m$ be the vehicle mass and $J\in\mathbb{R}^{3\times3}$ its body-frame inertia tensor. Because the body frame is attached at the center of mass, the generalized mass matrix is block diagonal and constant,
\begin{equation}
\mathcal M=\operatorname{blkdiag}(mI_3,J), \qquad \mathcal M^{-1}=\operatorname{blkdiag}(m^{-1}I_3,J^{-1}).
\end{equation}
The body-frame linear and angular momenta are $\mathbf p=mv_b$ and $\Pi=J\omega_b$. With $e_3=(0,0,1)^\top$, the potential and total Hamiltonian are
\begin{equation}
V(x_w,R)=mg\,e_3^\top x_w, \qquad H(q,\mathbf p,\Pi)=\frac{1}{2m}\|\mathbf p\|^2+\frac12\Pi^\top J^{-1}\Pi+mg\,e_3^\top x_w.
\end{equation}
Unlike the spherical pendulum, the gravitational potential depends only on height and not on attitude. Consequently, gravity acts as a force at the center of mass and produces no gravitational body torque.

\subsubsection{Port-Hamiltonian Dynamics Formulation}
\label{sec:se3-formulation}
On $\mathfrak{se}(3)^*$, the coadjoint term is $\operatorname{ad}^{*}_{(v,\omega)}(\mathbf p,\Pi)=(\mathbf p\times\omega,\Pi\times\omega+\mathbf p\times v)$. For the present mass matrix, $\mathbf p=mv_b$ and hence $\mathbf p\times v_b=0$. The resulting controlled rigid-body dynamics are
\begin{equation}
\begin{aligned}
\dot{x}_w &= Rv_b, \\
\dot{R} &= R[\omega_b]_{\times}, \\
\dot{\mathbf p} &= \mathbf p\times\omega_b-mgR^\top e_3-D_v(v_b)\,v_b+(Gu)_{1:3}+R^\top F_{\mathrm{wind}}, \\
\dot{\Pi} &= \Pi\times\omega_b-D_\omega(\omega_b)\,\omega_b+(Gu)_{4:6}
\end{aligned}
\end{equation}
Equivalently, the deterministic dynamics can be written in port-Hamiltonian form as $\dot{x}=(\mathcal J(x)-\mathcal R)\nabla H+\mathcal G u+$ wind ports, where $\mathcal J(x)=-\mathcal J(x)^\top$ contains the $SE(3)$ kinematic and coadjoint couplings, $\mathcal R\succeq0$ contains translational and rotational viscous dissipation, and $\mathcal G$ maps the four rotor thrusts into the six-dimensional body wrench.

\subsubsection{Dissipation and Rotor Input Map}
The dissipation is rate-dependent,
\begin{equation}
D=\operatorname{blkdiag}\big(D_v(v_b),D_\omega(\omega_b)\big), \qquad D_v(v_b)=c\,m\,(1+\|v_b\|)\,I_3, \qquad D_\omega(\omega_b)=c\,(1+\|\omega_b\|)\,J,
\end{equation}
with $c=0.5$. It produces the body-frame force $-c\,m(1+\|v_b\|)v_b$ and torque $-c(1+\|\omega_b\|)J\omega_b$, so that $\mathcal M^{-1}D=\operatorname{blkdiag}\big(0.5(1+\|v_b\|)I_3,\ 0.5(1+\|\omega_b\|)I_3\big)$.

The control input $u=[F_1,F_2,F_3,F_4]^\top\in\mathbb{R}^4$ contains the four rotor thrusts ($F_i=k_f\Omega_i^2$, $0\le F_i\le F_{\max}=0.149$\,N), so thrust and torque are exactly linear in $u$. For the ``+''-configuration Crazyflie (CF2P) used here,
\begin{equation}
\begin{bmatrix}F_b\\ \tau_b\end{bmatrix}=Gu, \qquad G=\begin{bmatrix}0&0&0&0\\0&0&0&0\\1&1&1&1\\0&\ell&0&-\ell\\-\ell&0&\ell&0\\-\kappa&\kappa&-\kappa&\kappa\end{bmatrix}, \qquad \ell=0.0397\,\mathrm{m},\quad \kappa=\frac{k_m}{k_f}=0.0251\,\mathrm{m},
\end{equation}
with $m=0.027$\,kg and $J=\mathrm{diag}(2.40,\,2.40,\,3.23)\times10^{-5}$\,kg\,m$^2$.

Thus the system is underactuated: $G\in\mathbb{R}^{6\times4}$ has four input channels for six rigid-body velocity degrees of freedom. The first two force rows are zero because the rotors generate thrust only along the body $z$ axis; lateral motion is achieved indirectly by changing attitude.

\subsubsection{Stochastic Wind and Full SDE}
Stochastic wind is represented by independent translational and rotational Wiener processes $W_f,W_\tau\in\mathbb{R}^3$ acting on the vehicle. The complete Stratonovich SDE is
\begin{equation}
\begin{aligned}
dx_w &= Rv_b\,dt, \\
dR &= R[\omega_b]_{\times}\,dt, \\
d\mathbf p &= \left(\mathbf p\times\omega_b-mgR^\top e_3-D_v(v_b)\,v_b+(Gu)_{1:3}\right)dt+\sigma_fR^\top\circ dW_f, \\
d\Pi &= \left(\Pi\times\omega_b-D_\omega(\omega_b)\,\omega_b+(Gu)_{4:6}\right)dt+\sigma_\tau\circ dW_\tau.
\end{aligned}
\end{equation}
The force channel is multiplicative through $R^\top$, whereas the torque channel is additive. The corresponding diffusion in the embedded 18-dimensional state has zero configuration rows and momentum block $\operatorname{blkdiag}(\sigma_fR^\top,\sigma_\tau I_3)$. Because the diffusion depends only on $R$ while all nonzero diffusion rows lie in the momentum variables, the Stratonovich-to-It\^o correction vanishes for this environment. In our dataset $\sigma_f=0.5\,m$ and $\sigma_\tau=0.5\,J$, so the true twist diffusion is $\xi=0.5$ in all six channels.

\begin{figure}
    \centering
    \includegraphics[width=0.8\linewidth]{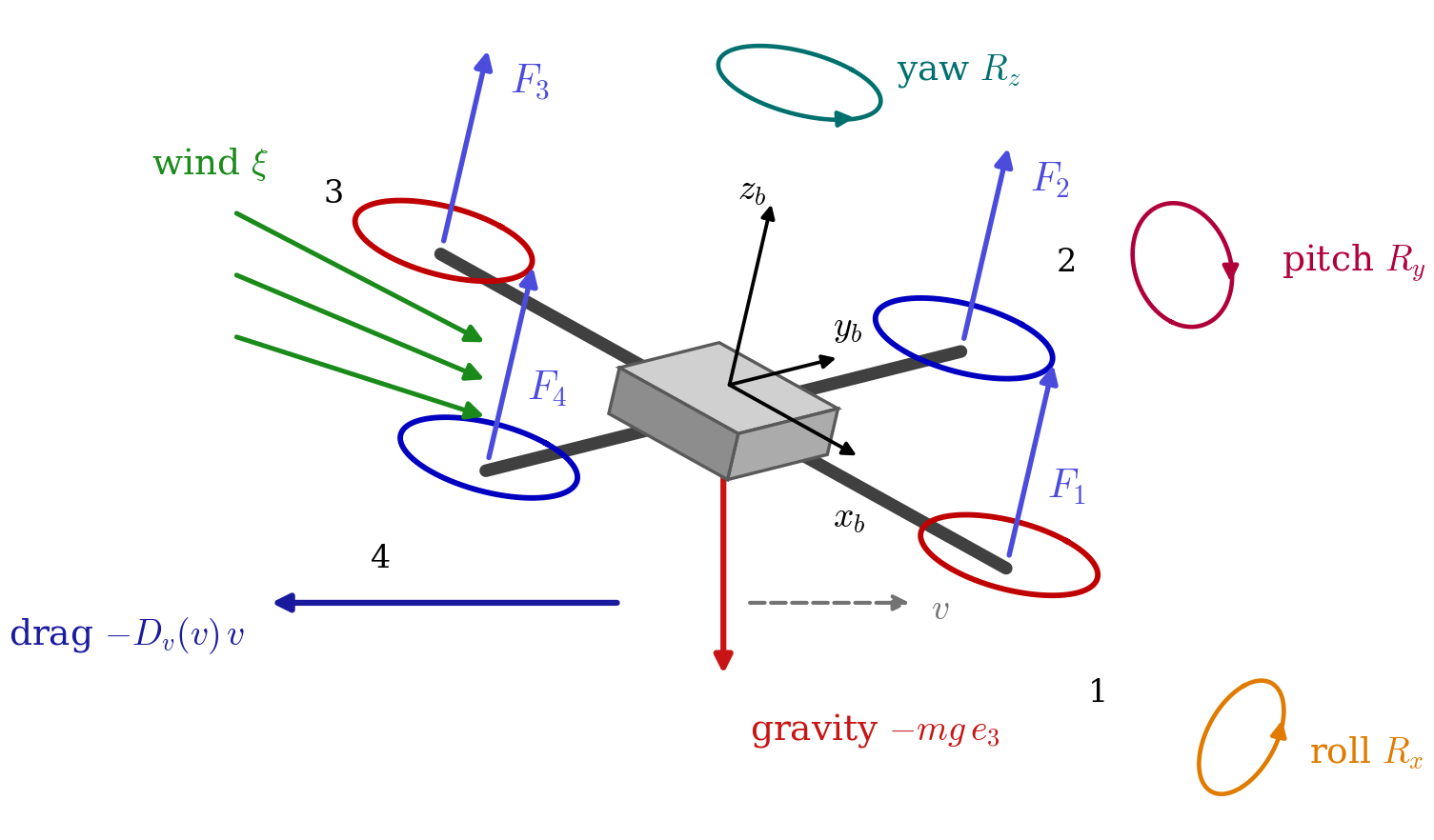}
    \caption{$SE(3)$ windy quadrotor. The four rotor thrusts $F_1,\dots,F_4$ are the inputs; gravity, velocity-dependent air drag $-D_v(v)\,v$ and random wind forces and torques $\xi\,dW_t$ act on the body.}
    \label{fig:quadrotor}
\end{figure}

Figure \ref{fig:quadrotor} provides a visualization of this $SE(3)$ quadrotor system.

\subsubsection{Observation Noise}
For observation-noise standard deviation $\sigma_{\mathrm{obs}}$, Euclidean state components are perturbed additively while the rotation remains on the manifold:
\begin{equation}
R_{\mathrm{obs}}=R\exp([\epsilon]_{\times}), \qquad \epsilon\sim\mathcal N(0,\sigma_{\mathrm{obs}}^2I_3),
\end{equation}
\begin{equation}
x_{w,\mathrm{obs}}=x_w+\nu_x, \qquad v_{b,\mathrm{obs}}=v_b+\nu_v, \qquad \omega_{b,\mathrm{obs}}=\omega_b+\nu_\omega, \qquad \nu_x,\nu_v,\nu_\omega\sim\mathcal N(0,\sigma_{\mathrm{obs}}^2I_3).
\end{equation}
This produces noisy observations that retain a valid $SO(3)$ attitude block while independently perturbing position and velocity measurements.

\subsubsection{Experimental Conditions}
The training data are $300$ flights of $5$\,s sampled at $50$\,Hz ($\Delta t=0.02$\,s; PyBullet physics at $1000$\,Hz), flown by a noisy PID tracker with short torque kicks for excitation. Observation noise is $\sigma_{\mathrm{obs}}=0.25$ on $x_w$, $v_b$ and $\omega_b$ and $R\exp([\epsilon]_\times)$, $\epsilon\sim\mathcal N(0,0.25^2I_3)$, on the attitude, the stochastic wind of the previous section is the only disturbance, and all physical parameters are fixed. Evaluation uses $10$ held-out noise-free reference flights of $10$\,s.

The learned structured components have direct physical ground truths: $\mathcal M^{-1}=\operatorname{blkdiag}(m^{-1}I_3,J^{-1})$, $V=mg e_3^\top x_w$, the rate-dependent $D$ above, the sparse rectangular rotor map $G$, and the twist diffusion $\xi=0.5$. Because the individual components are identifiable only up to the scaling discussed in Appendix~\ref{sec:appendix:physics-id}, we compare the gauge-invariant products ($M^{-1}G$, $M^{-1}D$, the gravitational acceleration, and $\xi$) with their true values, in addition to trajectory-level metrics.

\subsection{Real World Quadrotor Dataset}
\label{sec:appendix:idsia}
We take the real-flight nano-drone benchmark of \citet{busetto2026nonlinear}: a Crazyflie~2.1 Brushless flown in a motion-capture room, logged at $100$~Hz with the measured rotor speeds as input, which we convert to a wrench with the benchmark's own mixer. We follow the benchmark's split and protocol: train on the Square, Random and Chirp flights, test on the unseen Melon flights, and from every admissible start state ($19{,}347$ windows) predict $0.5$~s ($50$ steps) open loop from the recorded inputs, with the input at time $t$ driving the step from $t$ to $t+1$ as in the benchmark. Table~\ref{tab:multistep-prediction} reports the mean absolute error averaged over the $50$ horizons; every row is scored by the same script on each Melon run separately (Eq.~(11) of \citet{busetto2026nonlinear}), with the baselines taken from the benchmark's released predictions. Our models train on all four runs of each training trajectory and are evaluated at the final checkpoint, whereas the benchmark's learned baselines train on runs $1$--$3$ and keep the epoch with the lowest loss on run $4$. Our models impose physical structural priors through the port-Hamiltonian formulation, while learning the physical coefficients from data. The Physics and Hybrid baselines are built on the vehicle's known parameters (mass, gravity, the identified thrust and torque coefficients and the thrust limit and Physics holds the angular velocity constant). LieSPHGP and PH-NODE receive none of them: mass and inertia, potential energy, damping and the map from the wrench to the accelerations are learned, and only the conversion of the measured rotor speeds to a wrench uses the benchmark's mixer. PH-NODE is trained with the same recipe as in Table~\ref{tab:quadrotor-closed-loop-comparison}. The benchmark data are not raw measurements: the authors low-pass filter every signal with a zero-phase fourth-order Butterworth filter (cutoffs between $10$ and $25$~Hz) to reduce high-frequency measurement noise. This experiment therefore tests prediction on preprocessed trajectories. Using predictions with posterior-mean GP weights, LieSPHGP achieves the lowest position and velocity errors among all models, including the
benchmark's physics model. Its attitude error is higher than that of every physics-based and learned baseline, and its angular-velocity error is higher than every baseline's, including the naive hold: holding the measured angular velocity constant, as the physics and naive models do, is accurate over the short $0.5$-s horizon.

\begin{table}[t]
\centering
\caption{Comparison on the real-world quadrotor dataset from \citet{busetto2026nonlinear}. Multi-step prediction MAE averaged over rollout horizons $1$ through $50$ (step size $0.01$\,s, i.e.\ up to $0.5$\,s) from every start state of the held-out Melon flights ($19{,}347$ windows), scored per flight as in Eq.~(11) of \citet{busetto2026nonlinear}. Baselines: the benchmark's released predictions rescored by the same script. The published values pool the three flights (e.g.\ Physics $0.0464$, $0.2125$, $0.1020$, $0.5817$) and give the same ranking. LieSPHGP: predictions with posterior-mean GP weights of a single training run. PH-NODE \citep{duong2024phnode}: a single training run. Bold: best in each column.}
\label{tab:multistep-prediction}
\begin{tabular}{lcccc}
\toprule
\textbf{Model}
& $p$ Error (m)
& $v$ Error (m/s)
& $R$ Error (rad)
& $\omega$ Error (rad/s) \\
\midrule
Na\"ive \citep{busetto2026nonlinear}    & 0.3536 & 0.7745 & 0.1639 & 0.5797 \\
Physics \citep{busetto2026nonlinear}    & 0.0448 & 0.2094 & \textbf{0.1014} & 0.5797 \\
Res-MLP \citep{busetto2026nonlinear}    & 0.0795 & 0.2488 & 0.1210 & 0.7097 \\
Hybrid \citep{busetto2026nonlinear}     & 0.0456 & 0.2051 & 0.1228 & \textbf{0.5782} \\
Res-LSTM \citep{busetto2026nonlinear}   & 0.1182 & 0.3314 & 0.1129 & 0.7149 \\
PH-NODE \citep{duong2024phnode}         & 0.0393 & 0.2262 & 0.1451 & 0.7849 \\
\midrule
{LieSPHGP} (ours)                       & \textbf{0.0363} & \textbf{0.1926} & 0.1284 & 0.7693 \\
\bottomrule
\end{tabular}
\end{table}

\subsection{Real-World Underwater Vehicle (SE(3) Lie Group)}
\label{app:underwater}

\begin{figure}
    \centering
    \includegraphics[width=0.75\linewidth]{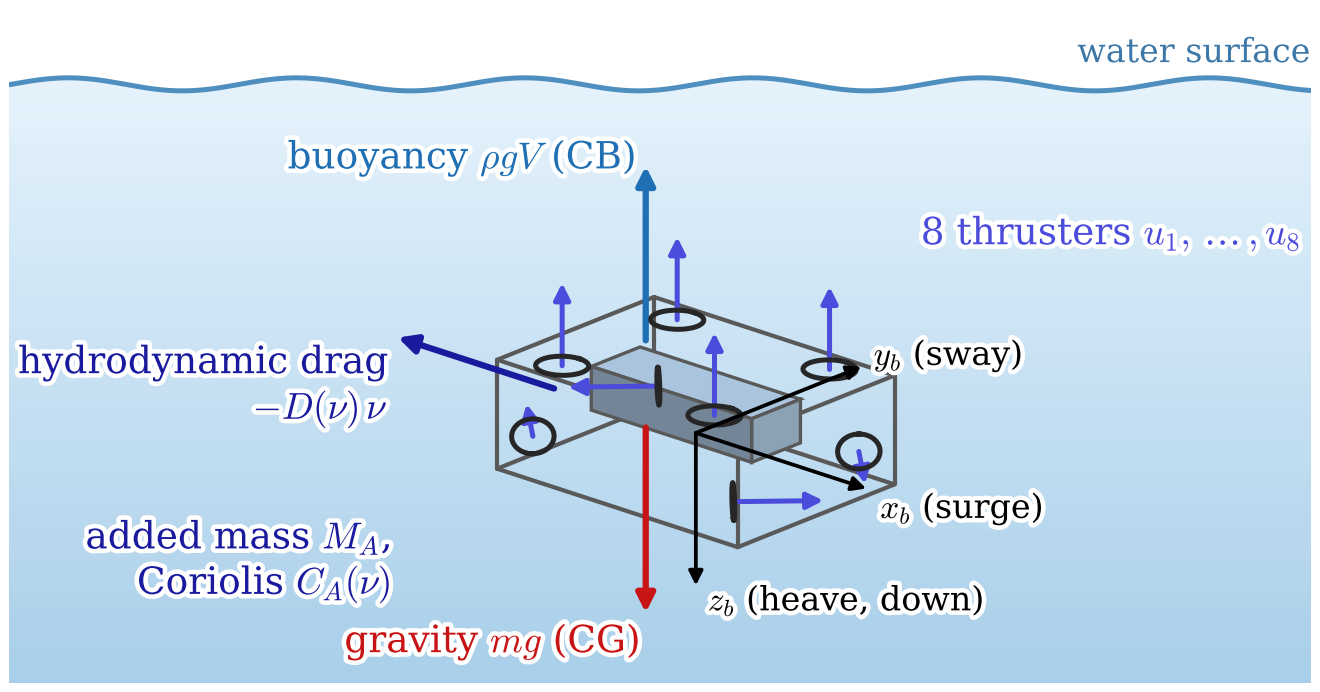}
    \caption{Real BlueROV2 Heavy underwater vehicle with eight thruster inputs
$u_1,\dots,u_8$. Gravity acts at the center of gravity (CG), and buoyancy
acts at the center of buoyancy (CB). Water contributes hydrodynamic drag
$-D(\nu)\nu$, added mass $M_A$, and the associated Coriolis term
$C_A(\nu)$. The body frame follows the marine convention, with $z_b$
pointing down.}
    \label{fig:underwater_vehicle}
\end{figure}

\subsubsection{Experimental Setup and Dataset}

We evaluate LieSPHGP on real-world underwater vehicle trajectories from the Marinarium dataset \citep{torroba2026marinarium}. The dataset was collected using a BlueRobotics BlueROV2 Heavy remotely operated underwater vehicle in the Marinarium facility at KTH Royal Institute of Technology. The facility contains a $9\times5\times3$ m water basin equipped with an underwater Qualisys motion-capture system, enabling accurate pose measurements during controlled underwater experiments. The vehicle is equipped with eight thrusters and an onboard inertial measurement unit (IMU), with measurements synchronized through the ROS 2 control and logging infrastructure.

The dataset contains three recordings, each longer than $15$ minutes.
Following \citet{torroba2026marinarium}, we use the manual recording with
a chronological $80\%$/$20\%$ train--test split, resampled to $50$\,Hz
($\Delta t=0.02$\,s). The eight raw thruster commands serve as control
inputs. We remove motion-capture dropouts and pose glitches without
interpolation, leaving $114$ training trajectories of $5$\,s each.
Following the benchmark, each model predicts $H\in\{1,10,100\}$ steps
ahead from each of the $9{,}164$ test samples using the recorded commands.
We report endpoint RMSE across the 12 state variables (position, ZYX
Euler angles, body linear and angular velocities) in Table \ref{tab:rov-real}. We also measure valid
prediction time (VPT), the time until position error first exceeds
$0.1$\,m, on $15$ held-out $10$-s segments initialized from measured
states. We limit endpoint prediction to $2$\,s, focusing on short-horizon
prediction under commands issued by a pilot responding to the recorded
vehicle motion.

We train LieSPHGP using Section~\ref{sec:methods:loss}, with $2$-s windows
($101$ samples, stride $75$), a batch size of $16$, and $5{,}000$ steps.

We make three changes for this vehicle. The input map
$G\in\mathbb R^{6\times8}$ takes all eight thruster commands.
The translational inverse mass $M_1^{-1}$ is a full symmetric
positive-definite matrix that captures anisotropic added mass.
The potential includes an attitude-dependent mean,
\[
V_\theta(x,R)
= \lambda_x^\top x
+ \lambda_R^\top\mathrm{vec}(R)
+ \phi_V(x)^\top W_V.
\]
Its rotational gradient captures the restoring torque caused by the
offset between the centers of gravity and buoyancy. The inertia remains
block diagonal, excluding translation-rotation coupling from added mass.
Given the small motion-capture
noise, we set the minimum observation variance to $10^{-6}$.
We report results over two training seeds.
\subsubsection{SE(3) State Space and Dynamics}

As with the quadrotor, the underwater vehicle is a six-degree-of-freedom rigid body whose configuration evolves on the special Euclidean group $SE(3)$. We represent its configuration and body-frame velocity as
\begin{equation}
q=(x,R)\in SE(3),\qquad
\zeta=(v_b,\omega_b)\in\mathfrak{se}(3)\cong\mathbb{R}^6,
\end{equation}
where $x\in\mathbb{R}^3$ denotes the position in the inertial frame, $R\in SO(3)$ maps body-frame coordinates to the inertial frame, and $v_b,\omega_b\in\mathbb{R}^3$ denote body-frame linear and angular velocities. The corresponding kinematic equations are
\begin{equation}
\dot{x}=Rv_b,\qquad \dot{R}=R\widehat{\omega}_b.
\end{equation}
This formulation replaces the Euler-angle orientation representation used in the original dataset with a rotation matrix, avoiding coordinate singularities and allowing the Lie-IMEX integrator to preserve the $SO(3)$ constraint.

Underwater vehicle dynamics differ substantially from those of aerial rigid bodies because the surrounding fluid introduces added mass, nonlinear hydrodynamic damping, and buoyancy forces. The standard Fossen model used in \citet{torroba2026marinarium} is
\begin{equation}
M\dot{\zeta}+C(\zeta)\zeta+D(\zeta)\zeta+g(q)=w(u),
\end{equation}
where $M=M_{\mathrm{RB}}+M_{\mathrm{A}}$ combines rigid-body inertia and hydrodynamic added mass, $C(\zeta)=C_{\mathrm{RB}}(\zeta)+C_{\mathrm{A}}(\zeta)$ contains rigid-body and added-mass Coriolis and centripetal terms, $D(\zeta)$ represents hydrodynamic damping, $g(q)$ contains gravitational and buoyancy restoring forces, and $w(u)\in\mathbb{R}^6$ is the generalized wrench generated by the eight thrusters.

This model admits an energy-based mechanical interpretation compatible with the port-Hamiltonian framework. With body-frame generalized momentum $p=M\zeta$ and constant inertia matrix $M$, the Hamiltonian takes the form
\begin{equation}
H(q,p)=\frac{1}{2}p^\top M^{-1}p+V_{\mathrm{gb}}(q),
\end{equation}
where $V_{\mathrm{gb}}$ represents gravitational and buoyancy potential energy. The inertial and Coriolis couplings contribute to the energy-conserving interconnection structure, while hydrodynamic damping contributes to dissipation. Thruster forces enter through external control ports, and a stochastic diffusion term can represent unresolved environmental forces and modeling uncertainty. LieSPHGP learns the corresponding structured components from observed trajectories rather than requiring all hydrodynamic parameters to be specified analytically.

\subsubsection{Differences from the Windy Quadrotor}

Although both systems evolve on $SE(3)$ and share the same rigid-body kinematics, their physical dynamics and experimental conditions differ considerably. The quadrotor experiment uses simulated trajectories with four rotors, a rigid-body inertia matrix, gravity, prescribed viscous damping, and stochastic wind disturbances. In contrast, the underwater vehicle experiment uses measured trajectories from a physical eight-thruster vehicle operating in water. The surrounding fluid produces added-mass effects, nonlinear translational and rotational drag, and restoring forces associated with gravity and buoyancy. These interactions depend on the vehicle geometry, velocity, angular velocity, and hydrodynamic conditions.

The underwater vehicle also has a different actuation structure. The quadrotor generates thrust primarily along its body-fixed vertical axis and controls lateral motion indirectly through attitude changes. The BlueROV2 Heavy instead uses eight independently commanded thrusters to generate forces and torques in multiple body-frame directions. Consequently, its control allocation and hydrodynamic interactions differ from the quadrotor, even though both systems have six rigid-body degrees of freedom.

Unlike the simulated quadrotor, for which the underlying physical parameters and stochastic disturbances are prescribed, the underwater dataset contains measurement uncertainty and unmodeled physical effects arising from real hardware and fluid interactions. This experiment therefore evaluates whether the learned stochastic port-Hamiltonian representation generalizes to real-world dynamics without access to an exact ground-truth model.

\subsubsection{Physics and Koopman Baselines}

We compare against three baselines evaluated on the same test samples. \emph{Persistence} predicts
$\hat x_{k+H}=x_k$. \emph{Physics} is the Fossen model with the nominal BlueROV2 parameters used by
\citet{torroba2026marinarium} and the T200 thrust curve at the measured battery voltage, integrated with a Lie-group Heun
scheme ($10$ substeps per sample) and no parameter is fitted. The physics model therefore knows the vehicle's inertia and added mass ($M$), its gravity and buoyancy restoring forces ($V$), its hydrodynamic damping ($D$), its thruster map and thrust curve ($G$) in advance, and receives the thrust in newtons. LieSPHGP receives only the raw thruster commands and learns $M^{-1}$, $V$, $D$ and $G$ from the $114$ training trajectories. \emph{Koopman} is the EDMDc model of
\citet{torroba2026marinarium} with their code and hyperparameters ($K=500$ RBF centers, $\gamma=3$, $\lambda=0.1$),
fitted on the same $114$ training trajectories as LieSPHGP.

\paragraph{Koopman operator model.}
The Koopman baseline employs Extended Dynamic Mode Decomposition with control (EDMDc), using a dictionary of Gaussian radial basis functions (RBFs) to approximate nonlinear dynamics through a finite-dimensional linear system. The original 12-dimensional state $x_k$, consisting of position, Euler angles, and body-frame velocities, is lifted into an augmented observable space
\begin{equation}
z_k=\Theta(x_k)=\begin{bmatrix}x_k^\top & \phi_1(x_k) & \cdots & \phi_K(x_k)\end{bmatrix}^\top,
\qquad
\phi_i(x)=\exp\left(-\gamma\|x-c_i\|_2^2\right),
\end{equation}
where the RBF centers $c_i$ are obtained by $k$-means clustering. The lifted dynamics are approximated by
\begin{equation}
z_{k+1}=Az_k+Bu_k,\qquad \widehat{x}_k=Cz_k,
\end{equation}
where the matrices $A$ and $B$ are learned using ridge-regularized least squares and $C$ extracts the original state coordinates. The reported implementation uses $K=500$ RBF centers, width parameter $\gamma=3.0$, and ridge regularization $\lambda=0.1$.

The Koopman approach represents nonlinear dynamics through linear evolution in a learned observable space, providing efficient prediction without explicitly identifying physical mass, potential energy, dissipation, or control components. In contrast, LieSPHGP learns structured stochastic port-Hamiltonian dynamics directly on the rigid-body configuration manifold, preserving the rotational geometry through Lie-group integration. The comparison therefore evaluates physically structured probabilistic learning against both a parameterized first-principles model and a data-driven lifted linear representation.

\subsubsection{Evaluation on Unseen Trajectories}
The models trained on the manually piloted recording are tested, without retraining, on the separate recording of \citet{torroba2026marinarium}, which was flown in attitude-stabilized mode (Table~\ref{tab:rov-transfer}). At $2$\,s, LieSPHGP's endpoint error is half that of the Koopman model, whose basis functions are centered on training states, and $39\,\%$ lower than that of the physics model. LieSPHGP also has the lowest position error over $0$--$3$\,s. Persistence has the lowest endpoint errors because the stabilized vehicle moves slowly and hardly rotates.

\begin{table}[t]
\centering
\caption{\textbf{BlueROV2: evaluation on unseen trajectories.} Models trained on the manual recording are evaluated, without
retraining, on the attitude-stabilized recording of \citet{torroba2026marinarium} ($45{,}852$ test samples, $60$
segments of $10$\,s). Endpoint RMSE as in Table~\ref{tab:rov-real}; position error averaged over $0$--$3$\,s; VPT as in
Table~\ref{tab:rov-real}. LieSPHGP: mean $\pm$ std over two training seeds.}
\label{tab:rov-transfer}
\setlength{\tabcolsep}{4pt}
\begin{tabular}{@{}lcccc@{}}
\toprule
\textbf{Model} & $T=10$ ($0.2$\,s) & $T=100$ ($2$\,s) & Position $0$--$3$\,s (m) & \textbf{VPT (s)} \\
\midrule
Persistence                    & $\mathbf{0.218}$ & $\mathbf{0.397}$ & $0.479$ & $0.51$ \\
Physics (published parameters) & $0.359$ & $0.673$ & $0.282$ & $\textbf{1.19}$ \\
Koopman (EDMDc)                & $0.331$ & $0.844$ & $0.544$ & $0.58$ \\
\midrule
LieSPHGP (ours)                & $0.226 \pm 0.003$ & $0.412 \pm 0.007$ & $\mathbf{0.197 \pm 0.009}$ & $1.13 \pm 0.03$ \\
\bottomrule
\end{tabular}
\end{table}

\subsection{Physics Identification}
\label{sec:appendix:physics-id}

We evaluate the ability of the learned models to identify the physical quantities governing the observed dynamics of the windy pendulum on $SO(3)$ and the quadrotor on $SE(3)$, along ground-truth closed-loop trajectories. We compare the effective control map $M^{-1}G$, the damping operator $M^{-1}D$, the potential-induced acceleration $a_V$, and the diffusion $\xi$. For the pendulum, $a_V(q)=M^{-1}(q)\sum_{i=1}^{3}q_i\times\nabla_{q_i}V(q)$ with $q_i$ the rows of $q$; for the quadrotor, $a_V=-M_1^{-1}R^{\top}\nabla_x V(x)$. These quantities are invariant under the constant rescaling $(M,V,D,G)\mapsto(cM,cV,cD,cG)$, $c>0$, which leaves the dynamics unchanged, whereas the individual components are not uniquely determined by motion data; this invariance alone does not establish unique identifiability of the combined quantities. The diffusion $\xi$ acts directly on the velocity and is therefore unaffected by this rescaling: only the momentum diffusion $M\xi$ scales with $c$. Since $\xi$ is diagonal and positive, it is uniquely determined by the covariance $\xi\xi^{\top}$. We compare it with the true velocity diffusion, i.e.\ the simulator's force and torque noise divided by the mass and inertia.

\begin{figure}
    \centering
    \includegraphics[width=0.3\linewidth]{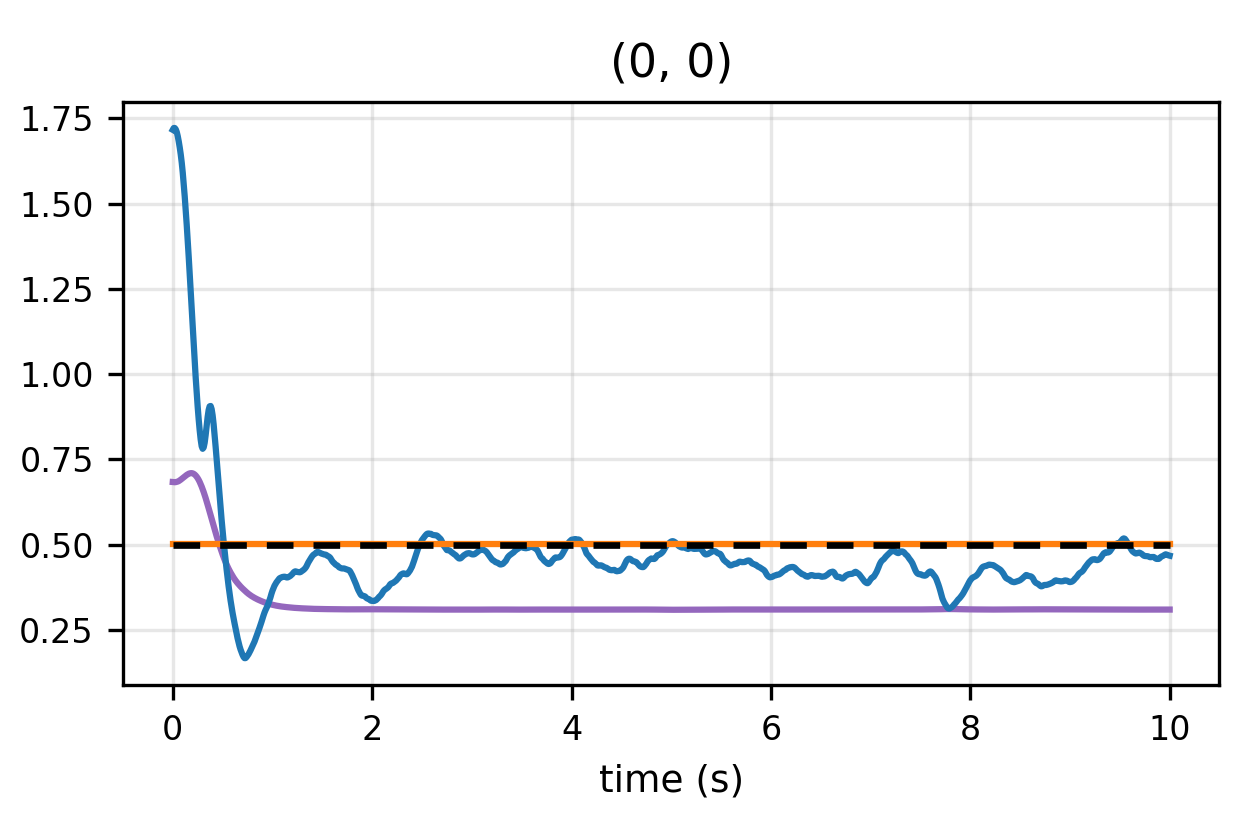}
    \includegraphics[width=0.3\linewidth]{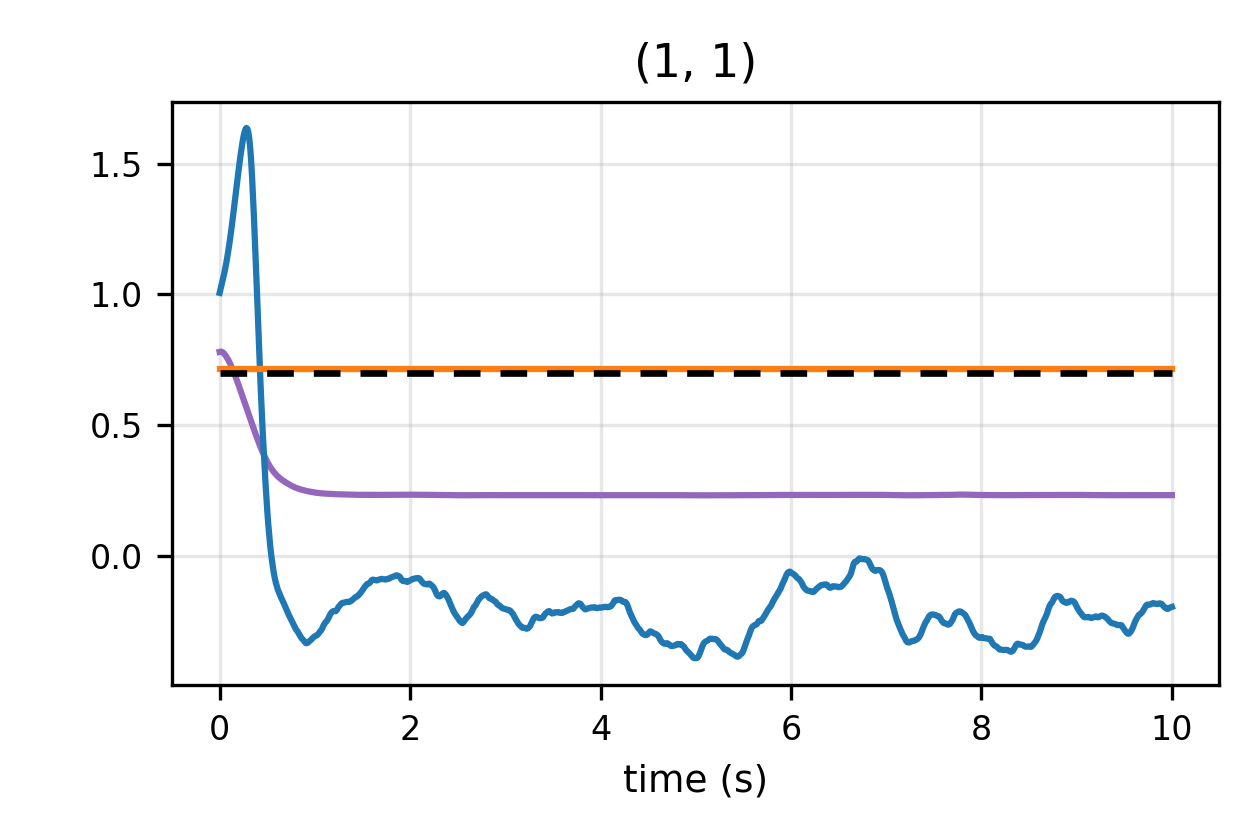}
    \includegraphics[width=0.3\linewidth]{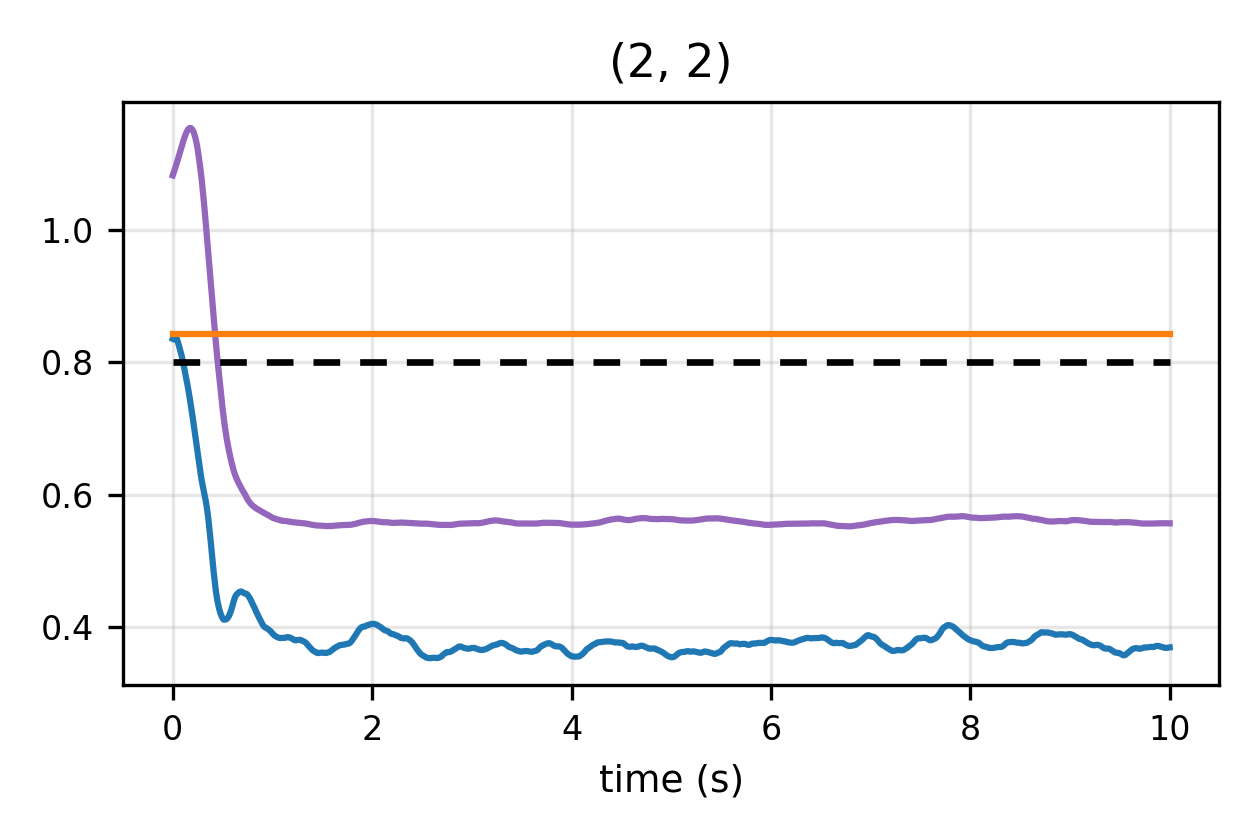}
    \includegraphics[width=0.3\linewidth]{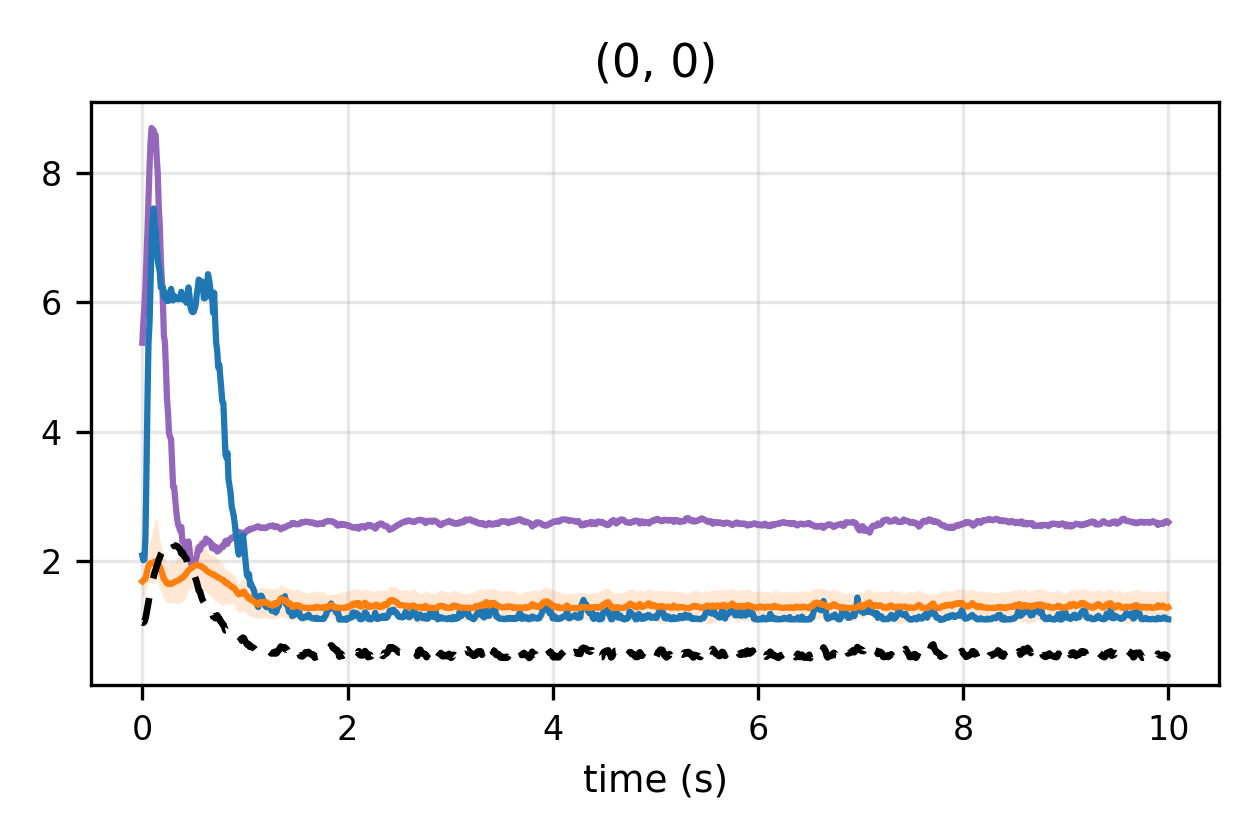}
    \includegraphics[width=0.3\linewidth]{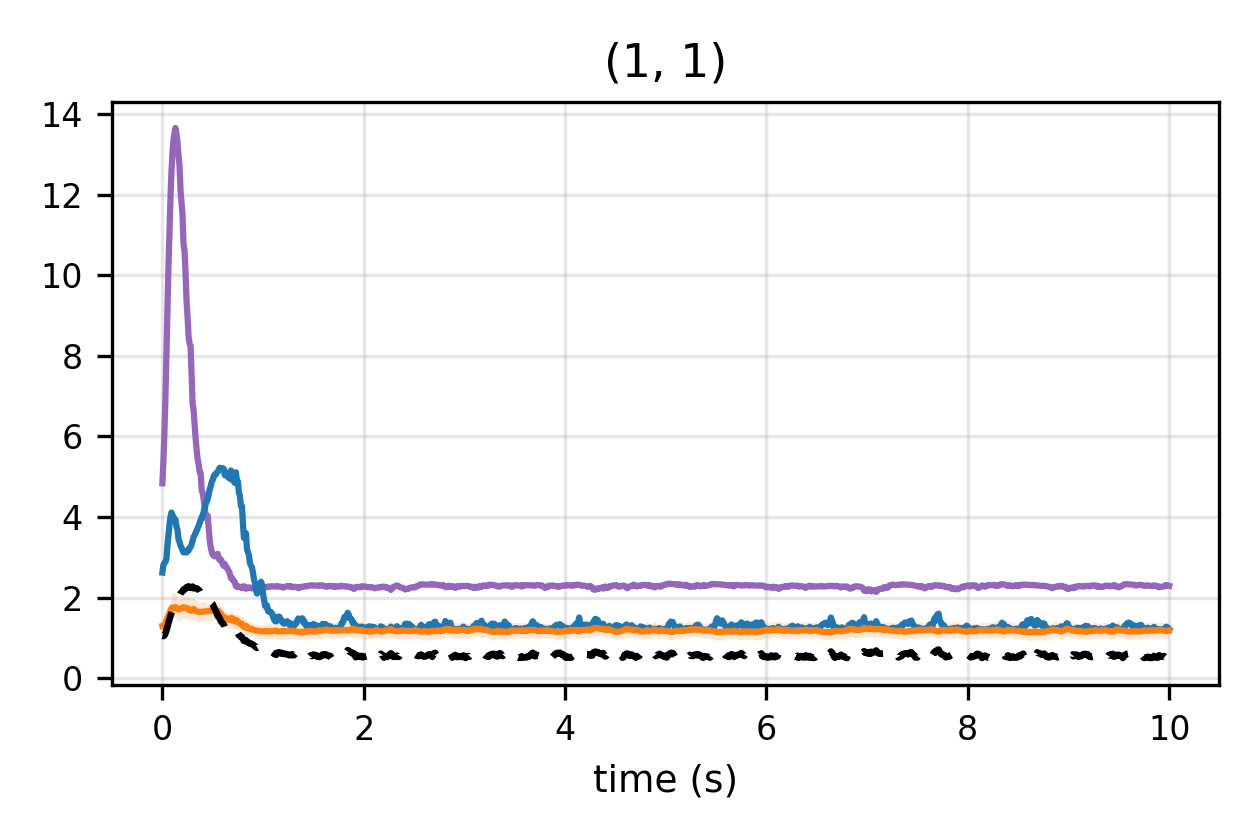}
    \includegraphics[width=0.3\linewidth]{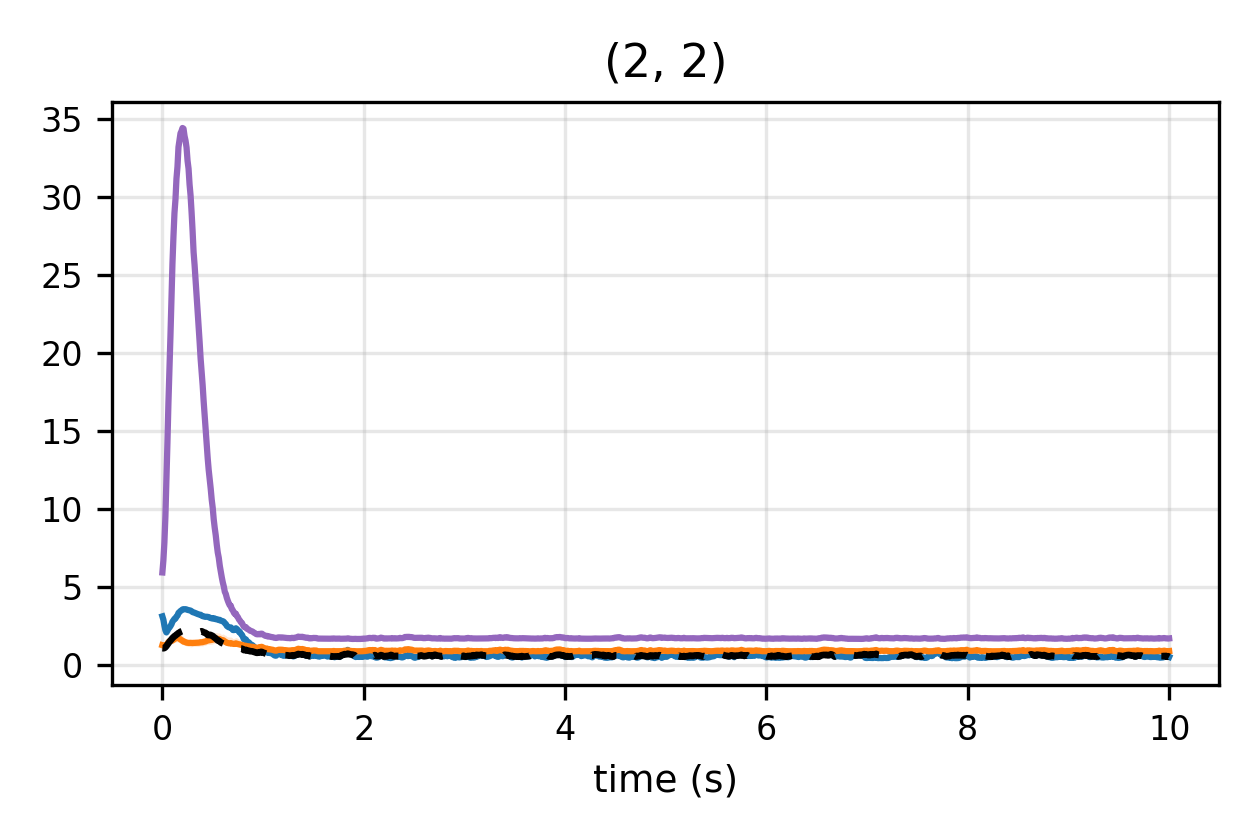}
    \caption{\textbf{Physics Identification in Windy Pendulum System:} Comparison of the learned diagonals of the matrices $M^{-1} G$ for control (top row) and $M^{-1} D$ for damping (bottom row) along the ground-truth swing-up flight (training seed $0$). The models compared are the ground truth (black, dashed), PH-NODE \citep{duong2024phnode} (purple), Lie-PH-NN-SDE (blue), and LieSPHGP (orange). LieSPHGP follows the ground truth for both products; PH-NODE over-estimates the damping about six times, and Lie-PH-NN-SDE has the wrong sign of the $y$ control gain near the upright.}
    \label{fig:pend-physics-id}
\end{figure}

For the windy pendulum, Figure~\ref{fig:pend-physics-id} compares the identified control and damping matrices, while Table~\ref{tab:pend-physics-id} reports the relative identification errors. LieSPHGP outperforms the baselines in identifying both the deterministic physical quantities and the diffusion.

\begin{figure}
    \centering
    \includegraphics[width=0.9\linewidth]{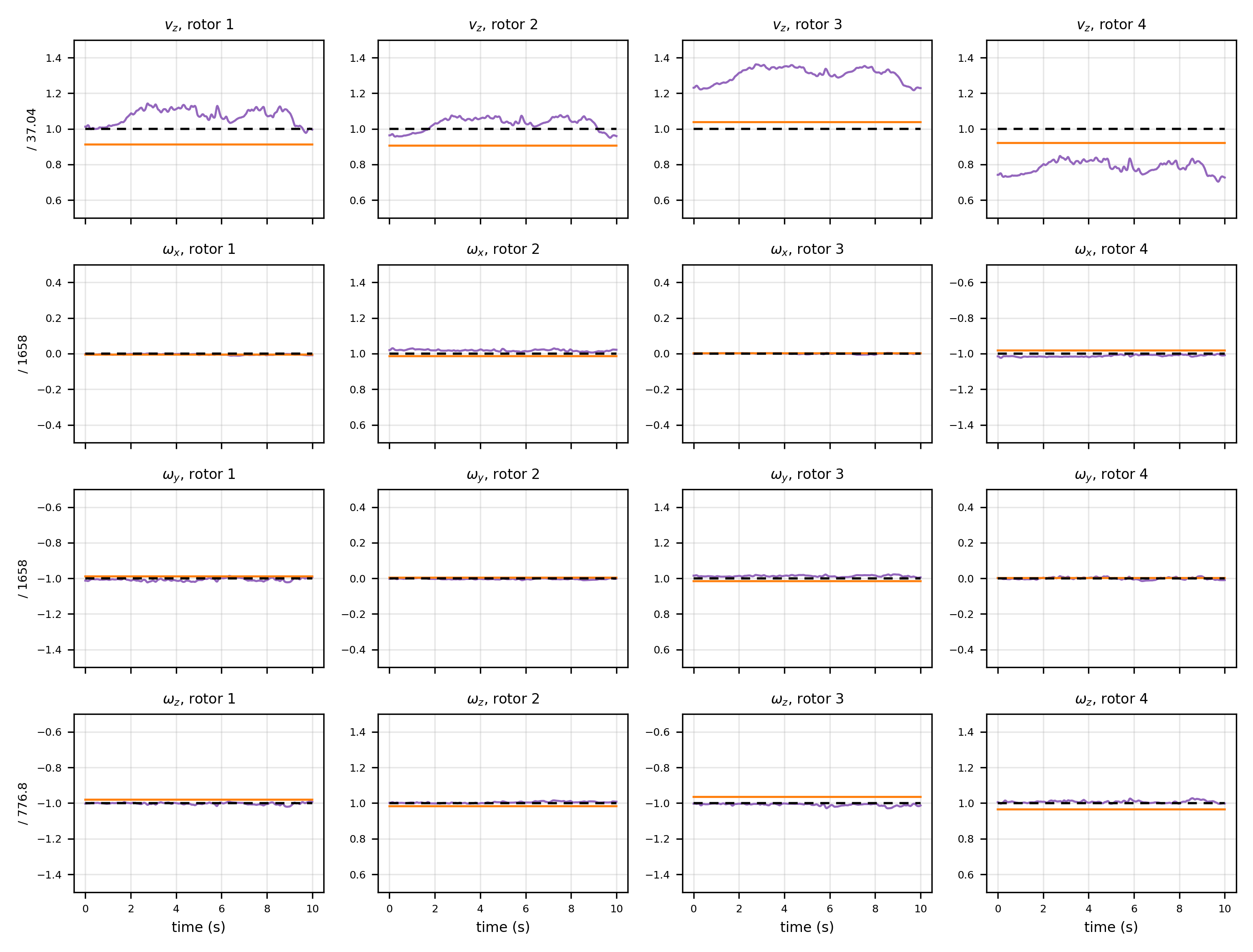}
    \caption{\textbf{Physics Identification in Quadrotor System (Control Matrix):} Actuated rows of the learned control map $M^{-1}G$ along a $10$-s
ground-truth tracking flight (training seed $0$). The top row shows
body-$z$ thrust ($v_z$), and the next three show body torques
($\omega_x,\omega_y,\omega_z$), with one column per rotor. The torque
rows contain the rotor mixing matrix $B_{mix}$. Each row is divided by
the magnitude of its nonzero ground-truth entries ($37.0$, $1658$,
$1658$, and $777$, from top to bottom). The normalized ground truth
(dashed black) is therefore $-1$, $0$, or $1$, and all panels share the
same vertical scale. PH-NODE \citep{duong2024phnode} is shown in purple
and LieSPHGP in orange. The body-$x$ and body-$y$ force rows are omitted
because they are zero in the true model and the controller does not use them.}
    \label{fig:quad-physics-id-control}
\end{figure}

\begin{figure}
    \centering
    \includegraphics[width=0.3\linewidth]{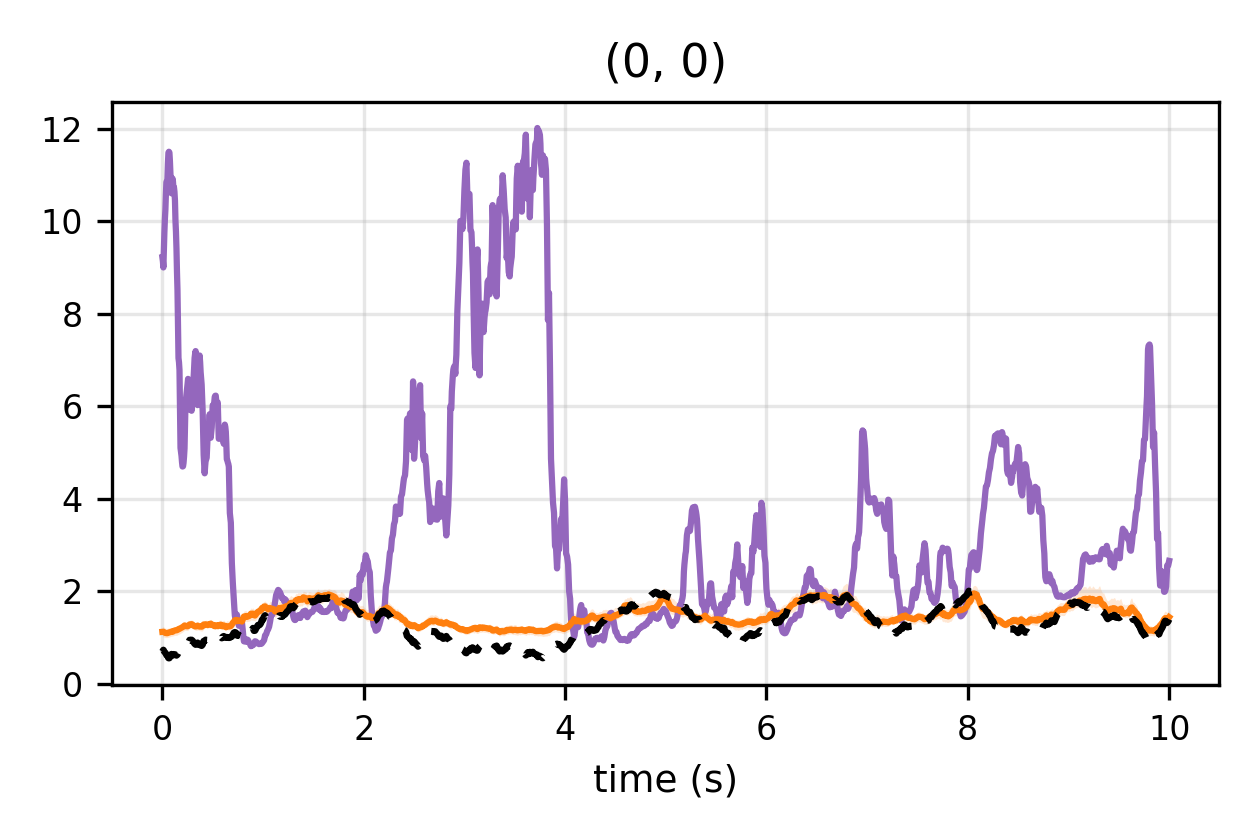}
    \includegraphics[width=0.3\linewidth]{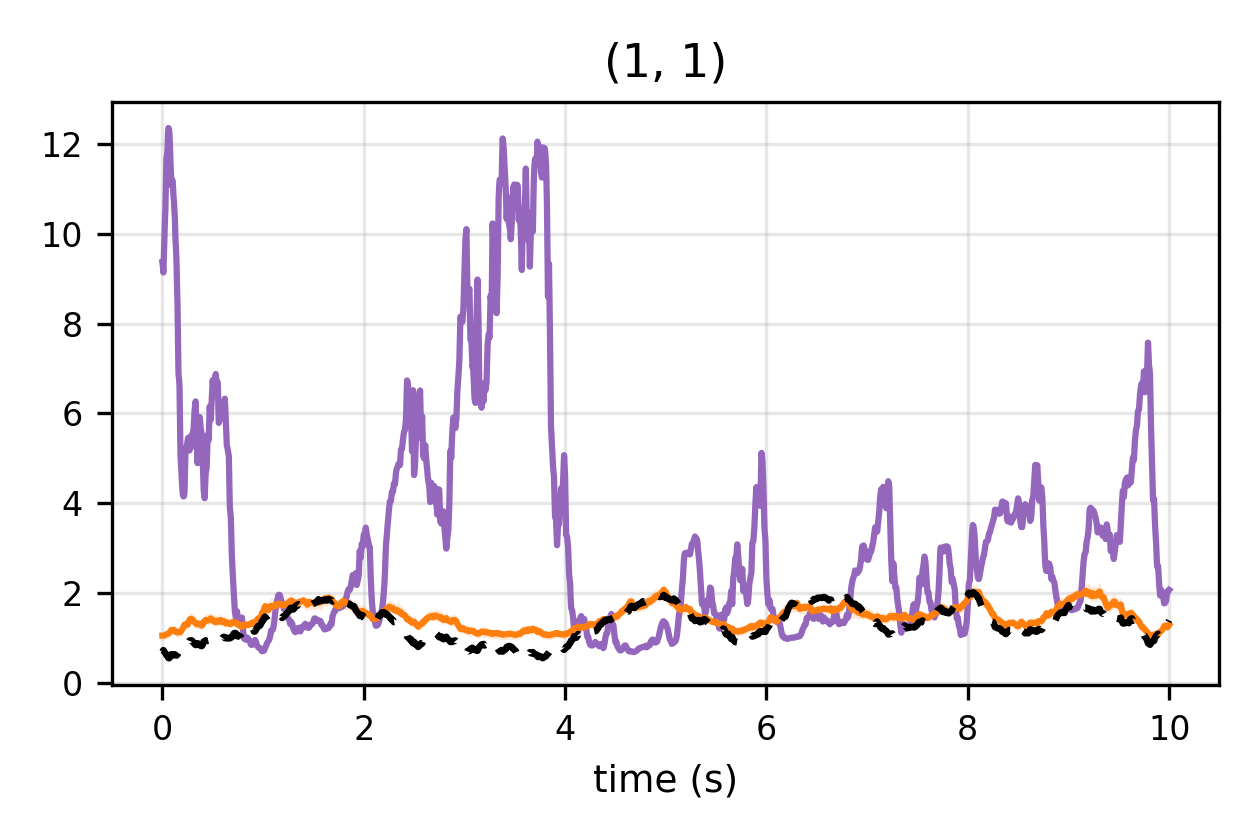}
    \includegraphics[width=0.3\linewidth]{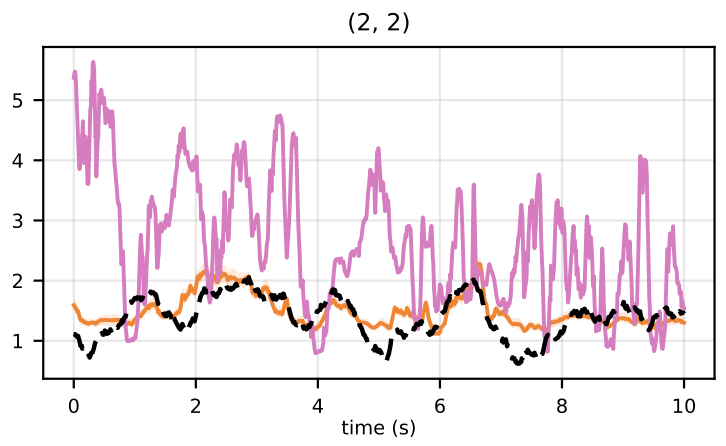}
    \includegraphics[width=0.3\linewidth]{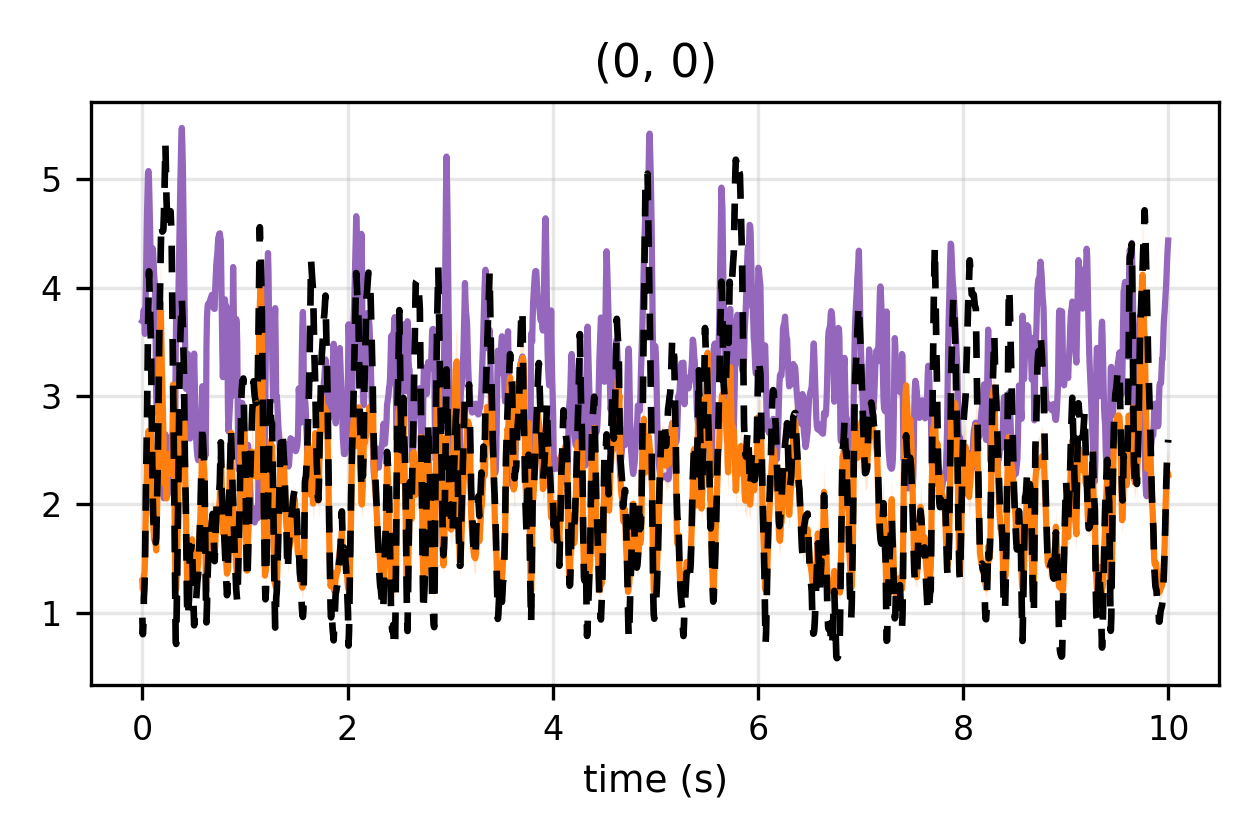}
    \includegraphics[width=0.3\linewidth]{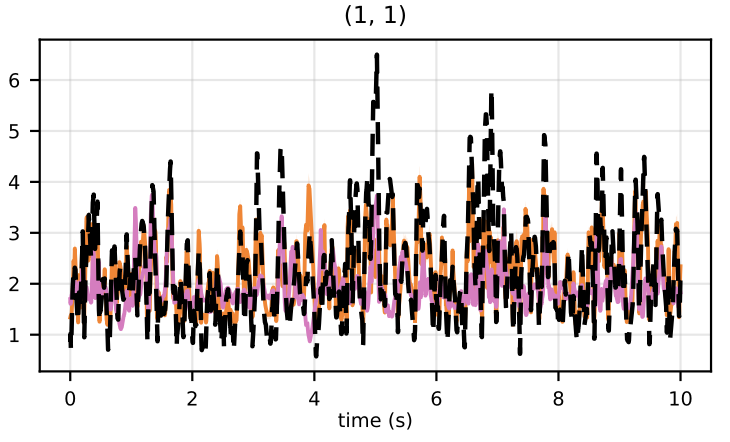}
    \includegraphics[width=0.3\linewidth]{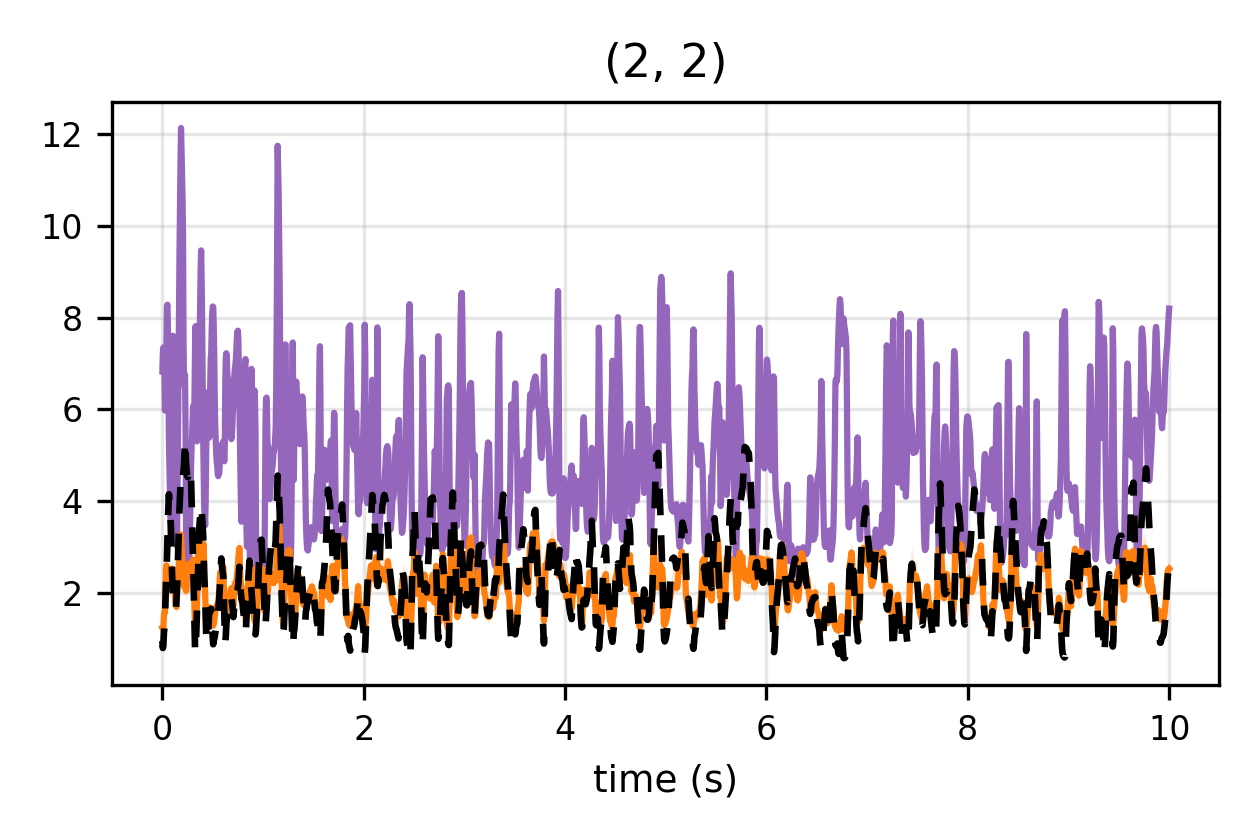}
    \caption{\textbf{Physics Identification in Quadrotor System (Dissipation):} Comparison of the learned diagonals of $M_1^{-1} D_v$ for translational damping (top row) and $M_2^{-1} D_\omega$ for rotational damping (bottom row) along a $10$-s ground-truth tracking flight (training seed $0$). The models compared are the ground truth (black, dashed), PH-NODE \citep{duong2024phnode} (purple), and LieSPHGP (orange). LieSPHGP is closer to the ground truth for both translational and rotational damping, while PH-NODE over-estimates both (Table~\ref{tab:quad-physics-id}).}

    \label{fig:quad-physics-id-damping}
\end{figure}

Figures~\ref{fig:quad-physics-id-control} and~\ref{fig:quad-physics-id-damping}
compare the learned control and damping matrices for the quadrotor.
Table~\ref{tab:quad-physics-id} reports the relative errors. LieSPHGP
recovers the correct torque signs for all four rotors, with a $1.7\,\%$
error in the torque rows of $M^{-1}G$. Its errors in thrust gain, gravity,
and wind intensity are all within $8\,\%$, and its damping errors are
$38\,\%$ (translational) and $46\,\%$ (rotational). PH-NODE achieves a
comparable error in the torque rows ($1.4\,\%$), but its damping errors are
$799\,\%$ and $261\,\%$ and its gravity error is $36\,\%$, that is, about
$21$, $6$ and $5$ times larger than those of LieSPHGP.

\begin{table}[htbp]
\centering
\caption{\textbf{Relative error of the learned pendulum physics:} $\operatorname{mean}\|Q-\widehat Q\|/\operatorname{mean}\|Q\|$
over the states of the $10$ held-out trajectories ($Q$: ground truth, $\hat{Q}$: learned), mean $\pm$ std over $5$
training seeds, bold = best learned model. Diffusion: relative error of the learned diagonal of
$\xi$ against $(0.5,0.5,0)$.}
\begin{tabular}{lccc}
\toprule
\textbf{Product} & \textbf{LieSPHGP (ours)} & \textbf{Lie-PH-NN-SDE (ours)} & \textbf{PH-NODE (prior work)} \\
\midrule
Damping $M^{-1}D$      & $\mathbf{0.44\pm0.11}$   & $0.67\pm0.11$   & $4.96\pm0.07$ \\
Control gain $M^{-1}G$ & $\mathbf{0.057\pm0.003}$ & $0.392\pm0.024$ & $0.347\pm0.009$ \\
Gravity $a_V$          & $\mathbf{0.054\pm0.011}$ & $0.285\pm0.040$ & $0.402\pm0.006$ \\
Diffusion $\xi$     & $\mathbf{0.20\pm0.07}$   & $0.30\pm0.05$   & -- \\
\bottomrule
\end{tabular}
\label{tab:pend-physics-id}
\end{table}

\subsection{Uncertainty Quantification}
\label{sec:appendix:uncertainty}
LieSPHGP separates three sources of uncertainty which are observation noise $\sigma_{\mathrm{obs}}$, process noise through the learned diffusion $\xi$, and the posterior over the GP weights. Table~\ref{tab:noise-identification} compares the learned noise with the values used to generate the data. We assess the EKF's uncertainty on held-out windows using the normalized
innovation squared (NIS),
\begin{equation}
\epsilon_k = r_k^\top S_k^{-1}r_k,
\qquad
\overline{\mathrm{NIS}}
= \frac{1}{n_{out}N}\sum_k \epsilon_k,
\end{equation}
where $r_k=y_k\ominus\hat x_k^-$ is the innovation,
$S_k=P_k^-+R_{\mathrm{obs}}$ is its predicted covariance, and $N$ is the
number of predicted observations
(Section~\ref{sec:methods:loss}, Eq.~\eqref{eq:ekf_nll_step}).
Under a correctly specified Gaussian innovation model,
$r_k\sim\mathcal N(0,S_k)$, so
$\epsilon_k\sim\chi^2_{n_{out}}$ and
$\mathbb E[\overline{\mathrm{NIS}}]=1$
\citep{barshalom2001estimation}. Values consistently above $1$ indicate
overconfidence, which may reflect underestimated noise or errors in the
learned dynamics. Values consistently below $1$ indicate underconfidence. On all three simulated systems the learned observation noise is within $2\,\%$ of the truth, and $\overline{\mathrm{NIS}}$ is $1.00$--$1.01$, i.e.\ the mean NIS is close to its ideal value. The learned diffusion is within $1$--$7\,\%$ of the truth on the quadrotor and moderately over-estimated on the windy pendulum. With the same EKF likelihood, Lie-PH-NN-SDE also recovers the observation noise $(0.494, 0.489)$ but under-estimates the diffusion, $\xi=(0.37, 0.34, 0.00)$, with $\overline{\mathrm{NIS}}=1.16$. SPH-NN, which has no observation-noise model, attributes the observation noise to the dynamics and over-estimates $\xi$ more than fivefold (Table~\ref{tab:sphnn}).

\begin{figure}[t]
    \centering
    \includegraphics[width=\linewidth]{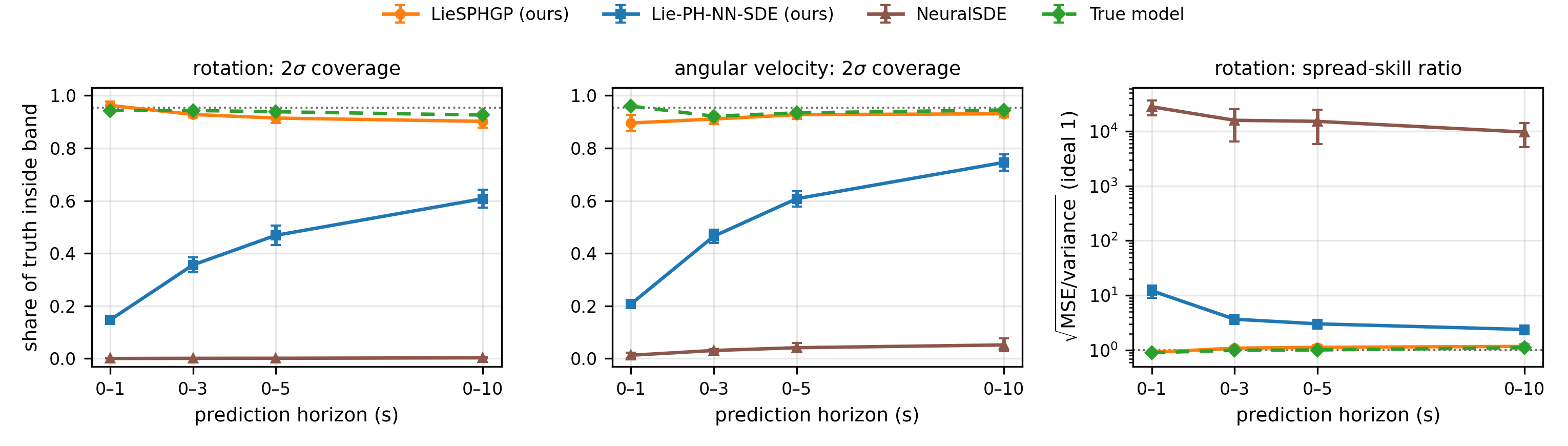}
    \caption{\textbf{Calibration of the predictive distribution on the windy pendulum.} $2\sigma$ coverage of rotation (left) and angular velocity (middle), i.e.\ the share of true states inside the predicted $\pm2\sigma$ band, and the spread-skill ratio $\sqrt{\mathrm{MSE}/\mathrm{variance}}$ of rotation (right, log scale, $1$ = calibrated, $>1$ = over-confident), averaged over $0$--$1$, $0$--$3$, $0$--$5$ and $0$--$10$\,s of open-loop prediction on the $10$ held-out trajectories. $32$ sampled paths per model and the true model is the ground-truth SDE with an independent wind draw. Mean $\pm$ std over training seeds. Dotted lines: nominal $0.954$ and $1$.}
    \label{fig:calibration}
\end{figure}

Figure~\ref{fig:calibration} shows that LieSPHGP's marginal coverage stays close to nominal over the whole $10$-s horizon where its $2\sigma$ coverage stays at $0.90$--$0.96$ for rotation and $0.90$--$0.93$ for angular velocity, close to that of the true SDE sampled with an independent wind draw ($0.92$--$0.96$), and its spread-skill ratio stays between $0.91$ and $1.17$. Lie-PH-NN-SDE learns a smaller diffusion and is over-confident, most strongly at short horizons ($2\sigma$ coverage $0.15$ over $0$--$1$\,s, spread-skill ratio $12$), because the error of its mean dynamics exceeds the predicted spread. The unstructured NeuralSDE is not calibrated (coverage below $0.06$). These marginal diagnostics use $32$ samples per prediction over cumulative horizons. They suggest useful uncertainty estimates but do not establish joint or conditional calibration.
\begin{table}[t]
\centering
\caption{\textbf{Learned noise against the simulators' ground truth.} The EKF likelihood learns the observation noise $\sigma_{\mathrm{obs}}$ and the process noise $\xi$ as separate quantities. Mean over 5 training seeds. NIS is normalized innovation squared of the EKF on the noisy test windows, per dimension (ideal $1$).}
\label{tab:noise-identification}
\begin{tabular}{@{}lcc@{}}
\toprule
\textbf{Quantity} & \textbf{Truth} & \textbf{LieSPHGP (ours)} \\
\midrule
\multicolumn{3}{@{}l}{\emph{Deterministic $SO(3)$ pendulum}} \\
Obs.\ noise $(\sigma_R,\sigma_\omega)$ & $(0.25, 0.25)$ & $(0.247, 0.252)$ \\
NIS per dimension (ideal $1$) & $1$ & $1.00\pm0.01$ \\
\midrule
\multicolumn{3}{@{}l}{\emph{Windy $SO(3)$ pendulum}} \\
Obs.\ noise $(\sigma_R,\sigma_\omega)$ & $(0.5, 0.5)$ & $(0.495, 0.500)$ \\
Diffusion $\xi$ & $(0.5, 0.5, 0)$ & $(0.61, 0.57, 0.06)$ \\
NIS per dimension (ideal $1$) & $1$ & $1.01\pm0.01$ \\
\midrule
\multicolumn{3}{@{}l}{\emph{$SE(3)$ quadrotor}} \\
Obs.\ noise $(\sigma_p,\sigma_R,\sigma_v,\sigma_\omega)$ & $0.25$ & $(0.249, 0.251, 0.250, 0.251)$ \\
Diffusion $\xi_v$ & $(0.5, 0.5, 0.5)$ & $(0.51, 0.51, 0.52)$ \\
Diffusion $\xi_\omega$ & $(0.5, 0.5, 0.5)$ & $(0.53, 0.53, 0.52)$ \\
NIS per dimension (ideal $1$) & $1$ & $1.00\pm0.02$ \\
\bottomrule
\end{tabular}
\end{table}

\subsection{Ablations on the windy pendulum}
\label{sec:appendix:pend-ablations}
Table~\ref{tab:pend-ablations} isolates the effect of the training loss, the posterior over GP weights, and the training window length. 

\emph{Training loss.} We replace the EKF likelihood with PH-NODE's deterministic rollout loss, which minimizes the squared geodesic rotation error and the squared angular velocity error over each training window. When initialized from the first noisy observation, the model estimates the control map and damping poorly, and the resulting controller fails to swing the pendulum up. Learning a separate initial state for each window through multiple shooting recovers most of the prediction accuracy, but $2\sigma$ coverage remains low ($0.27$--$0.43$), and low-gain control is less reliable. 

\emph{GP weights.} We replace the variational posterior over the random-feature weights with a point estimate with weights equal to learned mean ($w=w^{\theta}_{\mathrm{mean}}$). Maximum likelihood estimation (MLE) uses no weight prior, while maximum a posteriori estimation (MAP) replaces the KL term with the Gaussian prior penalty $\tfrac12\lVert w\rVert^2$. Both variants increase the rotation error by $50$--$78\,\%$, reduce $2\sigma$ coverage to $0.60$--$0.67$, and swing the pendulum up in at most $48\,\%$ of the flights at normal gains and $7\,\%$ at low gains.

\emph{Training windows.} Shortening the windows to five points (stride $4$, batch $64$) makes damping harder to estimate from noisy observations. The $M^{-1}D$ error reaches $3.01$, the rotation error rises to $0.92$ rad, compared with $1.07$ rad for PH-NODE, and the low-gain swing-up succeeds in only $8\,\%$ of the flights. In this experiment, the EKF likelihood and the variational weight posterior improve calibration and control, while longer windows improve damping estimation.

\begin{table}[t]
\centering
\caption{\textbf{Ablations on the windy pendulum.} Each variant changes one ingredient of LieSPHGP, except that the rollout-MSE variants, whose loss has no noise model, are also trained without the diffusion and the observation-noise model on same data, $5{,}000$ steps, final checkpoint. Mean $\pm$ std over 3 training seeds and swing-up success at gain scales
$s=1$ / $0.25$.}
\label{tab:pend-ablations}
\setlength{\tabcolsep}{3pt}
\begin{tabular}{@{}lcccccc@{}}
\toprule
\textbf{Variant} & \textbf{Rotation} & \textbf{Rot.\ CRPS} & \textbf{$2\sigma$ cov.} & \textbf{$M^{-1}G$} & \textbf{$M^{-1}D$} & \textbf{Swing-up} \\
\midrule
LieSPHGP                           & $0.288 \pm 0.024$ & $0.099$ & $0.91$ & $0.057$ & $0.43$ & $100$ / $81$ \\
\midrule
\multicolumn{7}{@{}l}{\emph{Loss}} \\
Rollout MSE, noisy start            & $0.628 \pm 0.175$ & $0.248$ & --            & $0.807$ & $2.96$ & $0$ / $0$ \\
Rollout MSE, learned start  & $0.308 \pm 0.024$ & $0.122$ & $0.27$--$0.43$ & $0.187$ & $0.32$ & $100$ / $66$ \\
\multicolumn{7}{@{}l}{\emph{GP weights}} \\
Point estimate, MLE                  & $0.512 \pm 0.030$ & $0.191$ & $0.60$ & $0.359$ & $0.94$ & $9$ / $0$ \\
Point estimate, MAP                & $0.436 \pm 0.109$ & $0.161$ & $0.67$ & $0.283$ & $0.96$ & $48$ / $7$ \\
\multicolumn{7}{@{}l}{\emph{Training windows}} \\
5-point windows                     & $0.922 \pm 0.018$ & $0.308$ & $0.79$ & $0.150$ & $3.01$ & $100$ / $8$ \\
\bottomrule
\end{tabular}
\end{table}

\subsection{Comparison with Stochastic Port-Hamiltonian Neural Networks (SPH-NN)}
\label{sec:appendix:sphnn}

\paragraph{Model.}
SPH-NN \citep{di2026stochastic} represents controlled stochastic port-Hamiltonian
dynamics in Euclidean coordinates as an It\^o SDE:
\begin{equation}
dx = \big[J_\theta(x)-\mathcal R_\theta(x)\big]\nabla H_\theta(x)\,dt
     + G_\theta(x)u\,dt + \xi_\theta(x)\,dW_t .
\label{eq:sphnn}
\end{equation}
The model learns the Hamiltonian $H_\theta$, input map $G_\theta$, and diagonal
diffusion $\xi_\theta$ with neural networks. It enforces skew symmetry through
$J_\theta=A_\theta-A_\theta^\top$ and positive semidefiniteness through
$\mathcal R_\theta=L_\theta^\top L_\theta$. A softplus activation constrains the learned
diffusion entries to be positive. Training maximizes the Euler--Maruyama
likelihood of consecutive observations:
\begin{equation}
y_{k+1}-y_k \mid y_k,u_k
\sim \mathcal N\!\left(
    f_\theta(y_k,u_k)\,\Delta t,\,
    \Sigma_\theta(y_k)\,\Delta t
\right),
\end{equation}
where $f_\theta$ is the drift in Equation~\eqref{eq:sphnn} and
$\Sigma_\theta$ is the transition covariance per unit time.

\paragraph{Windy pendulum setup.}
This is our adaptation of SPH-NN to the present $SO(3)$ observation and control setting. We use the data from Table~\ref{tab:3d-windy-pendulum-comparison} and represent
the state as $x=(\mathrm{vec}\,R,\omega)\in\mathbb R^{12}$. The control input
and wind act on the three angular velocity coordinates, so $G_\theta(x)$ and
$\xi_\theta(x)$ have zero rotation rows. For the transition likelihood, we use one
learned residual variance $s_R^2$ for all nine rotation entries, so
$\Sigma_\theta=\mathrm{diag}(s_R^2 I_9,\,\xi_\theta\xi_\theta^\top)$, where $\xi_\theta$ is
the diagonal angular velocity diffusion.

Each subnetwork has two hidden layers with $64$ tanh units each. We train
with Adam for $5{,}000$ steps, using an initial learning rate of $10^{-3}$,
cosine decay, and gradient clipping at $1$. We evaluate the final checkpoint.
SPH-NN uses all one-step transitions $(y_k,u_k,y_{k+1})$ at every update, while SPH-NN-mb uses random mini-batches of $1{,}024$ transitions. For prediction, we sample $32$ Euler--Maruyama paths.
We estimate $M^{-1}G$ by differentiating the angular velocity drift with
respect to $u$, and estimate gravity by evaluating that drift at
$\omega=0$ and $u=0$. The Euclidean dissipation matrix has no direct
counterpart to $M^{-1}D$.

\begin{table}[t]
\centering
\caption{\textbf{SPH-NN and LieSPHGP on the windy pendulum.}
The data and evaluation protocol follow
Table~\ref{tab:3d-windy-pendulum-comparison}. Values with error bars are
means and standard deviations over the same three training seeds ($0$--$2$).
Swing-up success is reported at normal and low IDA-PBC gains
(Table~\ref{tab:ida-pbc-gain-sweep}).}
\label{tab:sphnn}
\small
\setlength{\tabcolsep}{3pt}
\begin{tabular}{@{}lccc@{}}
\toprule
\textbf{Metric} & \textbf{SPH-NN} & \textbf{SPH-NN-mb} & \textbf{LieSPHGP (ours)} \\
\midrule
Rotation error (rad)                        & $1.71 \pm 0.03$   & $1.75 \pm 0.09$   & $\mathbf{0.288 \pm 0.024}$ \\
Velocity error (rad/s)                      & $4.95 \pm 1.21$   & $5.34 \pm 0.99$   & $\mathbf{0.441 \pm 0.013}$ \\
Energy error                                & $33 \pm 20$       & $27 \pm 15$       & $\mathbf{0.317 \pm 0.013}$ \\
$\max_t\|R_t^\top R_t-I\|_F$                & $347 \pm 105$     & $73 \pm 30$       & $\mathbf{1.9\times10^{-6}}$ \\
Rotation CRPS                               & $0.61 \pm 0.05$   & $0.69 \pm 0.08$   & $\mathbf{0.099 \pm 0.010}$ \\
Rotation $2\sigma$ coverage                 & $0.80 \pm 0.11$   & $0.56 \pm 0.05$   & $\mathbf{0.91 \pm 0.03}$ \\
$M^{-1}G$ relative error                    & $0.995 \pm 0.016$ & $0.255 \pm 0.016$ & $\mathbf{0.057 \pm 0.005}$ \\
Gravity relative error                      & $1.95 \pm 0.14$   & $1.16 \pm 0.08$   & $\mathbf{0.055 \pm 0.006}$ \\
Learned $\xi$ (true $(0.5,0.5,0)$)          & $(2.7,2.6,2.5)$   & $(2.9,2.8,2.6)$   & $(0.62,0.57,0.06)$ \\
Learned obs.\ noise $\sigma_\omega$ (true $0.5$) & not modelled & not modelled & $\mathbf{0.50 \pm 0.00}$ \\
Swing-up success (\%), $s=1$ / $0.25$       & $0$ / $0$         & $0$ / $0$         & $\mathbf{100}$ / $\mathbf{81}$ \\
\bottomrule
\end{tabular}
\end{table}

\paragraph{Results.}
Both SPH-NN variants have about six times the rotation error of LieSPHGP
and fail to swing up the pendulum at either gain setting
(Table~\ref{tab:sphnn}). Their Euclidean rollouts leave $SO(3)$ within the
first second, and the mean peak orthogonality error reaches $347$ for
full-batch SPH-NN. Both variants also overestimate the wind diffusion,
including along the third axis, where the true diffusion is zero. This
is consistent with the one-step likelihood attributing some observation
noise to process noise. LieSPHGP's EKF likelihood models these two noise
sources separately. Mini-batch training improves the control map estimate,
but the rollouts still leave $SO(3)$ and swing-up still fails.

\begin{table}[htbp]
\centering
\caption{\textbf{Relative error of the learned quadrotor physics:} $\operatorname{mean}\|Q-\widehat Q\|/\operatorname{mean}\|Q\|$
over the states of the held-out flights ($Q$: ground truth, $\hat{Q}$: learned), mean $\pm$ std over $3$ training
seeds; bold = best learned model. The control map $M^{-1}G$ ($6\times4$, one column per rotor) is split into its thrust
row (body $z$) and its $3\times4$ torque rows, which contain the rotor mixing matrix $B_{mix}$. Diffusion: relative error
of the six learned entries of $\xi$ against $0.5$.}
\label{tab:quad-physics-id}
\begin{tabular}{lcc}
\toprule
\textbf{Product} & \textbf{LieSPHGP (ours)} & \textbf{PH-NODE (prior work)} \\
\midrule
Thrust row of $M_1^{-1}G$             & $\mathbf{0.080\pm0.004}$ & $0.182\pm0.017$ \\
Torque rows of $M_2^{-1}G$ ($B_{mix}$) & $0.017\pm0.003$          & $\mathbf{0.014\pm0.001}$ \\
Gravity $-M_1^{-1}R^T\nabla V$        & $\mathbf{0.066\pm0.010}$ & $0.355\pm0.024$ \\
Translational damping $M_1^{-1}D_v$   & $\mathbf{0.38\pm0.19}$   & $7.99\pm0.42$ \\
Rotational damping $M_2^{-1}D_\omega$ & $\mathbf{0.46\pm0.22}$   & $2.61\pm0.03$ \\
Diffusion $\xi$                       & $\mathbf{0.050\pm0.039}$ & -- \\
\bottomrule
\end{tabular}
\end{table}

\subsection{Window-style ablation for Noisy SE(3) Quadrotor with stochastic wind}
\label{sec:appendix:quad-windows}
\paragraph{Training windows on the quadrotor.}
Table~\ref{tab:quad-window-ablation} trains both methods in both data styles. With PH-NODE's $5$-sample windows
($0.08$\,s), LieSPHGP is already better than PH-NODE in closed loop ($0.199$ vs.\ $0.255$\,m) and in model-in-the-loop
prediction ($0.163$ vs.\ $0.422$\,m), but both over-estimate the damping, which is barely visible over $0.08$\,s of noisy
data. With $2$-s windows, LieSPHGP identifies the damping to within $33\,\%$ and tracks as well as the true model,
whereas PH-NODE becomes worse ($0.745$\,m closed loop): its deterministic rollout from the noisy first sample over
$100$ steps cannot fit the windows, and its thrust gain and torque map degrade. In our experiments, long windows helped only when the likelihood accounted for process and observation noise.

\begin{table}[t]
\centering
\caption{\textbf{Effect of the training data style on the quadrotor.} Both methods trained on the same rotor-input data
with PH-NODE's style ($5$-sample windows, stride $1$, batch $256$) and with LieSPHGP's style ($101$-sample, i.e.\ $2$-s,
windows, stride $75$, batch $16$); everything else is each method's own recipe ($5{,}000$ steps, final checkpoint).
Training seed $0$; closed loop and model-in-the-loop over $10$ held-out references $\times$ $5$ Brownian motion samples,
$0$--$10$\,s. Lower block: relative identification errors.}
\label{tab:quad-window-ablation}
\setlength{\tabcolsep}{3pt}
\begin{tabular}{@{}lccccc@{}}
\toprule
 & \multicolumn{2}{c}{\textbf{5-sample windows}} & \multicolumn{2}{c}{\textbf{101-sample windows}} & \\
\cmidrule(lr){2-3}\cmidrule(lr){4-5}
\textbf{Metric} & PH-NODE & LieSPHGP & PH-NODE & LieSPHGP & \textbf{True model} \\
\midrule
Closed-loop position RMSE (m)        & $0.255$ & $0.199$ & $0.745$ & $\mathbf{0.130}$ & $0.127$ \\
Closed-loop velocity RMSE (m/s)      & $0.867$ & $0.692$ & $1.615$ & $\mathbf{0.479}$ & $0.464$ \\
Rotor saturation share               & $0.103$ & $0.072$ & $0.296$ & $\mathbf{0.048}$ & $0.040$ \\
Model-in-the-loop position error (m) & $0.422$ & $0.163$ & $0.453$ & $\mathbf{0.113}$ & $0.103$ \\
Model-in-the-loop $2\sigma$ coverage & $0.00$  & $0.47$  & $0.00$  & $\mathbf{0.65}$  & $0.69$ \\
\midrule
Thrust row of $M_1^{-1}G$            & $0.183$ & $0.113$ & $0.429$ & $\mathbf{0.078}$ & -- \\
Torque rows of $M_2^{-1}G$           & $\mathbf{0.013}$ & $0.028$ & $0.131$ & $0.018$ & -- \\
Gravity                              & $0.342$ & $0.087$ & $0.176$ & $\mathbf{0.056}$ & -- \\
$M_1^{-1}D_v$                        & $8.20$  & $3.96$  & $2.92$  & $\mathbf{0.33}$  & -- \\
$M_2^{-1}D_\omega$                   & $2.63$  & $2.72$  & $0.59$  & $\mathbf{0.26}$  & -- \\
\midrule
Training time (min)                  & $3.8$   & $9.1$   & $32.0$  & $60.7$  & -- \\
\bottomrule
\end{tabular}
\end{table}

\subsection{Embedded and Intrinsic Mat\'ern Kernels on SO(3)}
\label{sec:appendix:kernel}

In our model we use Mat\'ern kernels embedded in $SO(3)$ rather than $SO(3)$-intrinsic Mat\'ern kernels because both kernels satisfy the invariance on $SO(3)$ and are positive definite, but the intrinsic features are more computationally expensive than the embedded features.

\paragraph{Both kernels use the group structure.}
Let $\theta$ be the angle of the relative rotation $R^\top R'$, so $\operatorname{tr}(R^\top R')=1+2\cos\theta$. For $R,R'\in SO(3)$,
\begin{equation}
\|R-R'\|_F^2 = 6-2\operatorname{tr}(R^\top R') = 8\sin^2(\theta/2).
\end{equation}
Hence the embedded kernel $k_{\mathrm{emb}}(R,R')=k_{\mathrm M}(\|R-R'\|_F)$ depends only on $\theta$, and it is bi-invariant: $k_{\mathrm{emb}}(QRP,QR'P)=k_{\mathrm{emb}}(R,R')$ for all $Q,P\in SO(3)$. The intrinsic Mat\'ern kernel \citep{borovitskiy2020matern,azangulov2024compact},
\begin{equation}
k_{SO(3)}(R,R')=\frac1Z\sum_{l\ge0}a_l\,(2l+1)\,\chi_l(\theta),\qquad \chi_l(\theta)=\frac{\sin\big((2l+1)\theta/2\big)}{\sin(\theta/2)},
\end{equation}
is also a bi-invariant function of $\theta$. Both kernels are positive definite. They differ only in their shape in $\theta$, and near the identity
\begin{equation}
\|R-R'\|_F=2\sqrt2\sin(\theta/2) = \sqrt2\,\theta +O\left(\theta^3\right),
\end{equation}
so the embedded kernel is close to a Mat\'ern kernel of the geodesic angle; the factor $\sqrt2$ is absorbed by the learned length scale.

\paragraph{The intrinsic features are slower.}
An embedded feature costs one inner product and one cosine, $O(F\,d)$. An intrinsic feature needs a Chebyshev recurrence up to degree $L=8$ plus a sum over degrees, $O(F\,(9+L))$. The EKF loss differentiates every GP twice (Jacobian of the step, then the gradient), which multiplies this cost.

\clearpage

\section{Algorithms}
\label{sec:appendix:algorithms}

\begin{algorithm*}[h]
\caption{Training Lie Group Stochastic Port-Hamiltonian Gaussian Process (Detailed)}
\label{alg:phase1_sde_long}
\small
\begin{algorithmic}[1]

\REQUIRE ~\\
    \qquad\textit{\textbf{Dataset}:} number of trajectory windows $N_{\mathrm{win}}$, trajectories $\mathcal{D} = \{\{y_k^{(j)}\}_{k=0}^{T-1},\{u_k^{(j)}\}_{k=1}^{T-1}\}_{j=1}^{N_{\mathrm{win}}}$, sample interval $\Delta t$ \\

    \qquad\textit{\textbf{GP subnets}:} $M^{-1}_{\theta_1}(q)$ (SPD),\; $V_{\theta_2}(q)$ (scalar),\; $D_{\theta_3}(\zeta)$ (SPD),\; $G_{\theta_4}(q)$ (matrix),\; $\xi_{\theta_5}(q)$ (diagonal); each with posterior $(\mu,\rho)$, level $\lambda$ and kernel hyperparameters $(\sigma_{\mathrm M},\ell_{\mathrm M})$;\; observation noise $(\log\sigma_R,\log\sigma_\omega)$ on $SO(3)$, $(\log\sigma_p,\log\sigma_R,\log\sigma_v,\log\sigma_\omega)$ on $SE(3)$ \\

    \qquad\textit{\textbf{Hyperparameters}:} steps $N_{\text{train}}$,\; batch $B$,\; substeps $N_{\mathrm{sub}}$,\; $\beta_{\max}$,\; anneal steps $K_\beta$,\; clip norm $c$

\ENSURE Trained parameters $\theta = \{\mu,\rho,\lambda,\sigma_{\mathrm M},\ell_{\mathrm M} \text{ of each subnet},\ \log\sigma_{\mathrm{obs}}\}$ after $N_{\text{train}}$ steps

\STATE \textbf{\COMMENT{Initialize}}
Draw the fixed random features $(\Omega,b)$ of each subnet (Alg.~\ref{alg:gp_layer}); initialize $\theta$ as in Sec.~\ref{sec:methods:loss} 
\STATE $N_{\mathrm{obs}} \gets N_{\mathrm{win}}\cdot(T-1)$, \quad $h \gets \Delta t / N_{\mathrm{sub}}$

\vspace{0.3em}
\FOR{$k = 1, \cdots, N_{\text{train}}$}
    \STATE \textbf{\COMMENT{ Sample GP posterior}}
    Draw a batch $\mathcal{B}$ of $B$ windows and one weight sample per subnet via reparameterisation (Sec.~\ref{sec:methods:gp}), shared by the whole batch, with $W = \mu + e^{\rho}\odot\epsilon, \; \epsilon \sim \mathcal{N}(0,I)$.

        \STATE For each window $j\in\mathcal{B}$, set $\hat x_0 = y_0^{(j)}$, $P_0 = R_{\mathrm{obs}}$, $\mathcal{N}^{(j)}\gets 0$ and, for $t = 1,\dots,T-1$:
    \STATE \textbf{\COMMENT{Forward dynamics integration}}
    Propagate the filtered state with $N_{\mathrm{sub}}$ \textsc{LieIMEXSolver} steps (Sec.~\ref{sec:methods:integrator}, Alg.~\ref{alg:lieimexsolver}) under the held control, with zero noise, and take the Jacobians of the step by forward-mode differentiation:
    \[
    \begin{aligned}
        \hat x_t^-,\ (A_t,B_t)
        \gets {}&\textbf{LieIMEXSolver}\big(\hat x_{t-1},\,u_t^{(j)},\,0;\ \theta^{(W)}\big)
    \end{aligned}
    \]
    \STATE \textbf{\COMMENT{EKF step}}
     Correct the state, predict the error covariance, score the noisy sample, and accumulate total loss (Sec.~\ref{sec:methods:loss}):
    \[
    \begin{aligned}
        \hat x_t,\ P_t,\ \ell_t^{(j)}
        \gets {}&\textbf{EKF}\big(\hat x_t^-,\,P_{t-1},\,A_t,\,B_t,\,y_t^{(j)};\ \xi,\,\sigma_R,\,\sigma_\omega\big), \qquad \mathcal{N}^{(j)} \gets \mathcal{N}^{(j)} + \ell_t^{(j)}
    \end{aligned}
    \]

    \STATE \textbf{\COMMENT{Compute loss}}
    Compute the loss $\mathcal{L}$ 
    (Sec.~\ref{sec:methods:loss}):
    \[
        \mathcal{L} \gets
        \frac{1}{n_{out}\,B\,(T-1)}\sum_{j\in\mathcal{B}}\mathcal{N}^{(j)} +
        \beta_k\,\frac{\mathrm{KL}(q\,\|\,p)}{n_{out}\,N_{\mathrm{obs}}}, \qquad \beta_k = \beta_{\max}\cdot\min\!\left(1,\, k/K_\beta\right)
    \]

    \STATE \textbf{\COMMENT{Backpropagate gradient}} $g\gets\nabla_\theta\mathcal{L}$ through the filter and the integrator. Clip $g$ to global norm $c$.

    \STATE \textbf{\COMMENT{Parameter update}} For every parameter group $k$ (the level, GP weights, $\sigma_{\mathrm M}$ and $\ell_{\mathrm M}$ of each subnet, and the observation noise): $\theta_k\gets\mathrm{Prodigy}_k(\theta_k,g_k)$, each with its own learning rate $\eta_k(t)$ and a cosine learning-rate schedule

\ENDFOR

\end{algorithmic}
\end{algorithm*}

\begin{algorithm}[p]
\caption{3D Windy Pendulum on $\SO$ Environment --- Data Generation}
\label{alg:windy_pendulum_3d}
\begin{algorithmic}[1]

\REQUIRE ~\\
    \qquad\textit{Physical:} mass $m$, length $\ell$, gravity $g$, timestep $\Delta t$ \\
    \qquad\textit{External force:} amplitude $A$, type $\in\{\texttt{sine},\texttt{square},\texttt{random},\texttt{constant}\}$, direction $\hat{\mathbf{d}}$ \\
    \qquad\textit{Friction:} coefficient $\bm{\mu}_0$, mode $\in\{\texttt{fixed},\texttt{varying}\}$ \\
    \qquad\textit{Stochastic (Wind):} process noise $\sigma$ \\
    \qquad\textit{Observation:} noise $\sigma_{\mathrm{obs}}$

\STATE \textbf{\COMMENT{--- State \& Parameters ---}}
\STATE $q = \bR \in \SO$,\quad Angular Velocity $\bomega \in \R^3$,\quad Control Input $\bu \in \R^3$
\STATE $\bI = m\ell^2\,\mathbf{I}_3,\quad \be_z = (0,0,1)^\top$
\STATE Bob position (world frame): $\mathbf{r}_t = \ell\,q_t\,\be_z$

\vspace{0.3em}
\STATE \textbf{\COMMENT{--- External Forces ---}}
\[
    w(t) = \begin{cases}
        A\sin(\pi t)                    & \texttt{sine}     \\
        A\cdot\mathrm{sign}(\sin(\pi t))& \texttt{square}   \\
        \mathcal{N}(0,A^2)              & \texttt{random}   \\
        A                               & \texttt{constant}
    \end{cases}, \qquad
    \btau_{\mathrm{ext}}(q_t, t) = q_t^\top\!\bigl(\mathbf{r}_t \times w(t)\,\hat{\mathbf{d}}\bigr)
\]

\vspace{0.3em}
\STATE \textbf{\COMMENT{--- Friction ---}}

\[
    \bm{\mu} = \begin{cases}
        \bm{\mu}_0
        & \text{constant} \\[4pt]
        \bm{\mu}_0 \,\bigl(1 + \norm{\bomega_t}\bigr)
        & \text{varying}
    \end{cases}, \qquad
    \btau_{\mathrm{fric}}(\bomega_t) = \bm{\mu} \odot \bomega_t
\]

\vspace{0.3em}
\STATE \textbf{\COMMENT{--- Torque ---}}
\[
    \btau(q_t,\bomega_t) = \underbrace{q_t^\top(\mathbf{r}_t \times {-mg\be_z})}_{\text{gravity}}
          + \underbrace{\btau_{\mathrm{ext}}(q_t, t)}_{\text{external}}
          + \underbrace{G\,\bu_t}_{\text{control}}
          - \underbrace{\btau_{\mathrm{fric}}(\bomega_t)}_{\text{friction}}
\]

\vspace{0.3em}
\STATE \textbf{\COMMENT{--- Angular Acceleration ---}}
\[
    \dot{\bomega}(q_t, \bomega_t) = \bI^{-1}\!\bigl(\btau(q_t,\bomega_t) - \bomega_t \times \bI\bomega_t\bigr)
\]

\vspace{0.3em}
\STATE \textbf{\COMMENT{--- Trajectory Rollout ---}}
\FOR{$t = 0$ to $T-1$}
    \STATE Advance the physical state with \textsc{LieIMEXSolver} (Alg.~\ref{alg:lieimexsolver}), using the ground-truth drift, damping, and diffusion for data generation.
\ENDFOR

\vspace{0.3em}
\STATE \textbf{\COMMENT{--- Observation Noise (Geometrically Correct) ---}}

\STATE For each $t$:
\[
    \bm{\varepsilon}_t \sim \mathcal{N}(\mathbf{0},\,\sigma_{\mathrm{obs}}^2\,\mathbf{I}_3), \qquad
    \bm{\delta}_t \sim \mathcal{N}(\mathbf{0},\,\sigma_{\mathrm{obs}}^2\,\mathbf{I}_3)
\]

\STATE Rotation noise via $\expmap$ (preserves $\SO$):
\[
    \tilde{q}_t = q_t\;\expmap(\bm{\varepsilon}_t), \qquad
    \theta = \norm{\bm{\varepsilon}_t},\quad E = \hatmap{\bm{\varepsilon}_t}, \quad
    \expmap(\bm{\varepsilon}_t) = \mathbf{I}
        + \frac{\sin\theta}{\theta}\,E
        + \frac{1-\cos\theta}{\theta^2}\,E^2
\]
\STATE Angular velocity noise (Euclidean addition):
$
    \tilde{\bomega}_t = \bomega_t + \bm{\delta}_t
$
\STATE Control $\bu_t$ is \textbf{not} corrupted. Noisy trajectory:
\[
    \tilde{\mathbf{o}}_t = \bigl[\,\mathrm{vec}(\tilde{q}_t)^\top,\;\tilde{\bomega}_t^\top,\;\bu_t^\top\,\bigr]^\top \in \R^{15}
\]

\end{algorithmic}
\end{algorithm}

\begin{algorithm}[p]
\caption{\textsc{LieIMEXSolver}: Stochastic Lie group IMEX step on $SE(3) \times \mathfrak{se}(3)$}
\label{alg:lieimexsolver}
\small
\begin{algorithmic}[1]
\REQUIRE State $s_n=(x_n,R_n,\zeta_n)$, control $u_n$, step $h$, full drift $f_\theta$, effective damping $K_\theta$, diffusion $\Gamma_\theta$
\ENSURE $s_{n+1}=(x_{n+1},R_{n+1},\zeta_{n+1})$ with $R_{n+1}\in SO(3)$
\STATE Sample one increment $\Delta W\sim\mathcal N(0,hI)$ and reuse it in predictor and corrector.
\STATE Evaluate $f_n=f_\theta(s_n,u_n)$ and $K_n=K_\theta(s_n)$.
\STATE Extract $\dot\zeta_{\mathrm{full},n}$ from $f_n$ and set $F_n\gets\dot\zeta_{\mathrm{full},n}+K_n\zeta_n$.
\STATE Evaluate $\Gamma_n=\Gamma_\theta(s_n)$.
\STATE Solve the predictor
\[ (I+hK_n)\zeta_p=\zeta_n+hF_n+\Gamma_n\Delta W. \]
\STATE Predict configuration: $x_p\gets x_n+h\dot x_n$ and $R_p\gets R_n\exp(h\widehat\omega_n)$.
\STATE Form $s_p=(x_p,R_p,\zeta_p)$; evaluate $f_p,K_p$ and set $F_p\gets\dot\zeta_{\mathrm{full},p}+K_p\zeta_p$.
\STATE Solve the linearly implicit corrector
\[ \left(I+\tfrac h2K_p\right)\zeta_{n+1}=\zeta_n+\tfrac h2(F_n+F_p-K_n\zeta_n)+\Gamma_n\Delta W. \]
\STATE Correct translation: $x_{n+1}\gets x_n+\tfrac h2(\dot x_n+\dot x_p)$.
\STATE Correct rotation: $R_{n+1}\gets R_n\exp\!\left(\tfrac h2\widehat{(\omega_n+\omega_p)}\right)$ using Rodrigues' formula.
\STATE \textbf{return} $s_{n+1}$
\end{algorithmic}
\end{algorithm}

\begin{algorithm}[p]
\caption{Mat\'ern random-feature GP subnetwork (used for every learned term on $SO(3)$ and $SE(3)$)}
\label{alg:gp_layer}
\small
\begin{algorithmic}[1]
\REQUIRE ~\\
    \qquad\textit{Input:} $z\in\mathbb R^{d}$ (one of $\mathrm{vec}(R)\in\mathbb R^9$, $\omega\in\mathbb R^3$, $v\in\mathbb R^3$, $x\in\mathbb R^3$, or $[\mathrm{vec}(R),x]\in\mathbb R^{12}$) \\
    \qquad\textit{Kernel:} $F$ features, smoothness $\nu$ \\
    \qquad\textit{Trainable:} variational mean $\mu\in\mathbb R^{F\times d_{\mathrm{out}}}$, log-std $\log\sigma\in\mathbb R^{F\times d_{\mathrm{out}}}$, level $\lambda$, log-amplitude $\log\sigma_{\mathrm M}$, log-length-scale $\log\ell_{\mathrm M}$ \\
    \qquad\textit{Flags:} \textsc{sample} (training) or posterior mean (evaluation); \textsc{spd} (symmetric positive definite) for $M^{-1}$ and $D$; \textsc{lin} (linear level) for $V$; \textsc{exp} for each diffusion channel of $\xi$
\ENSURE Output $y$ (scalar, matrix, or symmetric positive definite matrix) and $\mathrm{KL}$

\FOR{$k=1,\dots,F$}
    \STATE $\zeta_k\sim\mathcal N(0,I_d)$,\quad $c_k\sim\chi^2_{2\nu}$,\quad $b_k\sim\mathcal U(0,2\pi)$
    \STATE $\Omega_k\gets \dfrac{\zeta_k}{\sqrt{c_k/(2\nu)}}$ \COMMENT{Student-$t$ spectral sample of the Mat\'ern-$\nu$ kernel}
\ENDFOR

\STATE \textbf{\COMMENT{--- Forward pass ---}}
\STATE $\mathbf h\gets\sigma_{\mathrm M}\sqrt{2/F}\,\cos(\Omega z/\ell_{\mathrm M}+b)\in\mathbb R^{F}$ \COMMENT{$\mathbf h^\top\mathbf h'\approx k_\nu(\|z-z'\|)$; chordal distance for $\mathrm{vec}(R)$}
\IF{\textsc{sample}}
    \STATE $\epsilon\sim\mathcal N(0,I)$,\quad $W\gets\mu+\exp(\log\sigma)\odot\epsilon$ \COMMENT{one sample per training step, shared by all states in the batch}
\ELSE
    \STATE $W\gets\mu$
\ENDIF
\IF{\textsc{lin}}
    \STATE $a\gets\lambda^\top z_{\mathrm{lin}}+W^\top\mathbf h$ \COMMENT{$z_{\mathrm{lin}}=\mathrm{vec}(R)$ on $SO(3)$, $z_{\mathrm{lin}}=x$ on $SE(3)$}
\ELSE
    \STATE $a\gets\lambda+W^\top\mathbf h\in\mathbb R^{d_{\mathrm{out}}}$
\ENDIF
\IF{\textsc{spd}}
    \STATE $L\gets\begin{bmatrix} e^{a_1}&0&0\\ a_4&e^{a_2}&0\\ a_5&a_6&e^{a_3}\end{bmatrix}$,\quad $y\gets LL^\top$ \COMMENT{$d_{\mathrm{out}}=6$}
\ELSIF{\textsc{exp}}
    \STATE $y\gets\exp(a)$ \COMMENT{$d_{\mathrm{out}}=1$; $\Sigma_i(q)=\exp(\lambda_i+\mathrm{GP}_i(q))>0$, one GP per channel: $3$ on $SO(3)$, $6$ on $SE(3)$}
\ELSE
    \STATE $y\gets\mathrm{reshape}(a)$ \COMMENT{scalar for $V$; $3\times3$ for $G$ in $SO(3)$ and $6\times4$ for $G$ in $SE(3)$}
\ENDIF
\STATE $\mathrm{KL}\gets\tfrac12\sum_{k,j}\big(\sigma_{kj}^2+\mu_{kj}^2-1-2\log\sigma_{kj}\big)$ \COMMENT{$\mathrm{KL}(q(W)\,\|\,\mathcal N(0,I))$}
\STATE \textbf{return} $y,\ \mathrm{KL}$
\end{algorithmic}
\end{algorithm}

\end{document}